\documentclass[journal]{IEEEtran}

\usepackage{cite}
\usepackage{amsmath,amssymb,amsfonts}
\usepackage{graphicx}
\usepackage{textcomp}

\usepackage{algorithm}
\usepackage{algpseudocode}

\floatname{algorithm}{Algorithm}

\usepackage{subcaption}

\usepackage{booktabs}
\usepackage{tabularx}
\usepackage{ragged2e}
\usepackage{array}
\usepackage{makecell}
\usepackage{enumitem}
\usepackage{float}

\usepackage[T1]{fontenc}
\usepackage{seqsplit}

\usepackage{listings}
\usepackage{xcolor}
\usepackage{hyperref}
\hypersetup{
    colorlinks=true,
    urlcolor=blue,
    linkcolor=blue,
    citecolor=blue
}
\usepackage{caption}
\usepackage{framed}

\newenvironment{titlebox}[1]{%
  \def\boxTitle{#1}%
  \begin{framed}
    \noindent
    \colorbox{black}{%
      \parbox{\dimexpr\linewidth-2\fboxsep\relax}{%
        \bfseries\color{white}\boxTitle
      }%
    }%
    \par\smallskip
}{%
  \end{framed}%
}

\usepackage{bm}

\usepackage{url}
\usepackage{xurl}
\def\BibTeX{{\rm B\kern-.05em{\sc i\kern-.025em b}\kern-.08em
    T\kern-.1667em\lower.7ex\hbox{E}\kern-.125emX}}

\providecommand{\history}[1]{}
\providecommand{\doi}[1]{}
\providecommand{\corresp}[1]{}

\begin{document}
% \title{Towards EnergyGPT: A Large Language Model Specialized for the Energy Sector}
\title{Towards EnergyGPT: A Large Language Model Specialized for the Energy Sector%
\thanks{Repository: \url{https://github.com/fitila/energygpt-release}}}

\IEEEaftertitletext{%
\vspace{-1.6em}
\begin{center}
\small
\setlength{\tabcolsep}{18pt}
\begin{tabular}{c c}
\begin{minipage}[t]{0.38\textwidth}
\centering
\textbf{Amal Chebbi, Ph.D.}\\
Data Scientist\\
Fitila Technologies\\
Chicago, IL\\
\texttt{amal.chebbi@fitila.ai}
\end{minipage}
&
\begin{minipage}[t]{0.38\textwidth}
\centering
\textbf{Babajide Kolade, Ph.D., P.E.}\\
Founder and Technical Director\\
Fitila Technologies\\
Chicago, IL\\
\texttt{babajide.kolade@fitila.ai}
\end{minipage}
\end{tabular}
% \end{center}
% \vspace{2em}

% \vspace{1em}
% \small\textit{Repository:} \url{https://github.com/fitila/energygpt-release}
\end{center}
\vspace{2em}

}

% \author{Amal Chebbi and Babajide Kolade}

% \IEEEaftertitletext{%
% \vspace{-1.2em}
% \begin{center}
% \small
% \begin{minipage}[t]{0.47\textwidth}
%   \centering
%   \textbf{Amal Chebbi, Ph.D.}\\
%   Data Scientist\\
%   Fitila Technologies\\
%   Chicago, IL\\
%   \texttt{amal.chebbi@fitila.ai}
% \end{minipage}
% \hfill
% \begin{minipage}[t]{0.47\textwidth}
%   \centering
%   \textbf{Babajide Kolade, Ph.D., P.E.}\\
%   Founder and Technical Director\\
%   Fitila Technologies\\
%   Chicago, IL\\
%   \texttt{babajide.kolade@fitila.ai}
% \end{minipage}
% \end{center}
% \vspace{0.6em}
% }

\maketitle

\begin{abstract}
Large language models have demonstrated impressive capabilities across various domains. However, their general-purpose nature often limits their effectiveness in specialized fields such as energy, where deep technical expertise and precise domain knowledge are essential. In this paper, we introduce EnergyGPT, a domain-specialized language model tailored for the energy sector, developed by fine-tuning the LLaMA 3.1-8B model on a high-quality, curated corpus of energy-related texts. We consider two adaptation strategies: a full-parameter Supervised Fine-Tuning variant and a parameter-efficient LoRA-based variant that updates only a small fraction of the model parameters. We present a complete development pipeline, including data collection and curation, model fine-tuning, benchmark design, LLM-judge choice and calibration, evaluation, and deployment. Through this work, we demonstrate that our training strategy enables improvements in domain relevance and performance without the need for large-scale infrastructure. By evaluating the performance of both EnergyGPT variants using domain-specific question-answering benchmarks, our results  show that the adapted models consistently outperform the base model in most energy-related language understanding and generation tasks, with the LoRA variant achieving competitive gains at significantly reduced training cost.
\end{abstract}


\maketitle

\section{Introduction}
\label{sec:introduction}
Large language models (LLMs) have transformed natural language processing by demonstrating strong capabilities in text generation, question answering, summarization, and reasoning across a wide range of tasks \cite{brown2020language,openai2023gpt4}. Despite these advances, general-purpose LLMs often underperform in specialized technical domains, where effective responses depend not only on general linguistic fluency but also on accurate interpretation of domain-specific terminology, engineering context, and task-relevant knowledge \cite{gururangan2020dont,ling2024domain}. The energy sector is a particularly challenging domain in this regard. It spans multiple subfields, including power systems, renewable energy, oil and gas, energy policy, and engineering operations, each characterized by technical language, heterogeneous documentation, and highly contextual decision-making requirements. As a result, general LLMs may generate responses that are fluent yet incomplete, overly generic, or technically unreliable when applied to energy-related tasks \cite{gabber2024renewablehydrogenllm,cheng2025gaia,zhang2026llms_energy_systems,mirshekali2025llm_energy_review}.

\begin{figure}[ht]
    \centering
    \includegraphics[width=0.5\textwidth]{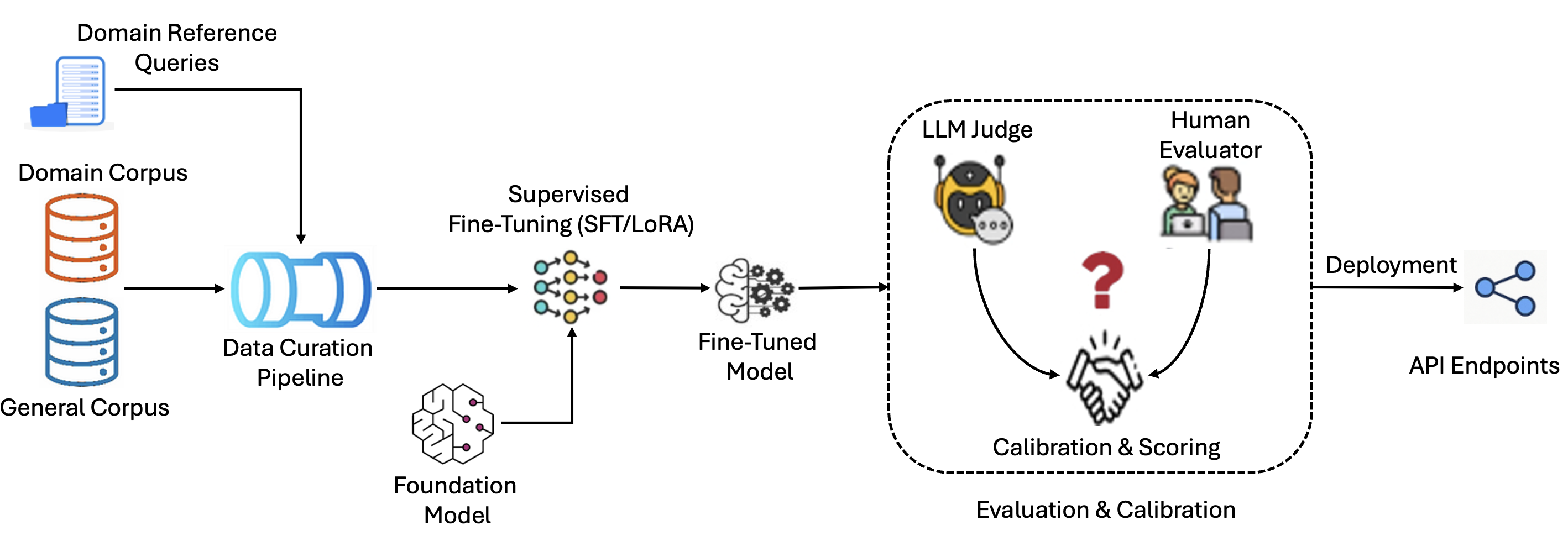}
    \caption{Generalizable pipeline for building domain-specialized assistants.}
    \label{fig:general_pipeline}
\end{figure}

The need for domain-specialized language models is growing as energy organizations seek intelligent tools for technical question answering, knowledge retrieval, maintenance support, research assistance, and operational decision support \cite{gabber2024renewablehydrogenllm,cheng2025gaia}. However, building such systems remains difficult in practice. Training a domain-specific model from scratch is computationally prohibitive for most organizations, while continual pretraining pipelines typically require large-scale domain corpora, substantial compute resources, and careful data engineering \cite{ke2023continual,xie2024efficient,ibrahim2024simple,cottier2024rising}. In contrast, supervised fine-tuning (SFT) and parameter-efficient adaptation methods offer a more practical path for adapting an existing open-weight LLM to a specialized domain under realistic resource constraints \cite{hu2021lora,dettmers2023qlora,pareja2024unveiling}. Nevertheless, the effectiveness of these adaptation strategies for broad energy-domain specialization remains insufficiently studied, particularly in settings that jointly consider corpus curation, model adaptation, evaluation, and deployment.

Recent studies have explored LLM adaptation in narrower energy-related subdomains, including renewable energy \cite{gabber2024renewablehydrogenllm}, oil and gas \cite{chebbi2024rag_oilgas}, and power-dispatch applications \cite{cheng2025gaia}. These efforts demonstrate the promise of domain-specialized LLMs for energy applications, but they do not yet provide a comprehensive end-to-end framework for adapting, evaluating, and deploying a language model using a broader energy domain corpus. To address this gap, we introduce \textit{EnergyGPT}, a specialized language model tailored for energy-sector applications. Starting from the open-weight LLaMA 3.1-8B model \cite{meta2024llama3}, we adapt the base model using a curated energy-focused corpus and investigate two complementary strategies: full-parameter supervised fine-tuning and parameter-efficient LoRA-based adaptation \cite{hu2021lora}. This controlled setup allows us to examine how a general-purpose LLM can be specialized for the energy domain while balancing domain performance against training cost.

Beyond model adaptation itself, this work aims to provide a practical and reproducible pipeline for domain specialization. We construct an energy-domain corpus from multiple sources, develop an energy-specific benchmark spanning multiple task formats, calibrate LLM-as-a-judge evaluation against human expert assessment, and demonstrate a deployment pathway from local serving to cloud-based access. In doing so, we frame EnergyGPT not only as a specialized model instance, but also as a case study in how open-weight LLMs can be adapted for specialized domains where accuracy, efficiency, and operational usability all matter.

The main contributions of this paper are as follows:
\begin{itemize}
    \item \textbf{Energy-domain corpus construction:} We curate a large-scale corpus for energy-domain specialization by combining scientific literature with semantically filtered energy-related text, and document the associated preprocessing, quality filtering, deduplication, and relevance-filtering pipeline.

    \item \textbf{Controlled domain adaptation of an open-weight LLM:} We specialize LLaMA 3.1-8B \cite{meta2024llama3} using two adaptation strategies: full supervised fine-tuning (EnergyGPT-SFT) and parameter-efficient LoRA tuning (EnergyGPT-LoRA); under the same corpus and training setup, enabling a direct comparison of performance and efficiency trade-offs.
    
    \item \textbf{Energy-specific evaluation framework:} We develop a benchmark covering multiple-choice, true/false, and open-ended energy questions across different difficulty levels, and combine human expert ratings with calibrated LLM-as-a-judge evaluation for scalable assessment.
    
    \item \textbf{Deployment-oriented specialization pipeline:} We demonstrate an end-to-end workflow that spans data curation, model adaptation, evaluation, and deployment through on-premises NIM serving and Azure-based API integration.
    
    \item \textbf{A reusable recipe for domain specialization of LLMs:} We present a scalable framework that can be extended beyond energy to support the development of specialized AI assistants in other fields as summarized in Figure~\ref{fig:general_pipeline}.
\end{itemize}

The remainder of this paper is organized as follows: Section~\ref{sec:2} reviews related work on domain-specific LLMs and supervised fine-tuning methods. Section~\ref{sec:3} details the data collection, preprocessing, and structuring techniques used to curate a high-quality energy-related dataset. Section~\ref{sec:4} describes the model adaptation strategy, including both the full SFT and LoRA-based variants, along with the training setup and hyperparameter choices. Section~\ref{sec:5} introduces the benchmark and the methodology for selecting and calibrating an LLM judge. Section~\ref{sec:6} presents the experimental results, comparing EnergyGPT against the baseline model and analyzing the differences between EnergyGPT-SFT and EnergyGPT-LoRA. Section~\ref{sec:7} details the deployment setup. Section~\ref{sec:generalizability} discusses how the proposed pipeline can be transferred to other specialized domains beyond energy. Section~\ref{sec:8} discusses limitations, challenges, and future research directions. Finally, Section~\ref{sec:9} concludes the paper.

%%%%%%%%%%%%%%%%%%%%%%%%%%%%%%%%%%%%%%%%%%%%%%%%%%%%%%%%%%%%%%%%%%%%%%%%%%%%%%%%%%%%%%%%%%%%%%%%%%%%%%%%%%%%%%%%%%%%%%%%%%%%%%%%%%%%

\section{Related Work}
\label{sec:2}

Large language models have achieved remarkable success in various domains, demonstrating strong capabilities in text generation, question answering, and reasoning. However, many specialized domains such as healthcare \cite{lee2019biobert}, finance \cite{wu2023bloomberggpt}, climate \cite{webersinke2022climatebert}, law and technology possess unique linguistic nuances, technical terminologies, and contextual dependencies that general-purpose LLMs fail to capture effectively \cite{ling2024domain}. This disparity between the broad linguistic knowledge encoded in pre-trained models and the specialized knowledge requirements of specific fields highlights the need for adapting LLMs to domain-specific applications.

\noindent\textbf{Domain-Specific Adaptation in Language Models}\\
In NLP literature, two primary strategies have emerged for developing domain-specific language models: \textbf{training a foundation model from scratch} and \textbf{continual pretraining} on domain-specific corpora.

Many research efforts have focused on building domain-specialized language models from the ground up, leveraging industry-specific knowledge to enhance their capabilities in targeted applications. For instance, BloombergGPT \cite{wu2023bloomberggpt} was trained on an extensive corpus of financial texts, improving performance in economic forecasting, market analysis, and financial sentiment detection. Similarly, in the biomedical field, BioGPT \cite{Luo2022BioGPTGP} was developed using biomedical literature, enabling enhanced capabilities in biomedical text generation and mining. Likewise, BioMedLM \cite{bolton2024biomedlm}, trained on PubMed and other medical literature, was designed to advance medical language understanding and question answering. Meanwhile, Galactica \cite{taylor2022galactica} was developed to store, integrate, and reason over scientific knowledge, making it particularly valuable for researchers and scientists working in interdisciplinary fields.

Training a foundation model from scratch requires learning from more than 100 billion tokens \cite{hoffmann2022training, wu2023bloomberggpt, taylor2022galactica} of raw domain-specific data, necessitating the meticulous curation of large-scale domain datasets. These models undergo self-supervised learning on unlabeled text \cite{Radford2019LanguageMA}, leveraging techniques such as next-token prediction \cite{openai2023gpt4} and masked language modeling \cite{Devlin2019BERTPO}, where the model sequentially predicts words while reconstructing masked tokens. While this approach maximizes domain adaptation, it is highly resource-intensive, demanding substantial computational power, large-scale datasets, and extensive infrastructure \cite{cottier2024rising}. Consequently, despite its advantages, training domain-specific models from scratch remains impractical for many specialized fields due to the immense costs and data requirements \cite{cheng2024instruction}.

An alternative approach to training domain-specific models from scratch is domain-adaptive continual pretraining (DACP) \cite{gururangan2020dont}, also known as continual pretraining (CP). This method involves further pretraining an existing foundation model on additional domain-specific unlabeled datasets \cite{gururangan2020dont, ke2023continual, ibrahim2024simple} using self-supervised learning techniques. By extending the pretraining phase of a general-purpose LLM, such as GPT, LLaMA, or other transformer-based models, on a curated domain-specific corpus, the model gradually adapts to specialized terminologies, contextual nuances, and industry-specific patterns while retaining its foundational linguistic knowledge. Domain-adaptive pretraining offers a computational advantage over training a model from scratch, as it leverages an already well-trained base model, reducing the need for massive datasets and extensive compute resources while still achieving significant domain adaptation. Studies have shown that this method significantly improves performance on specialized domains without requiring complete retraining \cite{ke2023continual, jin2022lifelong, wu2022pretrained} and also facilitates the continuous integration of emerging domain knowledge as new data becomes available \cite{jang2022towards, jang2022temporalwiki}.  

%%%%%%%%%%%%%%%%%%%%%%%%%%%%%%%%%%%%%%%%%%%%%%%%%%%%%%%%%%%%%%%%%%%%%%%%%%%%%%%%%%%%%%%%%%%%%%%%%%%%%%%%%%%%%%%%%%%%%%%%%%%%%%%%%%%%
\subsection{SFT for Domain Adaptation of LLMs}
Supervised Fine-Tuning (SFT) has emerged as a practical and resource-efficient approach to adapt large language models for use in specialized domains. Unlike pretraining, which requires massive amounts of unlabeled data and extensive compute power, SFT focuses on refining a pretrained model using a smaller, carefully curated dataset of high-quality input-output pairs. These pairs are crafted to reflect the specific language, concepts, and reasoning patterns of a target field, allowing the model to better align with domain-specific tasks and knowledge requirements \cite{pareja2024unveiling}.

SFT typically begins by initializing from a general-purpose or continually pretrained checkpoint. The model is then fine-tuned using supervised examples tailored to the domain. These examples may include question-answer pairs, document summaries, or instruction-style prompts, each chosen based on the nature of the downstream application \cite{chen2024self}. By exposing the model to structured, domain-relevant supervision, SFT enhances the model's ability to understand and generate content that is accurate, coherent, and task-aligned.

One of the key advantages of SFT is its efficiency with data. Because it builds on top of existing pretrained models, the amount of labeled data required for meaningful adaptation is relatively modest than what is needed for pretraining. This is particularly valuable in fields where labeled data is limited or expensive to obtain, especially when expert annotation is needed. SFT also supports an iterative refinement process: the model can be updated over time as new data becomes available or as task requirements evolve, without needing to restart the training pipeline from scratch. In practice, SFT has been shown to improve factual reliability, reduce hallucinations, and offer greater control over model outputs \cite{mecklenburg2024injecting}.

\subsection{Other Cost Efficient Methods for Domain Adaptation of LLMs}
In recent years, several alternative strategies for domain adaptation have emerged as cost-efficient, scalable, and performance-preserving solutions to the challenges of full fine-tuning using SFT. Parameter Efficient Fine-Tuning (PEFT) techniques such as Low-Rank Adaptation (LoRA) \cite{hu2021lora}, Quantized LoRA (QLoRA) \cite{dettmers2023qlora}, and adapter-based methods \cite{houlsby2019parameter} modify pre-trained models by introducing lightweight, trainable components. These methods substantially reduce computational and memory requirements, making them well-suited for use in resource-constrained environments without significantly compromising performance for specific tasks or domains.

However, with these reduced computational footprint advantage come notable limitations that may restrict the effectiveness of those methods in fully specializing LLMs. For example, instead of updating all the weights of the pre-trained model, LoRA and QLoRA inject low-rank trainable matrices into specific layers, and only these additional parameters are optimized during fine-tuning, leaving the original model weights frozen \cite{hu2021lora}.

Adapter-based methods follow a similar philosophy, inserting small bottleneck modules (e.g., feed-forward neural networks) between transformer layers to enable task-specific learning with minimal parameter updates. While this dramatically improve efficiency, these methods may fall short in capturing complex domain-specific representations, particularly in tasks requiring deep semantic understanding or multi-step reasoning. The frozen nature of the base model can become a bottleneck when substantial adaptation is needed. Furthermore, in the case of QLoRA, the use of aggressive quantization, typically to 4-bit precision, enhances memory savings but can also introduce numerical instability, potentially degrading performance on tasks sensitive to precision, such as arithmetic or structured reasoning \cite{li2025quantization_meets_reasoning}.

\subsection{Other Strategies for Knowledge-Intensive NLP}
 Similarly, Retrieval-Augmented Generation (RAG) \cite{lewis2020retrieval}, which augments models with external knowledge at inference time, relies heavily on the quality and relevance of the retrieval corpus. Its performance is highly sensitive to retrieval parameters such as the number of retrieved chunks and the accuracy of the ranking algorithm, poor retrieval can introduce irrelevant or conflicting context \cite{barnett2024seven}. Moreover, external lookups at inference time adds infrastructure complexity, inference latency, and increased vulnerability to retrieval noise, especially in dynamic or low-resource domains \cite{chan2025dont_do_rag}.
%%%%%%%%%%%%%%%%%%%%%%%%%%%%%%%%%%%%%%%%%%%%%%%%%%%
\section{Data Preparation: Methodology}
\label{sec:3}
A critical step in developing a domain-specific LLM is the curation of a high-quality, domain-focused dataset. This section presents the methodology used to develop an energy-specific dataset for training EnergyGPT, including data collection, curation, preprocessing, and the strategy for generating input-output pairs.

\subsection{Data Collection}
Our training corpus is constructed by combining publicly accessible energy-related literature, including scientific research papers from ASME, along with a filtered subset of The Pile dataset \cite{gao2020pile} that is relevant to the energy sector. The Pile is an open-source \cite{eleutherAI2026PileDedup} diverse English text corpus sourced from 22 high-quality domains, spanning academic, technical, legal, and online communities. Further details on journal selection criteria and licensing are provided in Appendix~\ref{appendix:DataCollection}.

\subsection{Data Cleaning: Energy-related Scientific Literature}
For the scientific literature, we extracted textual data and mathematical equations from around 40k scientific papers spanning various energy-related disciplines. We applied a preprocessing pipeline to improve quality and consistency: we removed non-content metadata (e.g., publisher/submission information, DOI links, and acknowledgments), stripped inline citation markers to avoid disrupting readability, normalized Unicode to standardize encoding across sources, corrected common OCR artifacts, and repaired broken LaTeX in equations to preserve mathematical structure. These steps were crucial to reduce noise and help the model learn from clean, domain-relevant content.

\subsection{Data Curation: The Pile}
\begin{figure}[ht]
    \centering
    \includegraphics[width=0.5\textwidth]{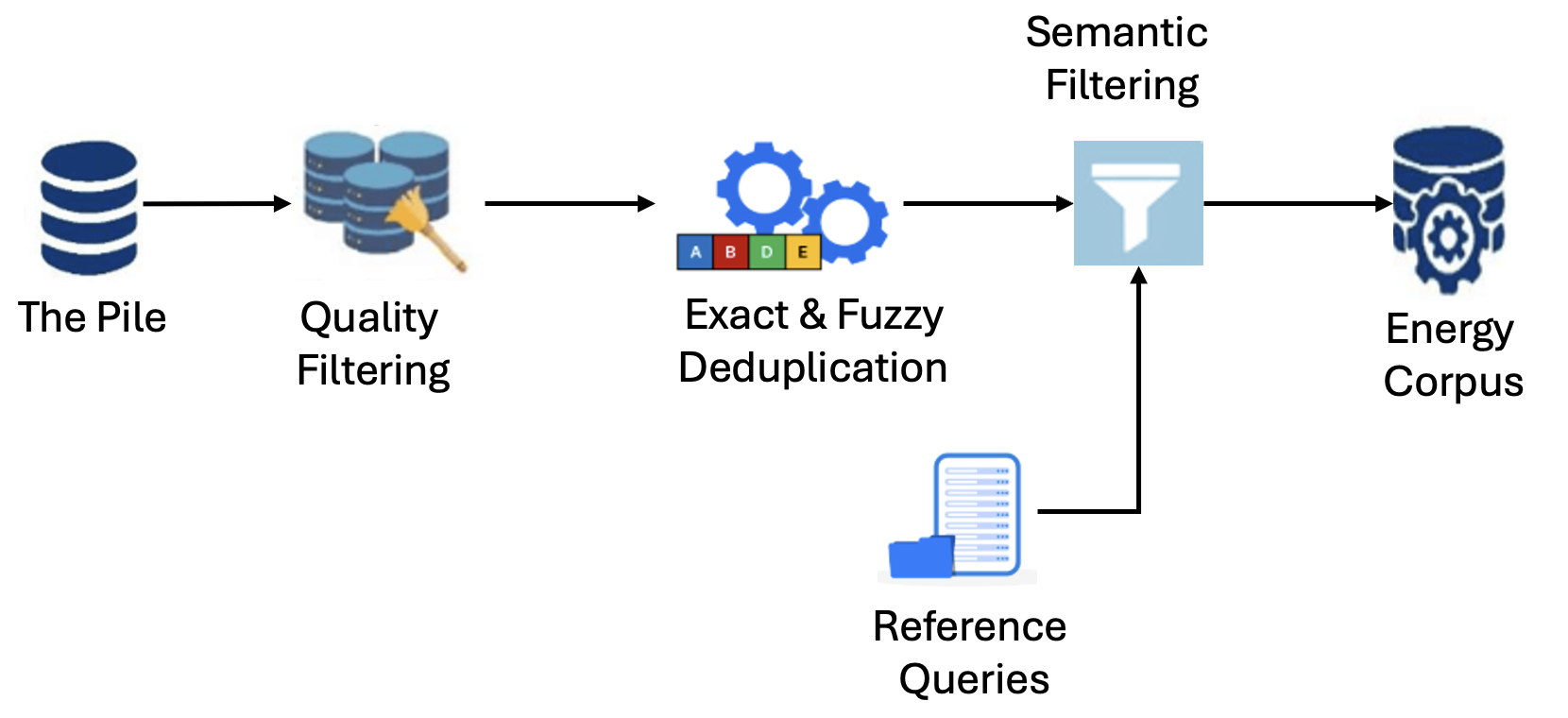}
    \caption{The Pile data processing pipeline.}
    \label{fig:thepilediagram}
\end{figure}

The following subsections describe the methodology used to extract relevant subsets from The Pile for training EnergyGPT. For this step, we used \textbf{NVIDIA NeMo Curator} \cite{nemo_curator, nvidia2026curatorDocs}, which is an open-source data processing tool \cite{nvidia2026curator} designed for large-scale corpus filtering and transformation provided by NVIDIA. This process involves applying rigorous quality filtering, deduplications, and ifentifying energy-related documents from The Pile (see Figure ~\ref{fig:thepilediagram}).

\subsubsection{Quality Filtering}
A significant portion of raw web text is unsuitable for training language models, as it often contains broken HTML, non-printable characters, or incoherent content. Training on such noisy data can degrade model performance. In the GPT-3 data pipeline \cite{brown2020language}, this was mitigated via quality filtering using a logistic regression classifier to distinguish high-quality text from gibberish based on low-dimensional document representations \cite{hashingtf}. In our work, we used a similar classifier, Quality Classifier DeBERTa \cite{nvidia_quality_filtering, nvidia_quality_classifier} to label documents as high, medium, or low quality based on textual coherence and structure, retaining only high and medium quality documents for subsequent processing (approximately 107 million documents).

\subsubsection{Exact De-duplication}
Large-scale text corpora sourced from the web are known to contain a substantial amount of duplicated content. These duplicates often stem from repeated web pages, mirrored sites, or boilerplate text structures. For instance, \cite{penedo2023refinedweb} reports that 12\% of Common Crawl snapshots are composed of duplicate documents, with a significant portion, up to ~40\%, being exact duplicates. Training on such redundant datasets across multiple epochs can be suboptimal, leading to inefficiencies in both computational resource usage and generalization performance \cite{tirumala2023d4, muennighoff2023scaling}. Deduplication is therefore a critical preprocessing step. Prior work such as \cite{lee2022deduplicating} proposed algorithms for detecting exact duplicates based on suffix tree representations, which enable the identification of repeated substrings in linear time. While effective, such methods may introduce substantial computational overhead when applied to extremely large datasets (e.g. the Pile).

In our preprocessing pipeline, we adopt a more scalable and computationally lightweight approach based on hashing techniques. Each document is processed to generate a unique hash signature, and only one instance is retained per unique hash, ensuring that exact duplicates are efficiently removed from the corpus before further processing \cite{nvidia_deduplication}.

\subsubsection{Fuzzy De-duplication}
While exact deduplication effectively removes identical documents, it fails to capture near-duplicate documents that share significant textual overlap but are not byte-for-byte identical. Such near-duplicates often result from minor edits, formatting differences, or slight paraphrasing \cite{smith2022megatron}. To address this limitation, we employed fuzzy deduplication techniques within NVIDIA NeMo Curator\cite{nvidia_deduplication} to detect and eliminate near-duplicate documents, which are particularly prevalent in large-scale corpora like The Pile. A detailed description of our fuzzy deduplication pipeline, settings, and Algorithm ~\ref{alg:fuzzy_deduplication} is provided in Appendix~\ref{appendix:Deduplication}. At the end of this stage, we retained approximately 94 million documents.
% documents 94,044,890

\subsubsection{Semantic Filtering}
Since our source corpora contain a broad mix of content spanning numerous domains, performing semantic filtering is essential to isolate energy-related data \cite{abbas2023semdedup}. 
To achieve this, we constructed a set of expertly curated reference queries representing various topics within the energy domain (see Appendix~\ref{appendix:topics}). We then leveraged transformer-based embedding models from the Sentence Transformers library \cite{reimers2019sentencebert} to encode both the reference queries and candidate documents. Specifically, we experimented with several state-of-the-art models, including \texttt{intfloat/e5-large-v2} \cite{intfloat_e5_large_v2}, \texttt{BAAI/bge-base-en-v1.5} \cite{baai_bge_base_en_v1.5}, and \texttt{\seqsplit{sentence-transformers/all-mpnet-base-v2}} \cite{sentence_transformers_all_mpnet_base_v2}, to evaluate their effectiveness in capturing semantic similarity. We run the filtering setp on multiple GPUs by sharding the corpus, computing normalized document embeddings, and retaining documents whose maximum cosine similarity to any reference query exceeds 0.8. This yields approximately 133 thousand energy-relevant documents from The Pile.

\subsection{Data Balancing}
\begin{figure}[!htbp]
    \centering
    \includegraphics[width=0.5\textwidth]{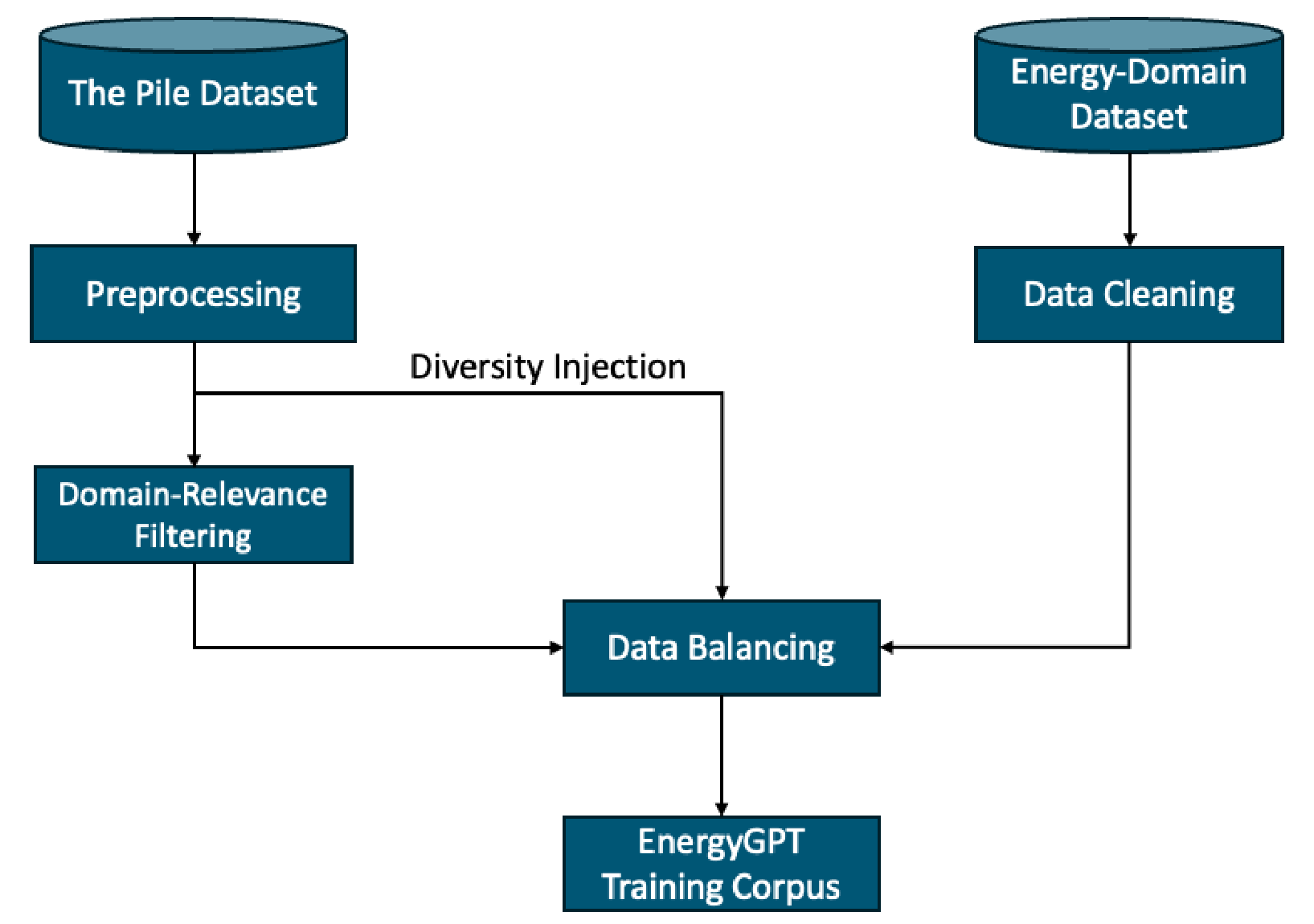}
    \caption{Data preparation pipeline for fine-tuning EnergyGPT.}
    \label{fig:datapipeline}
\end{figure}

The final corpus used to train EnergyGPT was constructed by mixing two primary sources to create an energy-specialized dataset: (1) energy-related text extracted from scientific papers published in reputable journals, and (2) a filtered, semantically relevant subset of The Pile dataset. This mixture ensures a balance between highly specialized technical content and broader contextual information within the energy domain. Following best practices recommended in \cite{zhang2023balancing, luo2025catastrophic}, we included a small proportion of randomly selected, high-quality general-domain content from The Pile. This deliberate injection of general data, approximately 1.4\% of the total corpus, helps preserve the model's general language capabilities and mitigate catastrophic forgetting. The data preparation process is illustrated in Figure~\ref{fig:datapipeline}. \\

The table ~\ref{tab:datacomposition} below summarizes the composition of the final dataset used for fine-tuning EnergyGPT.
\begin{table}[!htbp]
\centering
\begin{tabular}{lll}
\toprule
\textbf{Dataset} & \textbf{Quantity (tokens)} & \textbf{Weight in training mix} \\
\midrule
Scientific Papers & \textasciitilde 1.8 Billion & 82.9\% \\
The Pile (relevant)  & \textasciitilde  0.34 Billion  & 15.7\% \\
The Pile (filtered)                &  \textasciitilde 30 Million  & 1.4\% \\
\bottomrule
\end{tabular}
\vspace{10pt}
\caption{\textbf{Datasets used for fine-tuning EnergyGPT}. “Weight in training mix” refers to the fraction of examples during training that are drawn from a given dataset.}
\label{tab:datacomposition}
\end{table}
% nb tokens from asme papers 914.986.247
\subsection{Data Pairs Generation for Supervised Fine-Tuning}
Fine-tuning a large language model using supervised fine-tuning requires the training data to be organized into input-output pairs. To support this, we developed an efficient pipeline that transforms each document into structured data pairs. For data extracted from The Pile, each document was first split into sentence-aware and equation-aware chunks, ensuring that sentence boundaries were preserved and that mathematical equations remained intact and not broken apart (i.e., not fragmented across chunks).  We limited each chunk to around 600 tokens to stay within the model's capacity and to keep the content manageable. We used LLaMA 3.1-8B tokenizer throughout this process. From the resulting chunks, we generated overlapping input-output pairs with a stride of $C_{n+2}$, allowing contextual information to flow smoothly across sequences. This approach helped maintain consistency in structure and token length across pairs (see Appendixs~\ref{appendix:training_pairs}).
The pairing strategy we followed is illustrated below:
\begin{center}
\begin{tabular}{rl}
\textbf{Pair 1:} & Input $(C_1\ C_2\ C_3)$ $\longmapsto$ \textit{Output} $(C_4\ C_5\ C_6)$ \\
\textbf{Pair 2:} & Input $(C_3\ C_4\ C_5)$ $\longmapsto$ \textit{Output} $(C_6\ C_7\ C_8)$ \\
\end{tabular}
\end{center}
where $C_i$ denotes the $i^{th}$ chunk extracted from a document $d$.

For data extracted from the scientific literature, each document is first segmented into paragraph-aware, sentence-aware, and equations-aware chunks. These segments are then organized into input-output pairs using a sliding window strategy with a stride of one paragraph. Each pair is constrained to a maximum of 4,096 tokens, with an effort made to balance input and output lengths, typically targeting around 2,000 tokens per side, to prevent the model from overfitting on short output. Thus the number of chunks included in the input and output is variable, depending on the token lengths of paragraphs (see Appendix ~\ref{appendix:training_pairs}). An example of the pairing strategy is illustrated below:
\begin{center}
\begin{tabular}{rl}
\textbf{Pair 1:} & Input $(P_1\ P_2\ P_3)$ $\longmapsto$ \textit{Output} $(P_4\ P_5)$ \\
\textbf{Pair 2:} & Input $(P_3\ P_4)$ $\longmapsto$ \textit{Output} $(P_5\ P_6\ P_7)$ \\
\end{tabular}
\end{center}
where $P_i$ denotes the $i^{th}$ paragraph extracted from a document $d$.\\
The total number of pairs generated and their sources are summarized in Table~\ref{tab:data-pairs}.
\begin{table}[!htbp]
\centering
\begin{tabular}{lll}
\toprule
\textbf{Dataset} & \textbf{Number of pairs}  \\
\midrule
Scientific Papers &  517,716 pairs \\
The Pile (relevant) & 193,448 pairs \\
The Pile (filtered) & 20,781 pairs  \\
\bottomrule
\end{tabular}
\vspace{10pt}
\caption{Data pairs used for fine-tuning EnergyGPT.}
\label{tab:data-pairs}
\end{table}

Note that all data pairs used for training EnergyGPT are constructed directly from these corpora using the pairing strategies described above; we do not incorporate any additional generic or synthetic datasets.

%%%%%%%%%%%%%%%%%%%%%%%%%%%%%%%%%%%%%%%%%%%%%%%%%%%%%%%%%%%%%%%%%%
\section{EnergyGPT Training Approach}
\label{sec:4}
 This section describes the overall training strategy, infrastructure, and experimental setup used to develop the EnergyGPT variants (see Figure~\ref{fig:energygptpipeline}). More details can be found in Appendix~\ref{appendix:Training}.

 \begin{figure}[!htbp]
    \centering
    \includegraphics[width=0.5\textwidth]{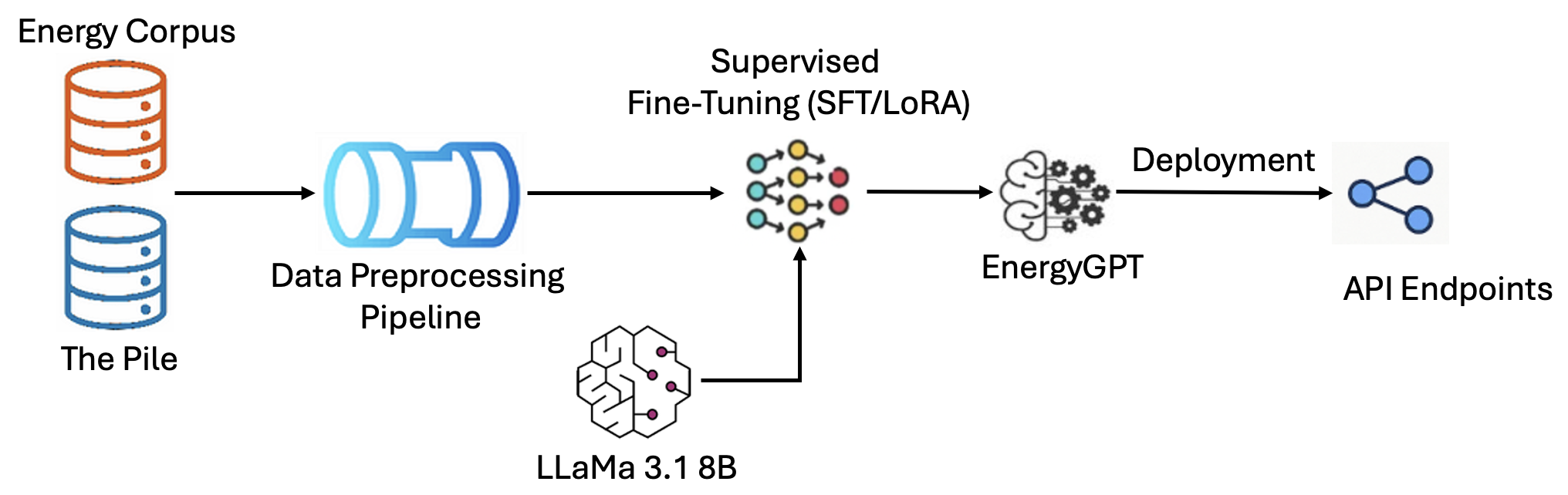}
    \caption{EnergyGPT pipeline.}
    \label{fig:energygptpipeline}
\end{figure}

\subsection{Base Model}
In our work, we adopted the open-source LLaMA 3.1-8B \cite{metaLlama3_1_8B} as the foundation model of EnergyGPT. The architectural specifications of LLaMA 3.1-8B are presented in Table~\ref{tab:llama3_8b_architecture}.
\begin{table}[!htbp]
\centering
\begin{tabular}{lc}
\toprule
\textbf{Parameter} & \textbf{Configuration} \\
\midrule
Hidden size & 4096 \\
Heads & 32 \\
Layers & 32 \\
Vocab size & 128,256 \\
Sequence length & 131,072 \\
\bottomrule
\end{tabular}
\vspace{10pt}
\caption{Architectural specifications of the LLaMA 3.1-8B model used as the base for EnergyGPT.}
\label{tab:llama3_8b_architecture}
\end{table}

Note that in our experiment, we retain the original LLaMA 3.1-8B tokenizer and input/output embedding layers without any vocabulary modifications.

\subsection{SFT Training Configuration and Hardware}
For the fine-tuning of EnergyGPT, we leveraged NVIDIA's high-performance hardware and software stack. The training was conducted on a compute cluster equipped with 4 NVIDIA A100 GPUs (80GB each), using the \textbf{NVIDIA NeMo} framework and \textbf{Megatron-LM} toolkit \cite{shoeybi2019megatron, narayanan2021efficient}, \cite{nvidia2026nemoSFTPEFT, nvidia2026NeMo, nvidia2026megatronBridgeDocs}. The training run was configured for up to 2 epochs over the curated dataset, with early stopping based on validation loss. To maximize resource utilization while avoiding out-of-memory errors, we employed intra-layer model parallelism, which distributes matrix operations within each transformer layer across multiple GPUs \cite{smith2022megatron}.\\
Throughout training, we regularly monitored progress and evaluated performance by calculating the validation loss (see Appendix~\ref{appendix:Training}). The 10 best-performing checkpoints were saved at the end for subsequent selection. A detailed summary of the training configuration and hyperparameters is provided in Table~\ref{tab:instruction-synthesizer-hparams}.

\begin{table}[!htbp]
\centering
\begin{tabular}{ll}
\toprule
\textbf{Hyper-Parameter} & \textbf{Configuration} \\
\midrule
Base model & LLaMA 3.1-8B \\
Compute infrastructure & 4 × A100-80GB GPUs \\
Training duration & $\approx$ 6 days \\
Epochs & 2 \\
Global batch size & 256 \\
Micro batch size & 8 \\
Max sequence length & 8192 \\
Max learning rate & 2e-5 \\
Optimizer & Distributed Fused Adam \\
Adam beta weights & [0.9, 0.95] \\
Learning rate scheduler & Cosine Annealing \\
Weight decay & 0.01 \\
Warm-up steps & 280 \\
Gradient clipping & 1.0 \\
\bottomrule
\end{tabular}
\vspace{10pt}
\caption{Hyperparameters for fine-tuning EnergyGPT using SFT.}
\label{tab:instruction-synthesizer-hparams}
\end{table}
\vspace{0.5\baselineskip}
\textit{Note: While the LLaMA 3.1-8B model architecture supports sequences up to 131,072 tokens, our fine-tuning was conducted using a maximum sequence length of 8,192 tokens due to hardware constraints. Most training examples, however, were truncated to 4,096 tokens or fewer to match the average size of domain-specific documents. This conservative input length improved batch efficiency and enabled stable training of the 8B model on a modest 4 × A100 GPU setup.}

In terms of size, the SFT variant of EnergyGPT preserves the full architecture and parameter count of the original LLaMA~3.1-8B model (8B parameters); no structural changes are introduced. In our experiments, the SFT model is stored both as a NeMo \texttt{.nemo} archive and as an equivalent Hugging Face checkpoint for downstream deployment.

\subsection{LoRA Training Configuration and Hardware}

In addition to full-parameter tuning with SFT, we train a parameter-efficient variant of EnergyGPT using Low-Rank Adaptation (LoRA)~\cite{hu2021lora} on the same base model and dataset. Specifically, we attach LoRA adapters of rank $r=32$ (with scaling factor $\alpha=32$) to the attention query-key-value (attention\_qkv) projections in all transformer layers of LLaMA 3.1-8B, while keeping all original model weights frozen. This results in approximately 10.5M trainable parameters, i.e., about $0.13\%$ of the 8B-parameter base model.

As with SFT, we use NVIDIA NeMo’s PEFT stack to fine-tune the LoRA adapters on the same GPU cluster (4~$\times$~A100-80GB), with tensor model parallelism of 4 and pipeline model parallelism of 1. A detailed summary of the training configuration and hyperparameters is provided in Table~\ref{tab:energygpt-lora-hparams}.

\begin{table}[!htbp]
\centering
\begin{tabular}{ll}
\toprule
\textbf{Hyper-Parameter} & \textbf{Configuration} \\
\midrule
Base model & LLaMA 3.1-8B (frozen) \\
PEFT method & LoRA (rank $r = 32$) \\
Target modules & Attention QKV projections \\
Trainable parameters & $\approx 10.5$M ($\sim$0.13\% of 8B) \\
Compute infrastructure & 4 $\times$ A100-80GB GPUs \\
Global batch size & 256 \\
Micro batch size & 8 \\
Max sequence length & 8192 \\
Epochs & 2  \\
Optimizer & Fused Adam \\
Learning rate & 1e-4 \\
Adam beta weights & [0.9, 0.98] \\
Learning rate scheduler & Cosine Annealing \\
Warm-up steps & 280 \\
Weight decay & 0.01 \\
Gradient clipping & 1.0 \\
\bottomrule
\end{tabular}
\vspace{10pt}
\caption{Hyperparameters for fine-tuning EnergyGPT using LoRA.}
\label{tab:energygpt-lora-hparams}
\end{table}

The resulting LoRA variant therefore consists of the frozen 8B-parameter LLaMA~3.1-8B base model plus approximately 10.5M trainable adapter parameters. In our setup, these adapters are stored as separate NeMo PEFT checkpoints that are merged with the base weights at inference time.

%%%%%%%%%%%%%%%%%%%%%%%%%%%%%%%%%%%%%%%%%%%%%%%%%%%%%%%%%%%%%%%%%%%%%%%%%%%%%%%%%%%%%%%%%%%%%%%%%%%%%%%%%%%%%%%%%%%%%%%%%%%%%%%%%%%%
\section{EnergyGPT Benchmark}
\label{sec:5}
This section introduces the composition of the benchmark used to evaluate EnergyGPT, along with the methodology for selecting a high-quality LLM to serve as an automated judge for grading and evaluation of the models responses on open-ended queries.
\subsection{Benchmark Design}
 We developed a specialized benchmark for evaluating EnergyGPT, consisting of 476 question-answer pairs designed to assess the model's performance across a range of cognitive tasks related to energy science. The benchmark includes multiple-choice, true/false, factual, and open-ended questions categorized by difficulty levels: easy, medium, hard, and challenging. Questions were either manually authored or adapted from public-domain sources. These questions are divided into three main sets, each serving a distinct purpose:
(1) 100 true/false statements focused on evaluating the model's grasp of fundamental energy concepts and its ability to handle binary factual reasoning;
(2) 233 multiple-choice questions aimed at testing the model's capability to recognize correct information among distractors and to perform basic reasoning under constrained options;
(3) 143 open-ended energy-related questions that vary in difficulty and include factual recall, designed to assess the model's depth of understanding, ability to generate informative-coherent answers, and performance on complex, open-form queries. To illustrate the structure and content of the benchmark, a sample of questions are provided in the supplementary text boxes in Appendix~\ref{appendix:Samples}.
\begin{table}[!htbp]
\centering
\scriptsize
\begin{tabular}{p{1.8cm}p{0.9cm}p{2.8cm}p{2cm}}
\toprule
\textbf{Type} & \textbf{Count} & \textbf{Focus} & \textbf{Difficulty} \\
\midrule
True/False & 100 & Factual validation, binary response & Easy-Hard \\
Multiple Choice & 233 & Distractor recognition and basic reasoning & Easy-Hard \\
Open-Ended & 143 & Free-form generation  & Easy-Challenging \\
\bottomrule
\end{tabular}
\caption{EnergyGPT Benchmark by Question Type, Evaluation Focus, and Difficulty.}
\label{tab:benchmark-overview}
\end{table}

\textit{Note: The benchmark was constructed independently of the training data and was excluded from model training.}

In our benchmarking strategy, we evaluated the model's ability to select the correct answer by employing a structured prompting approach, which constrained the model's response to a single word or letter, ideal for multiple-choice or binary (True/False) questions. Two distinct prompt templates were used to instruct the model: \\
\textbf{1. Prompt for Multiple-choice questions:}
\begin{quote}
\begin{ttfamily}
You are an expert in the energy field answering multiple-choice questions.

Please provide the correct answer key for the following question, you do not need to provide an explanation:

<Q>

A. <option A>

B. <option B>

C. <option C>

D. <option D>
\end{ttfamily}
\end{quote}
followed by:
\begin{quote}
\begin{ttfamily}
Answer Now:
\end{ttfamily}
\end{quote}
\textbf{2. Prompt for True/False statements:}
\begin{quote}
\begin{ttfamily}
\#\#\#Instruction:

You will be given a statement from the energy domain.

Decide whether this statement is factually correct.

If it is correct, answer with "True".

If it is incorrect, answer with "False".

You may provide a brief explanation.

Statement: \{q\}
\end{ttfamily}
\end{quote}
followed by:
\begin{quote}
\begin{ttfamily}
\#\#\#Response:
\end{ttfamily}
\end{quote}
The benchmarking was conducted with a deterministic setup without sampling (that is, T = 0) to ensure consistency in responses.

%%%%%%%%%%%%%%%%%%%%%%%%%%%%%%%%%%%%%%%%%%%%%%%%%%%%%%
\subsection{Choice of LLM Judge}
\label{subsec:5}
Evaluating the quality of generated responses in specialized domains like energy is both essential and resource-intensive. Human expert evaluation remains the gold standard but is time-consuming, costly, and difficult to scale across large datasets. As the diversity and volume of generated outputs increase, there is a growing need for automated, scalable, and reliable evaluation frameworks that still approximate human judgment.

Recent work has popularized the LLM-as-a-judge method as a scalable alternative to purely human evaluation for text generation systems~\cite{zheng2023judging, bai2024mtbench101, hada2023multilingualevaluators, shen2023llmabssum, llmasjudge_survey}. These studies show that strong, well-aligned LLMs such as GPT-4 can reach over 80\% agreement with human annotators, but also highlight important limitations: judge models may be sensitive to prompt design, exhibit style or length biases, and sometimes diverge from expert judgments on domain-heavy and safety-critical tasks~\cite{shen2023llmabssum, feuer2024style_bias, empirical_llm_judges}. As a result, LLM judges are increasingly treated as noisy but useful proxies rather than ground truth. 

In our setting, we follow best practices from this line of work by (i) calibrating candidate judges against human expert scores on a seed set, (ii) using deterministic decoding and fixed rubrics to reduce variance, and (iii) interpreting the resulting scores primarily as relative comparisons between models rather than absolute measures of correctness.

\subsubsection{Experimental Setup}

 To identify a judge model that closely mirror human expert preferences in the energy and technical QA domain, we evaluated several state-of-the-art LLMs. We used responses generated by LLaMA 3.1-8B for 118 energy-related queries. Each response was scored along seven evaluation criteria: Relevance, Correctness, Technical Level, Scientific Level, Explainability, Conciseness, and Coherence. Each LLM judge was instructed to act as a domain expert and return a structured JSON containing integer scores (0-10) for each dimension. More details about the judge prompt used and the generation settings are provided in Appendix~\ref{appendix:llm_judge}. For ground truth, expert human annotators independently scored the same set of responses using identical criteria.

\subsubsection{Results and Analysis}

For each of the seven evaluation dimensions, we first quantify agreement between the two human experts on the 118 calibration queries. Let $H_1$ and $H_2$ denote the per-item scores assigned by the two annotators, and let $\bar{H} = (H_1 + H_2)/2$ denote their average. For each dimension, we compute Pearson's correlation coefficient $r$, Spearman's rank correlation $\rho$, mean absolute error (MAE), quadratic-weighted Cohen's kappa $\kappa_{\text{quad}}$, and Krippendorff's $\alpha$ between $H_1$ and $H_2$ to characterize inter-annotator consistency. The results in Table~\ref{tab:human_agreement} show strong agreement across all seven criteria: Pearson $r$ ranges from 0.82 to 0.89, Spearman's $\rho$ from 0.77 to 0.88, and both $\kappa_{\text{quad}}$ and Krippendorff's $\alpha$ fall in the 0.81-0.89 range, indicating high reliability of the human reference scores.

\begin{table}[!htbp]
    \centering
    \caption{Inter-annotator agreement between the human experts on the calibration queries, across the seven evaluation dimensions.}
    \label{tab:human_agreement}
    \begin{tabular}{lccccc}
        \hline
        Metric & Pearson $r$ & Spearman $\rho$ & MAE & $\kappa_{\text{quad}}$ & Kripp. $\alpha$ \\
        \hline
        Relevance        & 0.887 & 0.862 & 0.42 & 0.893 & 0.886 \\
        Correctness      & 0.835 & 0.808 & 0.66 & 0.833 & 0.833 \\
        Technical level  & 0.867 & 0.880 & 0.61 & 0.823 & 0.819 \\
        Scientific level & 0.849 & 0.862 & 0.55 & 0.828 & 0.826 \\
        Explainability   & 0.838 & 0.802 & 0.57 & 0.824 & 0.823 \\
        Conciseness      & 0.818 & 0.774 & 0.77 & 0.815 & 0.815 \\
        Coherence        & 0.840 & 0.814 & 0.67 & 0.838 & 0.838 \\
        \hline
    \end{tabular}
\end{table}

\begin{figure}[!htbp]
    \centering
    \includegraphics[width=0.5\textwidth]{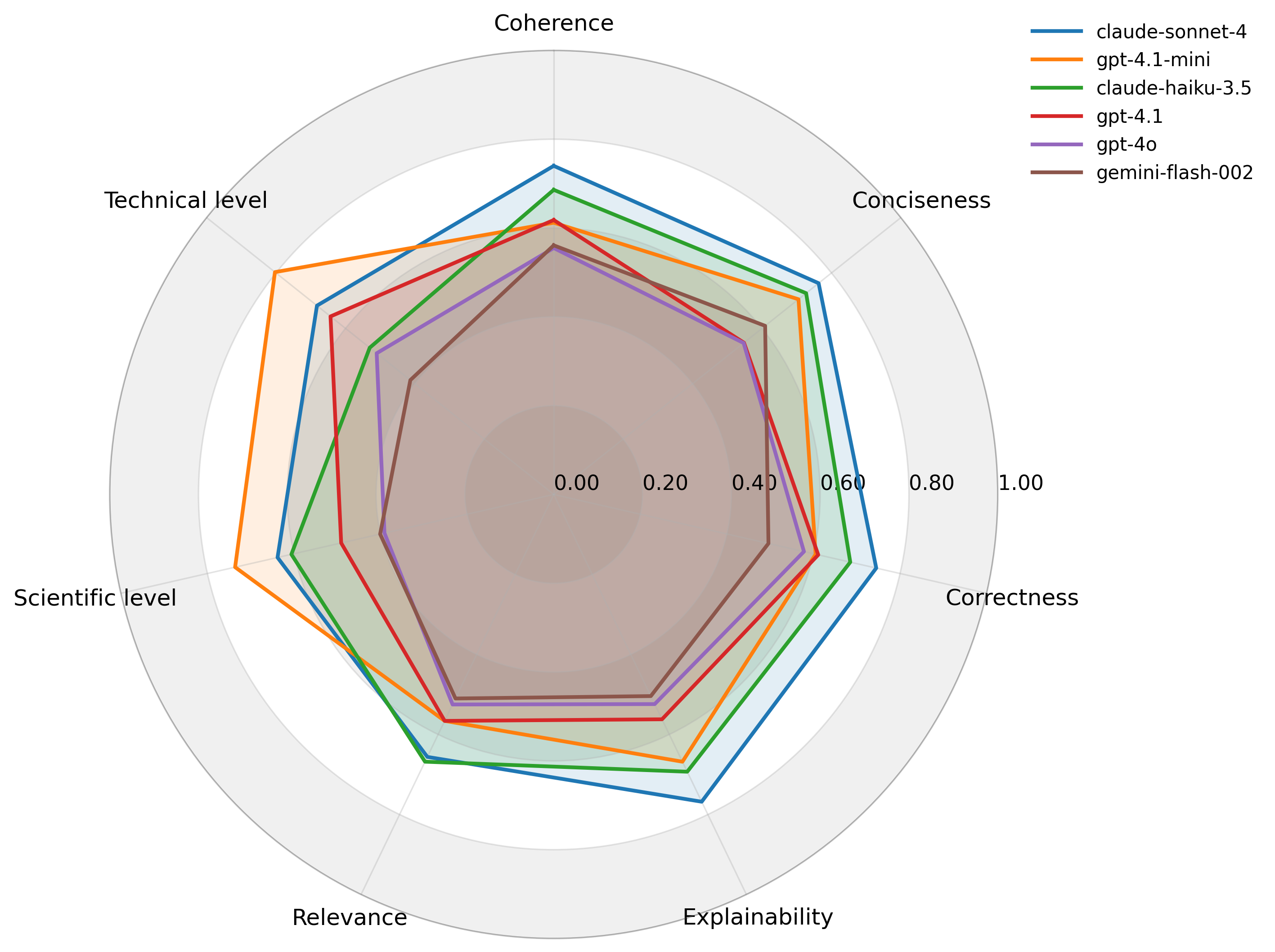}
    \caption{Cohen's kappa between selected candidate LLM judges and the mean human scores across the seven evaluation dimensions.}
    
    \label{fig:llm-judge-qwk_top6}
\end{figure}

Next, we evaluate a set of candidate LLM judges by comparing each judge's scores against the human reference $\bar{H}$ on the same 118 calibration items. We compute the same agreement statistics between each judge and $\bar{H}$ for every evaluation dimension. Table~\ref{tab:llm_judge_summary_metrics} summarizes judge-human agreement averaged across dimensions, and Figure~\ref{fig:llm-judge-qwk_top6} presents a radar plot of per-dimension agreement using $\kappa_{\text{quad}}$ for a subset of representative candidates (a corresponding plot including all evaluated models is provided in Appendix)~\ref{appendix:llm_judge}. More detailed per-dimension results for all judges are also provided in Appendix~\ref{appendix:llm_judge}.

\begin{table}[!htbp]
    \centering
    \small
    \caption{Judge-human agreement statistics (averaged across evaluation dimensions) for each candidate LLM judge.}
    \label{tab:llm_judge_summary_metrics}
    \setlength{\tabcolsep}{3pt}
    \resizebox{\columnwidth}{!}{%
    \begin{tabular}{lccccc}
        \hline
        Model &
        Pearson $r$ &
        Spearman $\rho$ &
        MAE &
        $\kappa_{\text{quad}}$ &
        Kripp.\ $\alpha$ \\
        \hline
        Claude Sonnet 4    & 0.753 & 0.649 & 1.100 & 0.713 & 0.704 \\
        Claude Haiku 3.5   & 0.727 & 0.617 & 1.206 & 0.656 & 0.641 \\
        GPT-4.1 Mini       & 0.816 & 0.750 & 1.400 & 0.671 & 0.623 \\
        GPT-4.1            & 0.801 & 0.743 & 1.768 & 0.577 & 0.499 \\
        Gemini 2.5         & 0.614 & 0.596 & 1.964 & 0.478 & 0.392 \\
        Gemini Flash-002   & 0.590 & 0.553 & 1.964 & 0.499 & 0.465 \\
        Gemini 1.5 Pro-002 & 0.589 & 0.578 & 2.031 & 0.459 & 0.369 \\
        GPT-4o             & 0.784 & 0.734 & 2.046 & 0.519 & 0.417 \\
        GPT-4o Mini        & 0.741 & 0.672 & 2.054 & 0.459 & 0.330 \\
        LLaMA 3.3 70B      & 0.730 & 0.613 & 2.231 & 0.465 & 0.322 \\
        GPT-4              & 0.699 & 0.597 & 2.855 & 0.367 & 0.135 \\
        \hline
    \end{tabular}
    }%
\end{table}

To select automated judges for downstream evaluation, we prioritize two complementary objectives: (i) calibrated agreement with human scoring, reflected by low MAE and high $\kappa_{\text{quad}}$ and Krippendorff's $\alpha$, and (ii) ranking fidelity, reflected by high Pearson $r$ and Spearman's $\rho$. As shown in Table~\ref{tab:llm_judge_summary_metrics}, Claude-Sonnet-4 provides the strongest overall calibrated agreement with human scores (lowest MAE and the highest $\kappa_{\text{quad}}$ and $\alpha$), while GPT-4.1-mini provides the strongest ranking fidelity (highest mean Pearson $r$ and Spearman's $\rho$). We therefore adopt two judges in our experiments: Claude-Sonnet-4 as the primary judge for human-aligned scoring and GPT-4.1-mini as a complementary judge emphasizing correlation and ranking consistency. Using two judges improves robustness to judge-specific biases and allows us to verify that model comparisons remain consistent across both agreement and correlation-oriented evaluators.

\subsubsection{Qualitative Analysis of Human-Judge Disagreements}

Beyond aggregate agreement statistics, we conducted a focused qualitative analysis of items with large discrepancies between human and LLM-judge scores. Using the calibration set, we examined all query-dimension pairs (118 queries $\times$ 7 dimensions = 826 pairs) where the absolute difference $|J^{(k)} - \bar{H}| \geq 3$ on the 0-10 scale, for each of the two selected judges $J^{(\text{Sonnet})}$ (Claude-Sonnet-4) and $J^{(\text{Mini})}$ (GPT-4.1-mini). Such large disagreements occur on roughly $5\%$ of query-dimension pairs for Claude-Sonnet-4 and about $11\%$ for GPT-4.1-mini, indicating that Sonnet-4 is, on average, closer to human scores. In this quantification, we focus on the mean signed differences $J^{(k)} - \bar{H}$ (rather than MAE) to characterize the direction of judge bias (over-scoring vs under-scoring). A sample of disagreement cases is provided in Table ~\ref{tab:human_judge_qual_examples} in Appendix~\ref{appendix:llm_judge_examples}.

Inspecting the highest-disagreement cases reveals two dominant patterns. First, both judges tend to be more lenient than human experts on fluent but partially incorrect answers. For questions requiring exact numerical values or recent cost figures, human annotators often assign low correctness scores when the model fabricates plausible-sounding numbers, whereas GPT-4.1-mini in particular assigns much higher scores (often 7-9). This is reflected in its positive mean signed differences on relevance, correctness, coherence, and explainability (around $+1.7$ on correctness and coherence on average), indicating a tendency to reward surface plausibility and narrative quality even when factual content is weak. Claude-Sonnet-4 exhibits a similar but milder trend, with a moderate positive bias on relevance (about $+0.6$ on average) and a small positive bias on coherence (about $+0.2$), while remaining essentially unbiased on conciseness (around $-0.1$).

Second, we observe cases where answers are conceptually correct but lack technical or scientific detail. In these situations, Claude-Sonnet-4 is slightly harsher than humans: it assigns lower scores on technical level and scientific level (average signed differences of roughly $-0.8$ and $-0.4$), penalizing missing physical justification and incomplete engineering arguments more than the annotators do. GPT-4.1-mini, by contrast, tends to be more forgiving in such cases, assigning higher explainability and conciseness scores (on the order of $+1.1$ on average) as long as the response is clear and well-structured.

Overall, these patterns support the quantitative findings: both judges broadly track expert preferences, but GPT-4.1-mini systematically overestimates the quality of fluent yet imperfect answers, whereas Claude-Sonnet-4 stays closer to human ratings and is comparatively more sensitive to missing technical and scientific depth. We therefore treat judge-based scores as calibrated but imperfect proxies for expert evaluation: human ratings remain the primary ground truth, and we use two complementary judges only to provide secondary, bias-mitigated support.

%%%%%%%%%%%%%%%%%%%%%%%%%%%%%%%%%%%%%%%%%%%%%%%%%%%%%%%%%%%%
\section{Results and Achievements}
\label{sec:6}
\begin{figure}[!htbp]
  \centering
  \begin{subfigure}{0.49\linewidth}
    \centering
    \includegraphics[width=\linewidth]{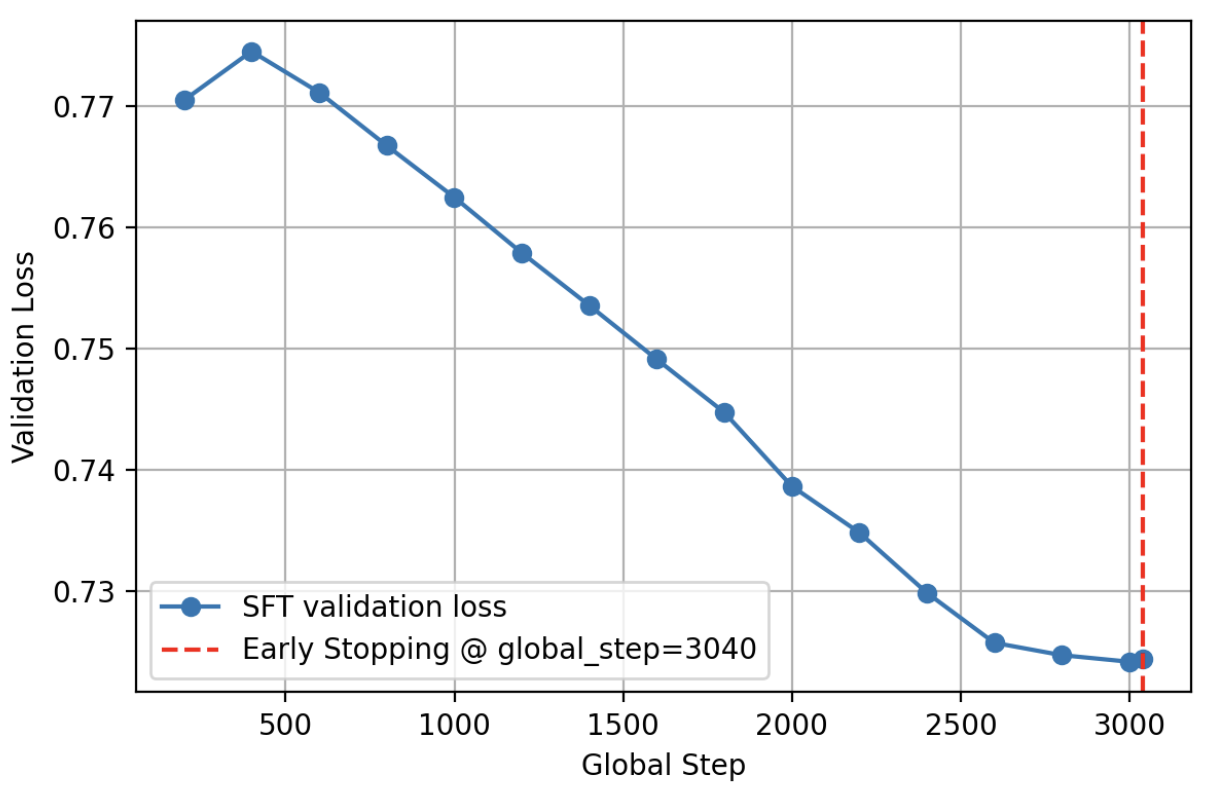}
    \caption{SFT fine-tuning}
    \label{fig:sft_val_loss}
  \end{subfigure}
  \hfill
  \begin{subfigure}{0.49\linewidth}
    \centering
    \includegraphics[width=\linewidth]{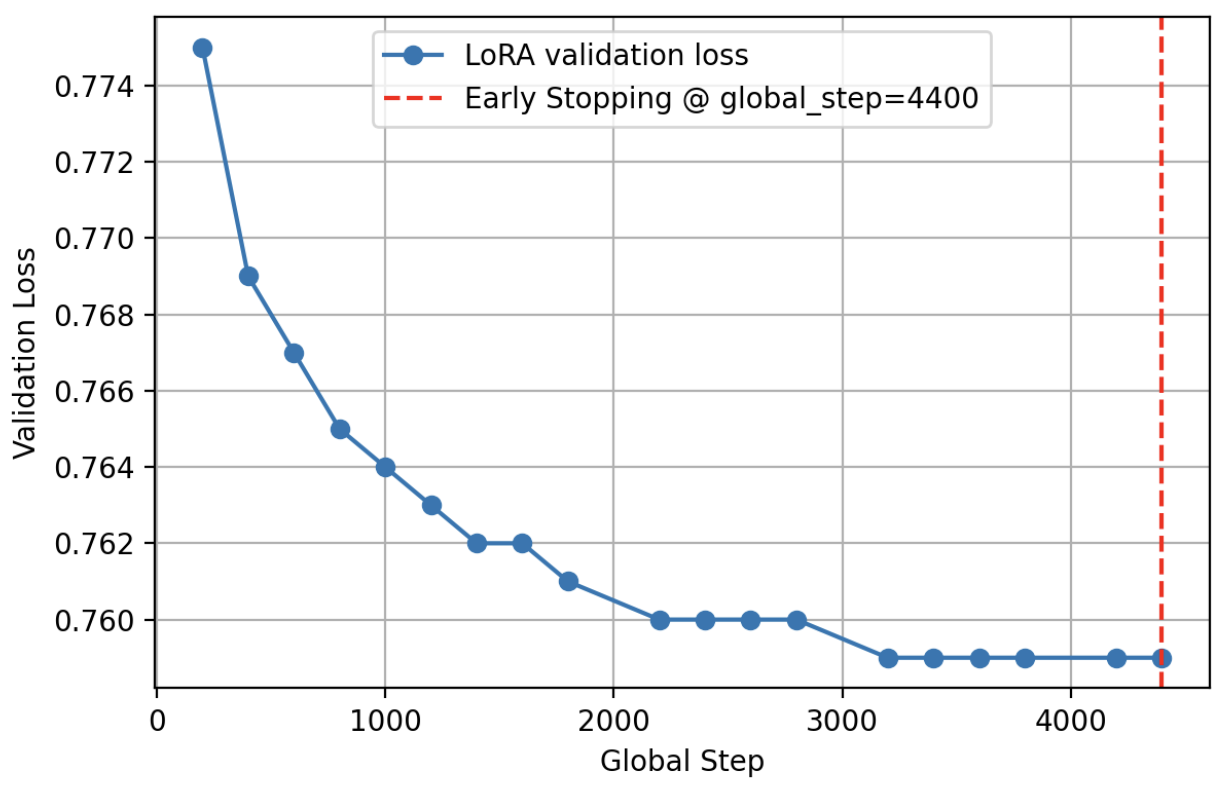}
    \caption{LoRA fine-tuning}
    \label{fig:lora_val_loss}
  \end{subfigure}
  \caption{Validation loss vs. global step for SFT (a) and LoRA (b) on LLaMA 3.1-8B. The red dashed line marks the checkpoint selected for downstream evaluation in each setting.}
  \label{fig:val-loss-sft-lora}
\end{figure}

To illustrate how model performance evolves during fine-tuning, Figure~\ref{fig:val-loss-sft-lora} reports the cross-entropy validation loss as a function of global training step for both settings. In all runs, validation loss is evaluated every 200 steps on a held-out set comprising 2\% of the corpus (approximately 15,000 input-output pairs). In the full SFT setting (Figure~\ref{fig:val-loss-sft-lora}\,(a)), we observe a small increase in loss at the beginning, followed by a steady decline as optimization progresses. Training is stopped early once the validation loss plateaus for three consecutive evaluations and then begins to increase while the training loss continues to decrease. The full fine-tuning process spanned 3,040 steps  and was completed in approximately 6 days. The corresponding checkpoint, indicated by the red dashed line, is the one with the minimum validation loss among the saved checkpoints and is used for all downstream evaluations ($L_{\text{val}} = 0.72$). The drop in cross-entropy in SFT translates into a reduction in perplexity roughly by 5\%.

The LoRA run (Figure~\ref{fig:val-loss-sft-lora}\,(b)) follows a similar qualitative pattern but with a shallower loss curve: the validation loss decreases steadily and then flattens, with early stopping triggered at global step 4,400 after observing the validation loss had plateaued for several consecutive evaluations. LoRA thus achieves a clear improvement over the base model while updating only about 0.13\% of the parameters, but its final validation loss remains slightly higher than that of the fully fine-tuned SFT model. We therefore select the LoRA checkpoint marked with red dashed line for all subsequent evaluations ($L_{\text{val}} = 0.759$).

For the open-ended question benchmark, we combine human and automated assessment.  As described in Section~\ref{subsec:5}, we first calibrated and selected two LLM judges (Claude-Sonnet-4 and GPT-4.1-mini), and use them primarily for large-scale, automated analysis reported in the appendix ~\ref{appendix:results_analysis}. To obtain reliable headline results, however, we treat human ratings as the primary source of truth and use LLM judges only as secondary, supporting evaluators. Thus, the headline open-ended results reported in this section are based on direct human evaluation of all 143 questions in the open-ended benchmark.

Human annotators scored each open-ended answer along seven dimensions (relevance, correctness, technical level, scientific level, explainability, conciseness, and coherence) on a 0-10 scale. The results in Table~\ref{tab:human_agreement_2} show consistently strong agreement across all criteria: Pearson $r$ ranges from 0.71 to 0.92, Spearman's $\rho$ from 0.67 to 0.90, and both $\kappa_{\text{quad}}$ and Krippendorff's $\alpha$ fall in the 0.7-0.92 range, indicating reliable and reasonably consistent human reference scores.

\begin{table}[!htbp]
    \centering
    \caption{Inter-annotator agreement between the human experts on the open-ended evaluation queries, across the seven evaluation dimensions.}
    \label{tab:human_agreement_2}
    \begin{tabular}{lccccc}
        \hline
        Metric & Pearson $r$ & Spearman $\rho$ & MAE & $\kappa_{\text{quad}}$ & Kripp. $\alpha$ \\
        \hline
        Relevance        & 0.920 & 0.898 & 0.65 & 0.918 & 0.918 \\
        Correctness      & 0.798 & 0.791 & 1.18 & 0.778 & 0.775 \\
        Technical level  & 0.708 & 0.717 & 0.72 & 0.789 & 0.787 \\
        Scientific level & 0.719 & 0.729 & 1.00 & 0.703 & 0.701 \\
        Explainability   & 0.768 & 0.702 & 1.15 & 0.735 & 0.730 \\
        Conciseness      & 0.748 & 0.674 & 1.14 & 0.738 & 0.737 \\
        Coherence        & 0.816 & 0.769 & 1.01 & 0.803 & 0.801 \\
        \hline
    \end{tabular}
\end{table}

We report two composite scores: \emph{factual/technical quality}, defined as the mean of (relevance, correctness, technical level, scientific level), and \emph{communication quality}, defined as the mean of (explainability, conciseness, coherence).  

Averaged over all open-ended questions, the frozen LLaMA~3.1-8B base model attains a factual/technical quality score of $3.83$ with 95\% CI $[3.58, 4.09]$. EnergyGPT (SFT) improves this to $4.76$ (CI $[4.53, 5.00]$), while the LoRA variant reaches $4.03$ (CI $[3.72, 4.33]$). In paired bootstrap comparisons against the base model, EnergyGPT (SFT) yields a mean gain of $+0.93$ points on the 0--10 scale (95\% CI $[+0.72, +1.14] $, $p < 10^{-3}$), whereas EnergyGPT (LoRA) shows a smaller and not statistically significant gain of $+0.23$ points (95\% CI $[-0.05, +0.50] $, $p \approx 0.10$). 

For communication quality, the base model obtains $4.26$ (CI $[3.98, 4.54] $), compared to $6.17$ (CI $[5.97, 6.36]$) for EnergyGPT (SFT) and $5.47$ (CI $[5.08, 5.83]$) for EnergyGPT (LoRA). Relative to the base model, this corresponds to gains of $+1.90$ points for SFT (95\% CI $[+1.59, +2.21]$, $p < 10^{-3}$) and $+1.25$ points for LoRA(95\% CI $[+0.83, +1.66]$, $p < 10^{-3}$). These results indicate that SFT delivers the largest improvements in factual and technical adequacy, while both SFT and LoRA enhanced the clarity and communicative quality of responses.

\begin{figure}[!htbp]
  \centering
  \includegraphics[width=0.75\linewidth]{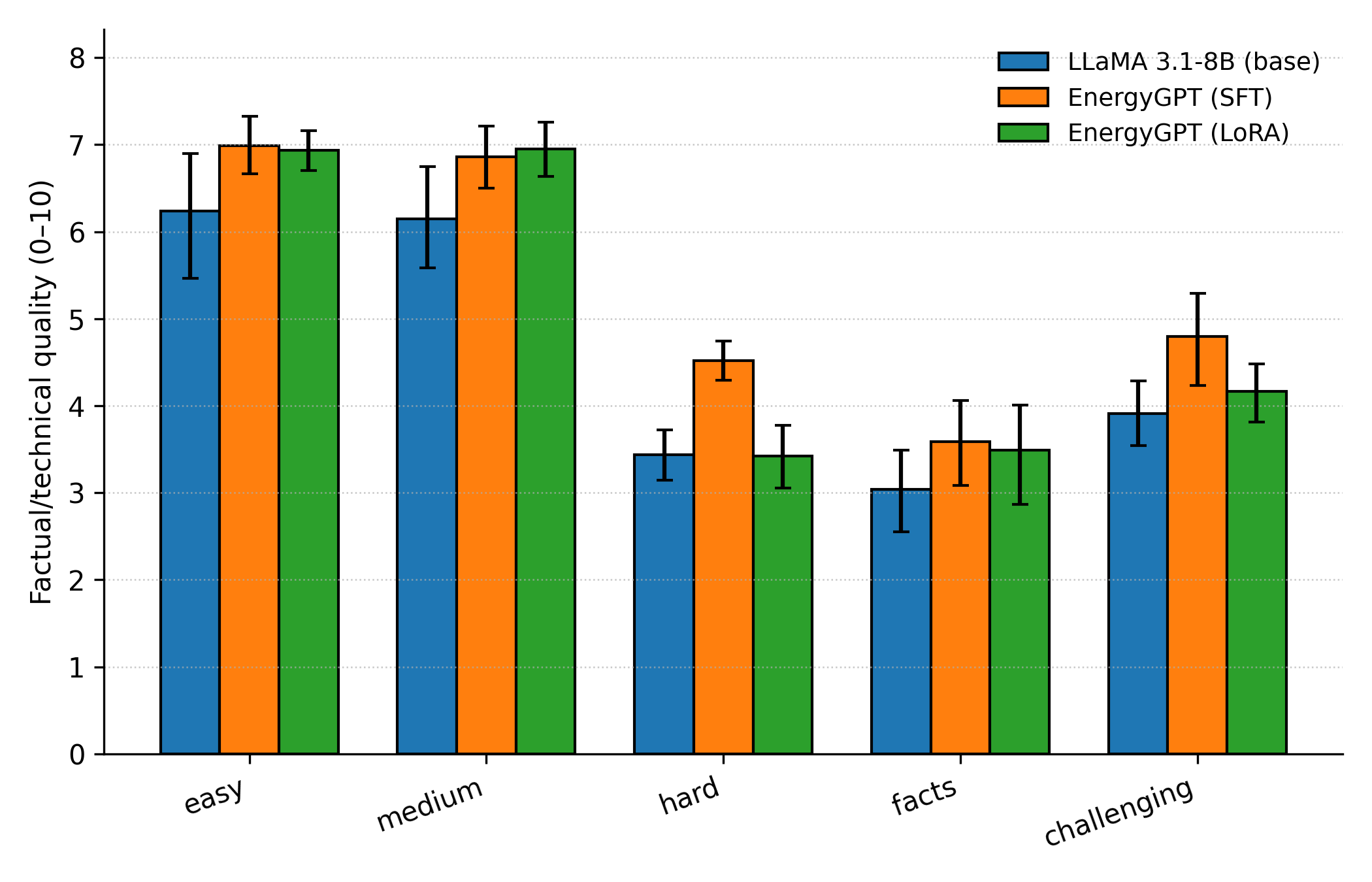}
  \caption{Open-ended factual/technical quality by difficulty, based on
  human ratings (0--10 scale). Error bars show 95\% bootstrap confidence
  intervals over questions.}
  \label{fig:open-ended-factual}
\end{figure}

\begin{figure}[!htbp]
  \centering
  \includegraphics[width=0.75\linewidth]{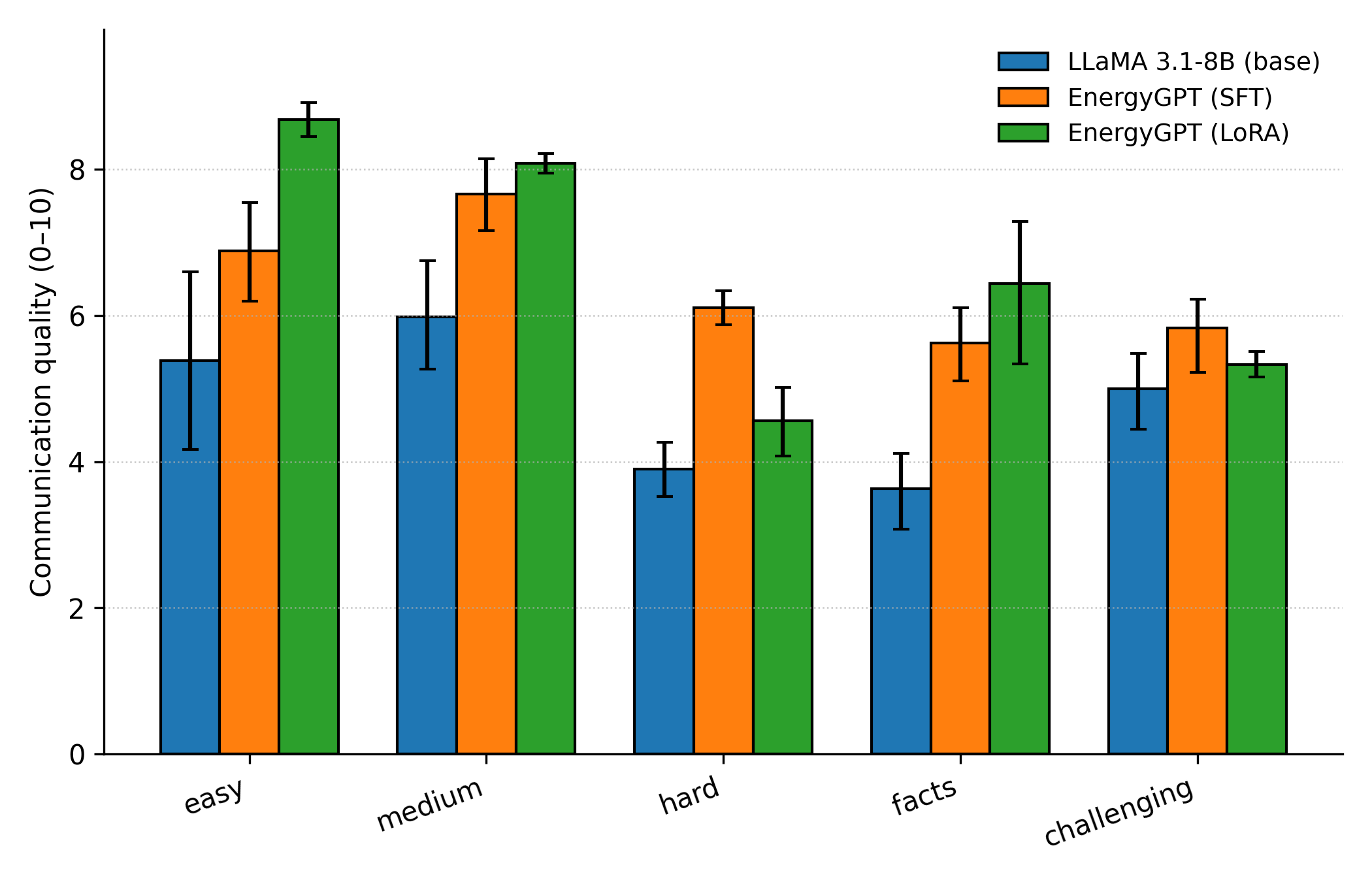}
  \caption{Open-ended communication quality by difficulty, based on
  human ratings (0--10 scale). Error bars show 95\% bootstrap confidence
  intervals over questions.}
  \label{fig:open-ended-comm}
\end{figure}

Figure~\ref{fig:open-ended-factual} breaks these factual/technical scores down by difficulty level (easy, medium, hard, fact-style, and challenging), showing that SFT’s gains are most pronounced on hard and challenging questions, while the two EnergyGPT variants are closer on easier questions. Figure~\ref{fig:open-ended-comm} presents the corresponding communication-quality breakdown, where both EnergyGPT variants consistently outperform the base model across all difficulty levels, with LoRA often slightly exceeding SFT on easier and fact-style questions.

To assess whether our findings are robust beyond human annotators, we also re-scored all open-ended answers with the selected LLM judges (Claude-Sonnet-4 and GPT-4.1-mini), using the same factual/technical and communication composite metrics.  Both judges reproduce the human raters' overall ranking: EnergyGPT (SFT) $>$ EnergyGPT (LoRA) $>$ LLaMA~3.1-8B (base) on factual/technical quality, and both EnergyGPT variants $>$ base on communication quality.  Moreover, the pattern across difficulty levels is highly consistent with human ratings: the largest factual/technical gains for EnergyGPT (SFT) appear on hard and challenging questions, while both SFT and LoRA deliver clear communication improvements across all difficulties, with LoRA fine-tuned model often slightly surpassing SFT fine-tuned model on the easier prompts (see Appendix~\ref{appendix:results_analysis}). Taken together with the strong human-LLM agreement scores reported in Table~\ref{tab:llm_judges_agreement_results}, these results suggest that our conclusions are stable across both human and LLM judges. Samples of generated responses on multiple queries by all models are presented in Appendix~\ref{appendix:responses_queries}.

\begin{figure}[!htbp]
  \centering
  \includegraphics[width=0.65\linewidth]{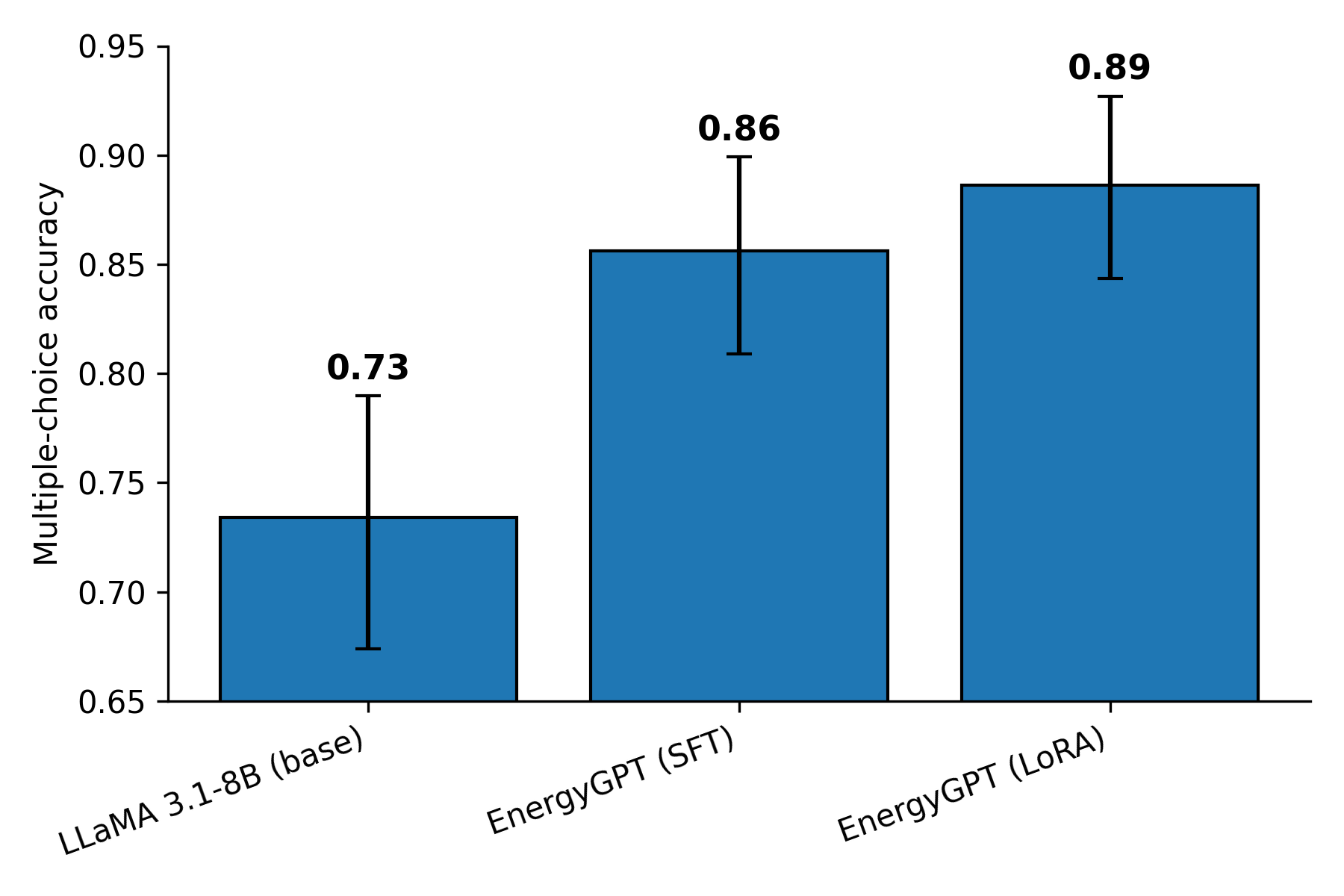}
  \caption{Multiple-choice accuracy of the base model and EnergyGPT
  variants with 95\% bootstrap CIs.}
  \label{fig:mc-accuracy}
\end{figure}

On the multiple-choice subset, we measure exact-match accuracy and estimate uncertainty using nonparametric bootstrap resampling over questions (95\% confidence intervals, CIs). The frozen LLaMA~3.1-8B base model attains an accuracy of $0.73$ with 95\% CI $[0.67, 0.79]$. Both EnergyGPT variants substantially outperform this baseline: EnergyGPT (SFT) reaches $0.86$ accuracy with CI $[0.81, 0.90]$, while EnergyGPT (LoRA) achieves $0.89$ with CI $[0.84, 0.93]$ (Figure~\ref{fig:mc-accuracy}). In terms of improvement over the base model, EnergyGPT (SFT) yields an absolute accuracy gain of $+0.13$ (95\% CI $[+0.07, +0.18]$, $p = 4 \times 10^{-4}$), and EnergyGPT (LoRA) improves by $+0.16$ (95\% CI $[+0.10, +0.21]$, $p < 10^{-3}$). Thus, both full SFT and LoRA-based fine-tuning models provide statistically significant gains on domain-specific multiple-choice questions, with LoRA slightly exceeding the SFT model despite updating only a small fraction of parameters. Samples of responses on multiple choice queries generated by all models are presented in Appendix~\ref{appendix:responses_multi_choice}.

These gains likely arise from improved alignment to energy-domain terminology and concepts, which helps the adapted models distinguish correct answers from plausible distractors in constrained-response settings. The slight advantage of EnergyGPT (LoRA) on this subset may indicate that parameter-efficient adaptation is sufficient to sharpen answer selection behavior for multiple-choice tasks, while the broader updates of full SFT appear to be more beneficial for open-ended factual and technical generation.

\begin{figure}[!htbp]
  \centering
  \includegraphics[width=0.65\linewidth]{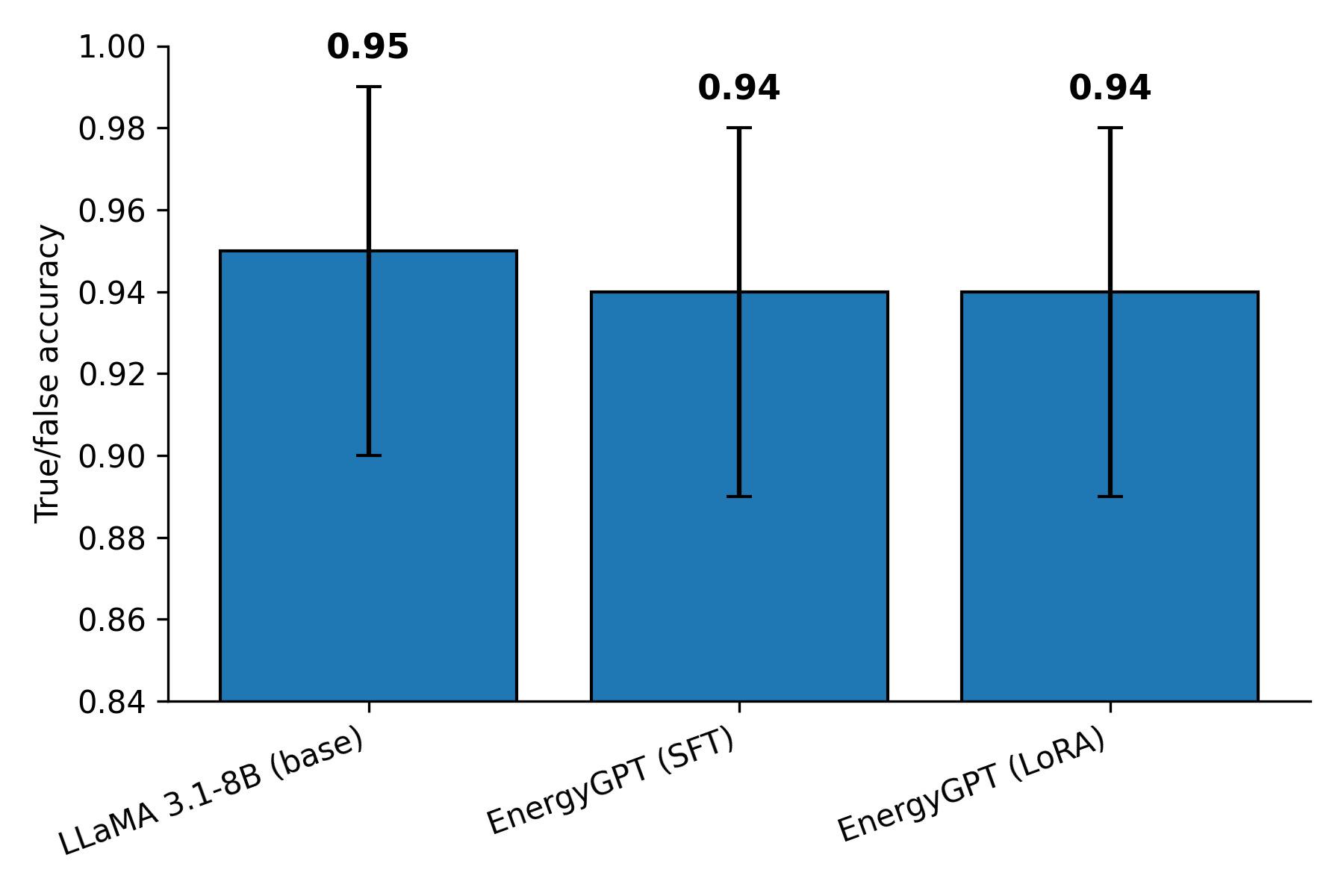}
  \caption{True/False accuracy of the base model and EnergyGPT
  variants with 95\% bootstrap CIs.}
  \label{fig:tf_accuracy}
\end{figure}

For binary factual consistency, we evaluate all models on the true/false subset of the benchmark. Figure~\ref{fig:tf_accuracy} reports accuracy with 95\% bootstrap confidence intervals. The base LLaMA~3.1-8B model achieves an accuracy of $0.95$ with a 95\% CI of $[0.90, 0.99]$. The fully fine-tuned EnergyGPT (SFT) model reaches $0.94$ accuracy with CI $[0.89, 0.98]$, and the LoRA variant likewise attains $0.94$ accuracy with CI $[0.89, 0.98]$. Pairwise differences with the base model are small ($-0.01$) and not statistically significant (95\% CIs include zero; $p > 0.75$), so we interpret these true/false results primarily as a sanity check rather than as evidence of meaningful differences between models. Samples of generated responses on multiple queries by all models are presented in Appendix~\ref{appendix:responses_true_false}.

\subsection{Failure Modes}
To better understand the limitations of \textsc{EnergyGPT} variants, we conducted a review of low-scoring responses across all question types and compared them against baseline outputs from \textsc{LLaMA 3.1-8B} (see examples in Appendix~\ref{appendix:responses_failure_queries}). This analysis revealed three dominant failure patterns. Notably, many of these errors were also present in the base model, often in a more pronounced form, suggesting that a substantial portion of the observed failure modes may be inherited rather than newly introduced by domain adaptation.

\begin{enumerate}
    \item \textbf{Factual Hallucinations:} 
    All models occasionally generated confident yet inaccurate factual statements. For example, EnergyGPT (SFT) reported a total global energy consumption of roughly 5.6~trillion~TWh for 2022, an implausible value inconsistent with authoritative data. 
    EnergyGPT (LoRA) sometimes produced more reasonable-sounding magnitudes, but still hallucinated specifics or mis-scoped quantities. LLaMA~3.1-8B showed similar tendencies, and sometimes returned placeholder text and irrelevant contextual data.

    \item \textbf{Limited Reasoning Depth:} 
    On tasks requiring multi-step or technically rigorous reasoning, EnergyGPT variants produced often coherent and contextually informed responses but sometimes omitted explicit intermediate steps and deeper analytical justification. 
    Relative to LLaMA~3.1-8B, its outputs were more structured, domain-aware, and stylistically consistent, reflecting clear gains from domain adaptation. 
    Nonetheless, the model's reasoning depth remained largely constrained by the underlying base architecture, as the fine-tuning process primarily targeted domain fluency rather than explicit reasoning enhancement.
    
    \item \textbf{Semantic and Conceptual Confusion:} 
    Both models occasionally conflated closely related planning or policy concepts (e.g., misattributing content between the 13\textsuperscript{th} and 14\textsuperscript{th} Five-Year Plans). EnergyGPT exhibited these errors less frequently in structured policy questions but still showed difficulty with fine-grained temporal and institutional distinctions, limitations that persisted from the base model despite fine-tuning.
\end{enumerate}

These findings indicate that, although domain adaptation yields clear gains in linguistic fluency, domain relevance, terminology use, and response structure compared to the base model, important limitations remain in factual calibration and multi-step analytical reasoning, including occasional catastrophic errors such as large numerical mistakes or policy inaccuracies. Our comparison further suggests that many of the dominant factual hallucinations observed in \textsc{EnergyGPT} are inherited from the base \textsc{LLaMA~3.1-8B} model, even though adaptation appears to reduce their frequency in some cases. Residual errors remain particularly evident for numerically sensitive, temporally specific, or policy-related questions, where domain-adapted responses may still sound plausible while being incorrect. In safety-critical applications, \textsc{EnergyGPT} should therefore be used only as a decision-support tool under expert supervision rather than as an autonomous authority. To improve factual reliability in practical use and move toward a more trustworthy version of \textsc{EnergyGPT}, future work will focus on integrating the model with retrieval-augmented generation over curated energy corpora, applying lightweight post-generation checks for numerical consistency and unit ranges against retrieved evidence, and using conservative prompting and abstention policies, where the model is instructed either to ground answers in explicit evidence or to acknowledge uncertainty~\cite{tonmoy2024hallucinationmitigation,zhang2023sirens,manakul2023selfcheckgpt}.

%%%%%%%%%%%%%%%%%%%%%%%%%%%%%%%%%%%%%%%%%%%%%%%%%%
\section{EnergyGPT Deployment}
\label{sec:7}
%%%%%%%%%%%%%%%%%%%%%%%%%%%%%%%%%%%%%%%%%%%%%%%%%%%%%%%%%%%%%%%%%%%%%%%%%%%%%%%%%%%%%%%%%%%%%%%%%%%%%%%%%%%%%%%%%%%%%%%%%%%%%%%%%%%%
To make the EnergyGPT model accessible for interactive inference and integration into downstream applications, we experimented with two deployment strategies:
(1) An on-premises deployment using NVIDIA NIMs, which enabled high-throughput inference within a secure local environment. As part of this deployment, we also developed a lightweight toolkit to generate API keys, track per-user usage, handle authentication, enforce per-user quotas, and logs detailed usage statistics for monitoring and auditing. (2) A cloud-based deployment using Microsoft Azure, where we integrated the model into Azure Machine Learning (AML) Workspace. For this setup, we leveraged Azure API Management (APIM) to expose the model via REST APIs, enforce API key-based access control, and monitor usage metrics.

\subsection{On Premises Deployment with NVIDIA NIMs}
 For secure and high-performance inference in a controlled environment, we deployed EnergyGPT on-premises using NVIDIA Inference Microservices (NIMs). NIM provides a simple and scalable mechanism to serve LLMs via standardized REST endpoints, wrapping the NVIDIA Triton Inference Server and TensorRT-LLM optimization under the hood. We used a server equipped with 4 x NVIDIA A100-SXM4-80GB GPUs. The deployment was containerized and launched using the official NIM LLaMA-3.1-8B Instruct container from NVIDIA's NGC registry. The fine-tuned model (stored in Hugging Face format) was mounted into the container, and served using the command shown in Listing~\ref{lst:nim-deploy}.
\renewcommand{\lstlistingname}{\textbf{Listing}}
\begin{lstlisting}[language=bash,
    caption={On-premises deployment of EnergyGPT using NVIDIA NIM.}, 
    label={lst:nim-deploy}, 
    float
]
docker run -it --rm --gpus all                       \
    --shm-size=64g                                   \
    --ulimit memlock=-1                              \
    --ulimit stack=67108864                          \
    --network=host                                   \
    -e NGC_API_KEY=$NGC_API_KEY                      \
    -e NIM_DISABLE_NGC_INJECTION=true                \
    -e NIM_FT_MODEL=/data/energygpt                  \
    -e NIM_SERVED_MODEL_NAME=energygpt               \
    -v /data:/data                                   \
    nvcr.io/nim/meta/llama-3.1-8b-instruct:1.3.0
\end{lstlisting}

This configuration exposed the EnergyGPT model through a RESTful API endpoint, accessible at \texttt{\seqsplit{http://localhost:8000/v1/completions}}, using an OpenAI-compatible interface. The model was served with full GPU acceleration, benefiting from NVIDIA's runtime optimizations for memory-efficient inference and prompt caching.\\
To ensure secure, multi-tenant access and enforce resource quotas, we implemented a modular API gateway using FastAPI. This service acts as a proxy in front of the EnergyGPT inference endpoint, and authenticates each request via a project-specific API key provided in the \texttt{Authorization} header, enabling role-based access control across multiple organizations and projects.\\
API keys are uniquely issued at the project level and governed by configurable quotas, including maximum request counts and total token budgets. Upon each request, the system performs quota validation, proxies the payload to the NIM backend, parses usage metadata (e.g., prompt and completion token counts), and logs the result to a relational database. %(SQLite or PostgreSQL). 
Usage is aggregated at daily, monthly, and yearly levels for each project and organization, enabling detailed metering and longitudinal analysis.\\
The system supports advanced key lifecycle operations, including:
\begin{itemize}
\item Automatic generation of API keys upon project creation.
\item Seamless regeneration of keys with historical linking and soft deactivation of prior keys.
\item Real-time enforcement of per-key request and token quotas.
\item Tracking of last usage timestamps and project-level usage roll-ups.
\end{itemize}
Administrators of added organizations are granted access to dedicated management endpoints, allowing them to:
\begin{itemize}
\item Create projects and add users dynamically.
\item Generate and revoke API keys, with support for seamless key regeneration and deactivation of prior keys.
\item Assign request and token quotas per key or project.
\item Export detailed API usage logs and key metadata in CSV format for billing, auditing, or analytics.
\end{itemize}

%%%%%%%%%%%%%%%%%%%%%%%%%%%%%%%%%%%%%%
\subsection{Deployment in Azure}
\begin{figure}[!htbp]
    \centering
    \includegraphics[width=0.5\textwidth]{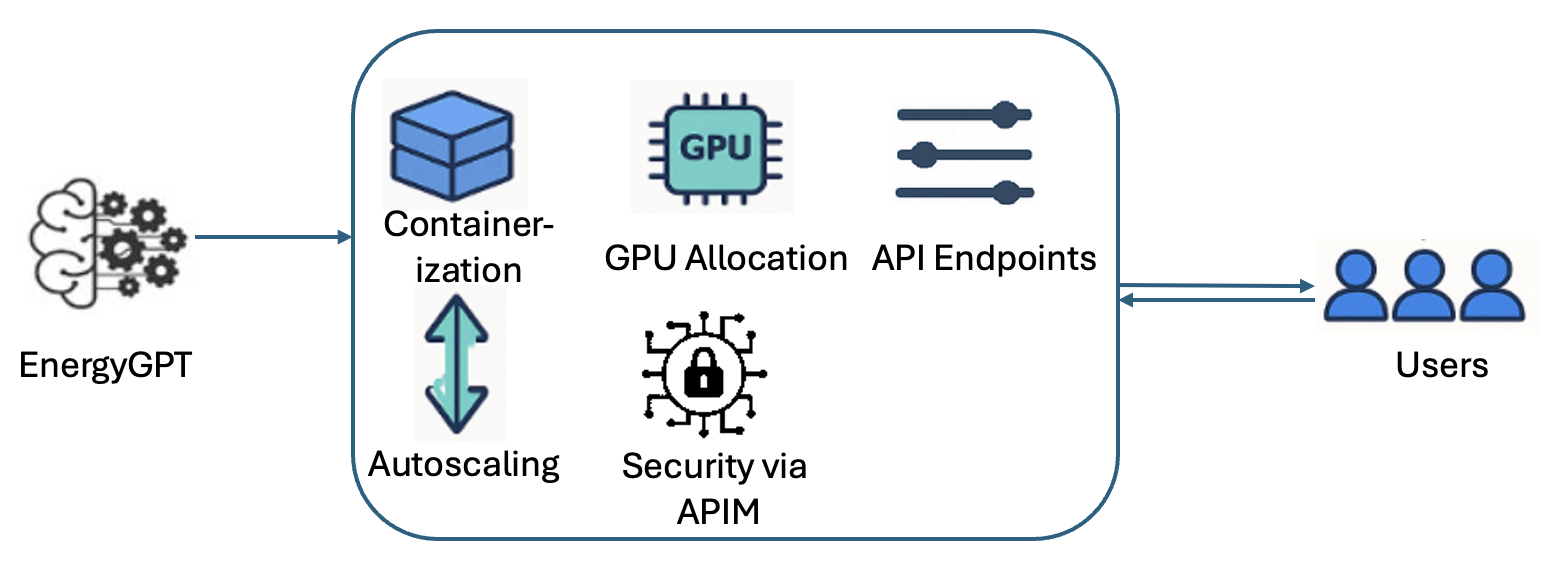}
    \caption{EnergyGPT deployment pipeline on Azure.}
    \label{fig:azure_deployment}
\end{figure}

We tested the deployment of  a production-ready version of EnergyGPT on Microsoft Azure using a managed online endpoint within the Azure Machine Learning (AML) workspace (see Figure~\ref{fig:azure_deployment}). The model was hosted on GPU-enabled virtual machines (\texttt{Standard-NV18ads-A10-v5}) and served via a custom Docker image registered in Azure Container Registry (ACR), which bundled all necessary dependencies for inference. To reduce memory footprint and improve throughput, the model was quantized using 4-bit precision. Azure ML also provided built-in support for versioning, reproducibility, and dynamic resource provisioning, enabling the deployment to elastically scale with workload demands.

To securely expose the model and manage external access, the AML endpoint was integrated with Azure API Management (APIM) \cite{azure_apim_overview, azure_apim_key_concepts}, which served as an authentication and governance layer. All inference requests were routed through APIM's centralized gateway and access to the EnergyGPT API was managed through an API product, \textit{EnergyGPT Access}, to which each organization or internal project subscribed. Upon subscription, a unique API key was issued per project, enabling isolated and traceable access. These keys were required in all client requests and were validated using declarative, header-based policies defined within the APIM configuration, which enforced strict authentication, rejected unauthorized or anonymous traffic, and ensured consistent request formatting prior to reaching the backend endpoint.

The API gateway enforced additional policies at the inbound stage, including key validation, request header normalization, and quota enforcement. These were defined declaratively using APIM's XML-based policy engine and applied uniformly across all requests. The policy framework also enabled advanced features such as token-based routing, header rewriting, and rate limiting for a secure and predictable service behavior.

The integration of Azure API Management (APIM) also enabled per-project usage isolation and flexible credential lifecycle management, including key regeneration and revocation. Administrators could monitor API activity through Azure Monitor and Application Insights, gaining visibility into key metrics such as request volumes, success and failure rates, and token consumption. A self-service developer portal allowed subscribed users to create and manage projects, retrieve keys, track usage, and test the model interactively. A step-by-step guide for setting up APIM can be found in Appendix~\ref{appendix:APIMDoc}.

\textit{For clarity, all experimental results reported in this paper were generated using the full model deployed locally on-premises. The Azure setup was included only as a proof-of-concept deployment test, in which a 4-bit quantized model was used due to resource constraints.}

\section{Generalizability to Other Domains}
\label{sec:generalizability}

Although this work focuses on the energy sector, the proposed pipeline in Figure~\ref{fig:general_pipeline} is largely domain-agnostic. Each stage can be instantiated with alternative corpora and relevance signals to obtain specialized assistants in other fields.

First, the data curation stage (quality filtering, exact and fuzzy deduplication, and semantic relevance filtering) operates on text statistics and embeddings rather than energy-specific heuristics. In our case, energy relevance is enforced via query templates and reference documents; in another domain these would simply be replaced by domain-appropriate prompts and seed documents, while reusing the same filtering and preprocessing infrastructure. Our input-output pair construction similarly relies only on generic document structure (sentences, paragraphs, equations) and a token budget, so the sliding-window pairing strategy can be applied to other corpora with minimal changes.

Second, the supervised fine-tuning stage uses a standard causal language modeling objective and does not depend on energy-specific architectural choices. Adapting the pipeline to biomedicine, for example, would involve constructing a corpus from PubMed articles and clinical guidelines, applying the same curation and pairing steps, and fine-tuning a suitable base model to obtain a clinical assistant. In finance, an analogous instantiation could be built from regulatory filings, analyst reports, and historical news, yielding a model similar in spirit to BloombergGPT but obtained via SFT or LoRA rather than full pretraining.

Third, the evaluation protocol and LLM-as-judge component are also reusable. Our benchmark is built from energy questions, but the scoring rubric relies on generic criteria such as correctness, relevance, and coherence. In another domain, one would substitute a domain-specific question set, reference answers, and an adapted judge prompt (with domain-appropriate instructions), while retaining the same high-level rubric and calibration procedure against human experts.

Finally, the deployment layer-containerization, GPU-backed inference, and API management is entirely domain-agnostic. The same NIM or Azure-based serving stack can host models fine-tuned on other specialized corpora. Thus, EnergyGPT should be viewed not only as a single model, but as a concrete case study of a general recipe for turning a foundation LLM into a production-grade assistant in other domains.

\section{Limitations \& Future Work}
\label{sec:8}

One important direction for future work is the integration of retrieval-augmented generation (RAG) strategies to address the temporal limitations of parametric domain specialization. In this paper, we benchmark the EnergyGPT variants (full SFT and LoRA) against the base LLaMA~3.1-8B model on standalone question-answering tasks, but we do not explore retrieval-based approaches. As a fully parametric model, EnergyGPT relies primarily on knowledge encoded during training, which is effective for relatively stable scientific and engineering concepts but less well suited to rapidly evolving aspects of the energy domain, such as regulations, market conditions, and policy commitments. Consequently, EnergyGPT inherits a practical knowledge cutoff tied to the time span and content of its training corpus, and may require periodic refresh to remain reliable on time-sensitive queries. Future work should therefore compare static domain specialization and retrieval-based methods more directly, and investigate update strategies such as scheduled corpus refresh and re-tuning, lightweight continual adaptation, and RAG pipelines for tasks requiring up-to-date external evidence. Such future work comparisons would help clarify the trade-offs among factual accuracy, latency, computational cost, and domain consistency across different classes of energy-related tasks.

A second open question concerns training dynamics. In this work, both SFT and LoRA training runs were halted early based on the plateauing of validation loss; however, the optimal number of effective epochs required to achieve maximal domain specialization remains an open question. Future research will aim to systematically explore the relationship between training duration, validation dynamics, and downstream task performance, with the goal of identifying convergence trends and balancing the risks of underfitting and overfitting for both full fine-tuning and PEFT-style adaptations.

Third, we observed that the EnergyGPT variants exhibited strong instruction-following behavior, particularly in multiple-choice settings, where the models often returned only the answer key, as prompted. Notably, this emerged without any explicit instruction-tuning stage. This suggests that the fine-tuning process itself may have implicitly reinforced instruction adherence, raising an interesting opportunity to study how domain-specific supervision shapes instruction-following in foundation models.

Fourth, while the benchmark is custom-curated and reviewed by domain experts, its size remains modest relative to the full diversity of energy-domain tasks. Future work will expand the benchmark to cover more fine-grained tasks and additional evaluation formats.

Finally, a fundamental limitation of the current EnergyGPT models is that, like other current large language models, they remain text-only, data-driven systems that reason primarily by matching patterns in the literature, without explicit mechanisms for multi-step reasoning or enforcing physical first-principles consistency. Many tasks in the energy and engineering domains require coherent chains of inference, numerical problem solving, equation handling, adherence to governing physical laws, and causal reasoning. While EnergyGPT generally produces factual and coherent responses on our benchmark, it does not verify its answers against explicit physical constraints or external simulation tools and therefore does not provide formal guarantees of globally consistent or physically valid explanations. As future work, we plan to develop a reasoning-augmented version of EnergyGPT that (i) is trained on carefully curated datasets emphasizing step-by-step analytical reasoning, explicit derivations, unit checks, and physics-aware supervision, and (ii) is coupled, where appropriate, with external tools such as equation solvers and simulators (agents), with the goal of improving physical consistency and robustness on complex, scientifically rigorous queries including under unseen distribution shifts.

From a practical perspective, it is also important to acknowledge the computational footprint of the training process. In our study, training was conducted on shared GPU infrastructure, and direct node-level power telemetry was not collected. For this reason, we do not report a precise carbon-footprint estimate. Nevertheless, the reported training configuration, including the use of 4~$\times$~A100-80GB GPUs and the observed wall-clock training duration, provides a practical indication of the resource demands associated with model adaptation. A more precise accounting of energy consumption and carbon footprint will be considered in future work.

%%%%%%%%%%%%%%%%%%%%%%%%%%%%%%%%%%%%%%%%%%%%%%%%%%
\subsection{Reproducibility and Release Plan}
\label{subsec:reproducibility_release}
A central goal of this work is to make the EnergyGPT pipeline as reproducible as possible. Throughout the paper, we document the data curation, training, evaluation, and deployment workflow in detail, including parameter settings for preprocessing, quality filtering, deduplication, and semantic selection, as well as the hyperparameters and optimization configurations used for supervised fine-tuning and LoRA training with the open-source NVIDIA NeMo and Megatron-LM frameworks. We also provide links and citations for the open-source tools, datasets, and base models used in our experiments (e.g., NeMo, Megatron-LM, The Pile, and LLaMA~3.1-8B), enabling readers to access the same software stack where licensing permits. In addition, we include the prompt templates used both for benchmark response generation and for instructing the LLM judges.

To support reproducibility, we release the following artifacts and scripts:
\begin{itemize}
    \item The EnergyGPT benchmark files, including metadata and reference keys,
    \item The reference query set used to extract the energy-related subset from The Pile, together with scripts for semantic filtering,
    \item The DOI list for the ASME articles included in our corpus,
    \item Training run scripts for both supervised fine-tuning (SFT) and LoRA, together with the corresponding configuration files used in our experiments,
    \item The benchmark and LLM-judge prompt templates used in the study,
    \item Sample script for evaluation using an LLM-judge,
    \item Sample data artifacts illustrating the released scripts.
\end{itemize}

The released materials are available at \url{https://github.com/fitila/energygpt-release} repository.

We do not redistribute ASME full-text articles themselves; only derived artifacts and bibliographic manifests (e.g., DOI lists) are shared. Together, these released resources are intended to allow other researchers to inspect the workflow, reproduce key parts of the pipeline, retrain comparable models, and extend the EnergyGPT benchmark to new tasks.

%%%%%%%%%%%%%%%%%%%%%%%%%%%%%%%%%%%%%%%%%%%%%%%%%%
\section{Conclusion}
\label{sec:9}
In this paper, we introduced EnergyGPT, a specialized language model for the energy sector. Our work spans the entire development pipeline, from the curation and preprocessing of domain-relevant dataset to fine-tuning, evaluation and deployment into production environment. Throughout this work, we proposed a cost-efficient and scalable approach for domain adaptation through supervised fine-tuning, offering a practical alternative to more resource-intensive methods such as training models from scratch or applying continual pretraining. We leveraged NVIDIA NeMo at multiple stages of this work, notably in The Pile data curation and in SFT process. Our results demonstrate that our fine-tuning strategy enhanced the model's performance in domain-specific language understanding and generation. Future work will focus on extending EnergyGPT with improved reasoning capabilities, enabling it to handle more complex, multi-step analytical tasks relevant to the energy domain.
Beyond the energy sector, the proposed data curation and fine-tuning pipeline is intentionally generic: by swapping the energy corpus and relevance queries for domain-specific counterparts, the same recipe can be used to construct specialized assistants in other fields. In this sense, EnergyGPT serves as a reference instantiation of a broader framework for domain adaptation of foundation LLMs.

%%%%%%%%%%%%%%%%%%%%%%%%%%%%%%%%%%%%%%%%%%%%%%%%%%%%%%%%%%%%%%%%%%%%%%%%%%%%%%%%%%%%%%%%%%%%%%%%%%%%%%%%%%%%%%%%%%%%%%%%%%%%%%%%%%%%

\appendices
\section{Data Preprocessing}
\subsection{Data Collection}
\label{appendix:DataCollection}
\noindent\textbf{ASME journals subset.}
We assembled an ASME-specific corpus from our access collection, accrued over two decades, using a two-stage screening based on topical fit and influence. We prioritized flagship titles central to EnergyGPT's scope, e.g., the \emph{ASME Journal of Heat and Mass Transfer}, the \emph{Journal of Mechanical Design}, and the \emph{Journal of Energy Resources Technology}. Journal prioritization combined topical fit (per publisher scope statements) with publicly available influence indicators (e.g., SJR/JCR), rather than convenience sampling.

The corpus was obtained through two access pathways: (i) open-access ASME articles made available under Creative Commons CC-BY licenses (e.g., CC BY 4.0), and (ii) subscription-access ASME articles available to our organization under the ASME Digital Collection subscription. All materials are processed and used exclusively within our lab's access-controlled research and development environment and are not made available outside our organization. We do not redistribute any subscription full text, cleaned or extracted text dumps, or document-level derived artifacts that would enable reconstruction of licensed content.

Accordingly, our public artifacts include only aggregated corpus statistics and bibliographic metadata (e.g., DOIs), and exclude any subscription full text. The final ASME-derived component used in our internal training runs comprises approximately 40k articles spanning roughly two decades of publication and yields on the order of 1.8B tokens after cleaning and normalization.

\subsection{Fuzzy Deduplication}
\label{appendix:Deduplication}
To identify near-duplicate documents in our dataset, we began by vectorizing each document using a hashing-based method, followed by computing MinHash signatures \cite{minhashlsh, leskovec2020mining} to compactly represent their content. These signatures allowed us to efficiently estimate Jaccard similarity without the need for exhaustive pairwise comparisons. We then applied Locality-Sensitive Hashing (LSH) on the MinHash representations \cite{leskovec2020mining} to group potentially similar documents into candidate buckets. Our goal with LSH was to ensure that documents with high similarity would likely be grouped into the same bucket (see the definitions of Jaccard similarity, MinHash, and LSH later in this appendix).
To achieve this, we configured LSH with 20 bands of 13 hash values each, resulting in a total of 260 hash functions per document.\\
Once candidate buckets were formed, we refined them to eliminate false positives introduced during the hashing step. For each bucket, we randomly selected around 10\% anchor documents, compared each sampled document $d_i$ to the remaining documents using Jaccard similarity and only documents with similarity score exceeding $\tau = 0.8$ were marked as duplicates and linked to their corresponding anchors. After processing all buckets, we built a sparse graph where nodes represented documents and edges connected documents identified as near-duplicates. Finally, we extracted connected components from the graph, with each component representing a group of near-duplicates and from each group, we retain a single representative document to construct the final deduplicated dataset. These steps were carried out using NVIDIA NeMo Curator \cite{nvidia_deduplication}, presented in ~\ref{alg:fuzzy_deduplication} is a comprehensive representation of the fuzzy deduplication workflow.

\begin{algorithm}[!htbp]
\caption{Fuzzy Deduplication Pipeline}
\label{alg:fuzzy_deduplication}
\begin{algorithmic}[1]
\State \textbf{Input:} Dataset $\mathcal{D} = \{d_1, d_2, \dots, d_n\}$, Jaccard threshold $\tau = 0.8$
\State \textbf{Output:} Deduplicated dataset $\mathcal{D}'$

\Statex
\State \textbf{MinHash Signature Generation}
\For{each document $d_i \in \mathcal{D}$}
    \State Compute MinHash signature $h_i$ for $d_i$.
\EndFor

\Statex
\State \textbf{Locality-Sensitive Hashing (LSH)}
\State Partition $\{h_i\}$ into 20 bands × 13 hashes per band $\Rightarrow$ 260 hash functions.
\State Group documents into candidate buckets based on matching LSH bands.

\Statex
\State \textbf{Candidate Bucket Refinement}
\For{each candidate bucket $B$}
    \State Randomly select anchor documents from $B$.
    \For{each anchor document $d_a \in B$}
        \For{each remaining document $d_j \in B$}
            \State Compute $J(d_a, d_j)$.
            \If{$J(d_a, d_j) \geq \tau$}
                \State Mark $d_j$ as a near-duplicate \& link it to $d_a$.
            \EndIf
        \EndFor
    \EndFor
\EndFor

\Statex
\State \textbf{Graph-Based Deduplication}
\State Construct a sparse graph $G = (V, E)$ where:
\begin{itemize}
    \item $V$ = set of documents.
    \item $E$ = edges between documents with $J(d_i, d_j) \geq \tau$.
\end{itemize}
\State Extract connected components $\{C_1, C_2, \dots\}$ from $G$.
\State Retain one representative document per component.
\Statex
\State \textbf{Final Dataset Construction}
\State Combine representatives to form $\mathcal{D}'$.
\State \Return $\mathcal{D}'$
\end{algorithmic}
\end{algorithm}

\subsection*{Jaccard Similarity}
The Jaccard similarity between two sets \( S \) and \( T \) is defined as:
\begin{equation}
J(S, T) = \frac{|S \cap T|}{|S \cup T|}
\end{equation}
This metric quantifies the degree of overlap between two sets as the ratio of the size of their intersection to the size of their union. The value of \( J(S, T) \) lies in the range \([0, 1]\), where:
\begin{itemize}
    \item \( J(S, T) = 1 \) indicates that the sets are identical,
    \item \( J(S, T) = 0 \) indicates that the sets are disjoint,
    \item Values closer to 1 indicate greater similarity.
\end{itemize}
In the context of document similarity, the sets \( S \) and \( T \) often represent collections of features such as character-level or word-level $n$-grams (shingles) extracted from the documents.\\
\subsection*{MinHash}
For a positive integer \( k \), a \emph{\( k \)-shingle} of a document \( D \) is any contiguous substring of \( D \) of length \( k \). Each document can thus be represented as a set of \( k \)-shingles.

We construct a binary incidence matrix \( M \) where:
\begin{itemize}
    \item Each row corresponds to a unique \( k \)-shingle in the corpus,
    \item Each column corresponds to a document,
    \item The entry \( M_{r, c} = 1 \) if the \( k \)-shingle in row \( r \) appears in document \( c \), and \( 0 \) otherwise.
\end{itemize}

Let \( \pi \) be a random permutation of the row indices of \( M \). The MinHash value of a document \( D \) under permutation \( \pi \), denoted \( h_\pi(D) \), is the index of the first row (i.e., the first shingle under \( \pi \)) where \( M_{r, D} = 1 \):
\begin{equation}
h_\pi(D) = \min \left\{ \pi(r) \mid M_{r,D} = 1 \right\}
\end{equation}

\paragraph{Proposition.}
For two documents \( D \) and \( E \), the probability that their MinHash values are equal under a random permutation is equal to their Jaccard similarity:
\begin{equation}
\Pr[h_\pi(D)=h_\pi(E)] = J(S_D, S_E)
\end{equation}
where $S_D$ is the shingle set of \( D \).

To approximate this probability, we generate \( n \) independent random permutations \( \pi_1, \dots, \pi_n \) and compute the MinHash signature vector for each document:
\begin{equation}
\text{sig}(D) = \left( h_{\pi_1}(D), h_{\pi_2}(D), \dots, h_{\pi_n}(D) \right)
\end{equation}
For two documents \( D \) and \( E \), the fraction of positions in which their signatures match approximates their Jaccard similarity:
\begin{equation}
\frac{1}{n} \sum_{i=1}^{n} \mathbb{I}\!\left[ h_{\pi_i}(D) = h_{\pi_i}(E) \right] \approx J(D, E)
\end{equation}

This signature provides a compact, lower-dimensional representation of the original document while preserving pairwise Jaccard similarities, enabling efficient near-duplicate detection and clustering.

In practice, generating true permutations of rows is computationally expensive. Instead, a family of hash functions \( h_i : \mathbb{Z} \rightarrow \{1, 2, \dots, m\} \), where \( m \) is the number of rows in \( M \), are used to simulate random permutations.

\subsection*{Locality Sensitive Hashing (LSH)}
Locality-Sensitive Hashing (LSH) is a technique for efficiently identifying similar pairs of items in high-dimensional spaces. When combined with MinHash signatures, LSH enables the detection of near-duplicate sets (or documents) without exhaustively computing pairwise similarities.

The core idea is to amplify the distinction between similar and dissimilar pairs by organizing MinHash signatures into a structured arrangement. Given a MinHash signature of length \( n \), we divide it into \( b \) \emph{bands}, each containing \( r \) rows, such that \( n = b \cdot r \). Within each band, the \( r \)-row sub-vectors of the signature are hashed into buckets.

Two documents are considered candidate duplicates if they hash to the same bucket in at least one band, that is, their sub-signatures are identical in that band.

\paragraph{Collision Probability.}  
Let \( s \in [0, 1] \) denote the Jaccard similarity between two documents. The probability that their sub-signatures match in all \( r \) rows of a given band is \( s^r \), and the probability that they do \emph{not} match in that band is \( 1 - s^r \). The probability that they fail to match in all \( b \) bands is:

\begin{equation}
(1 - s^r)^b
\end{equation}

Therefore, the probability that two documents are identified as candidate duplicates by LSH is:

\begin{equation}
P_{\text{LSH}}(s) = 1 - (1 - s^r)^b
\end{equation}

This function exhibits a sharp threshold behavior: documents with similarity above a chosen threshold are likely to be detected, while dissimilar documents have a low probability of collision. By tuning the parameters \( r \) and \( b \), one can balance false positives and false negatives.

\section{Semantic Filtering Reference Topics}
\label{appendix:topics}
The following list summarizes a sample of the main topics covered in the curated reference queries used to guide semantic filtering of The Pile dataset.
\begin{itemize}
  \item \textbf{Energy Systems and Infrastructure}: power grids, smart grids, microgrids, sector coupling.
  \item \textbf{Renewable Energy Sources}: solar, wind, hydropower, geothermal, biomass, ocean energy.
  \item \textbf{Fossil Fuels}: coal, oil, petroleum, natural gas, exploration, extraction, refining.
  \item \textbf{Nuclear Energy}: fission reactors, small modular reactors, nuclear fusion, waste management.
  \item \textbf{Hydrogen Economy}: production, storage, transport, fuel cells, industrial and transport applications.
  \item \textbf{Energy Policy and Economics}: climate policy, carbon pricing, subsidies, energy market structures.
  \item \textbf{Industrial Energy Systems}: combined heat and power, decarbonization, process electrification.
  \item \textbf{Carbon Management}: carbon capture, direct air capture, CCUS.
  \item \textbf{Emerging Technologies}: AI in energy, blockchain, advanced materials, next-generation photovoltaics.
  \item \textbf{Energy Science and Engineering}: thermodynamics, heat engines, gas turbines, energy physics, chemical systems.
  \item \textbf{Energy Efficiency and Conservation}: building efficiency, behavioral change, efficient systems.
  \item \textbf{Environmental Impacts}: emissions, ecosystem effects, sustainability assessments.
  \item \textbf{Energy Production and Consumption}: conversion processes, consumption patterns, efficiency metrics.
  \item \textbf{Energy Storage Systems}: batteries, thermal storage, mechanical storage, hydrogen storage.
  \item \textbf{Grid Integration of Renewables}: balancing, smart inverters, interconnection, stability.
  \item \textbf{Energy in Biological Systems}: bioenergy, biomass utilization.
\end{itemize}

%%%%%%%%%%%%%%%%%%%%%%%%%%%%%%%%%%%%%%%%%%%%%%%%%%%%%%%%%%
%%%%%%%%%%%%%%%%%%%%%%%%%%%%%%%%%%%%%%%%%%%%%%%%%%%%%%%%%%

\section{Training Pairs Generation}
\label{appendix:training_pairs}
For clarity, Algorithms~\ref{alg:pile_pair_generation} and~\ref{alg:scientific_pair_generation} summarize the data pair-construction procedures used for The Pile and the scientific-literature corpus, respectively.

\begin{algorithm}[H]
\caption{Input-Output Pair Construction for The Pile}
\label{alg:pile_pair_generation}
\footnotesize
\begin{algorithmic}[1]
\State \textbf{Input:} Document $d$, chunk token limit $L \approx 600$
\State \textbf{Output:} Training-pair set $\mathcal{P}$

\Statex
\State \textbf{Chunking}: Split $d$ into sentence-aware and equation-aware chunks $[C_1, \dots, C_T]$ using the LLaMA~3.1-8B tokenizer, while preserving sentence boundaries and equations and ensuring that each chunk remains within the token limit $L$

\Statex
\State \textbf{Pair Construction}
\State Initialize $\mathcal{P} \gets \emptyset$
\For{$i = 1$ to $T-5$ step $2$}
    \State $X \gets \text{concat}(C_i, C_{i+1}, C_{i+2})$
    \State $Y \gets \text{concat}(C_{i+3}, C_{i+4}, C_{i+5})$
    \State Add the pair $(X, Y)$ to $\mathcal{P}$
\EndFor

\Statex
\State \textbf{Return} $\mathcal{P}$
\end{algorithmic}
\end{algorithm}

%%%%%%%%%%%%%%%%%%%%%%%%%
%%%%%%%%%%%%%%%%%%%%%%%%%%%%%%%%%%%%%%%%%%%%%%%%%%%%%%%%%%%%
\begin{algorithm}[H]
\caption{Input-Output Pair Construction for Scientific Literature}
\label{alg:scientific_pair_generation}
\footnotesize
\begin{algorithmic}[1]
\State \textbf{Input:} Document $d$, maximum pair length $L_{\max} = 4096$
\State \textbf{Output:} Training-pair set $\mathcal{A}$

\Statex
\State \textbf{Segmentation and Tokenization}
\State Split $d$ into paragraph-aware, sentence-aware, and equation-aware segments $[P_1, \dots, P_T]$ and Tokenize all segments using the LLaMA~3.1-8B tokenizer

\Statex
\State \textbf{Pair Construction}
\State Initialize $\mathcal{A} \gets \emptyset$
\For{$i = 1$ to $T-1$}
    \State Initialize $X \gets \emptyset$, $Y \gets \emptyset$
    \State Initialize input token count $\ell_X \gets 0$ and output token count $\ell_Y \gets 0$
    
    \State Add consecutive paragraphs starting at $P_i$ to $X$ until $\ell_X$ is close to half of $L_{\max}$
    \State Add the subsequent consecutive paragraphs to $Y$ until $\ell_X + \ell_Y \leq L_{\max}$
    \State Adjust the boundary between $X$ and $Y$ to keep input and output lengths roughly balanced

    \If{$X \neq \emptyset$ and $Y \neq \emptyset$}
        \State Add $(X, Y)$ to $\mathcal{A}$
    \EndIf
\EndFor

\Statex
\State \textbf{Return}
\State \Return $\mathcal{A}$
\end{algorithmic}
\end{algorithm}

%%%%%%%%%%%%%%%%%%%%%%%%%%%%%%%%%%%%%%%%%%%%%%%%%%%%%%%%%%%%%%%%%%%
%%%%%%%%%%%%%%%%%%%%%%%%%%%%%%%%%%%%%%%%%%%%%%%%%%%%%%%%%%%%%%%%%%%
\section{Training Infrastructure and Optimization}
\label{appendix:Training}

To optimize computational efficiency and training stability, we use an optimized training infrastructure.\\

\textbf{Optimizer}: During training, We use the Adam optimizer with $\beta_1 = 0.9$ and a reduced $\beta_2 = 0.95$. Lowering $\beta_2$ decreases the influence of long-term gradient variance, enabling the optimizer to adapt more responsively to recent updates. \\

\textbf{Batching and Gradient Accumulation}: To support stable training and effective utilization of GPU memory, we employed gradient accumulation to achieve the desired global batch size without exceeding hardware limits. This technique allows for the accumulation of gradients over several micro-batches before performing a single optimization step, thereby simulating a larger effective batch size while maintaining manageable memory usage per device.
In our configuration, each GPU processed a micro-batch size of 8 samples per forward-backward pass. By accumulating gradients over multiple steps and leveraging tensor parallelism across 4 GPUs, we achieved a global batch size of 256 samples. This strategy enabled us to maintain a high throughput while preventing memory overflow on the A100-80GB GPUs.
The effective batch size B is given by:
\begin{equation}
    B = b \cdot k \cdot N
\end{equation}
where $b$ is the micro-batch size per GPU, $k$ is the number of accumulation steps, and $N$ is the number of GPUs used in training. 

\textbf{Random Seed and Number of Runs}:
All experiments are run with a fixed random seed $\texttt{SEED} = 1234$ in the NeMo/Megatron configuration. Each model variant is trained once, and all reported metrics are computed from this single run.

\textbf{Efficient Tensor Parallelism}: To enable the training of a large model on a limited number of GPUs, we utilized Tensor Parallelism (TP) to partition the model across devices at the level of individual tensor operations. Specifically, we adopted a TP size of 4, corresponding to the number of A100-80GB GPUs available in our compute cluster. Tensor Parallelism allows layers, particularly the attention and feedforward modules, to be split along their hidden dimensions, with each GPU computing a fraction of the operation and exchanging partial results via high-speed interconnects.
This intra-layer parallelism strategy significantly reduces per-device memory requirements and enables training models that would otherwise exceed the memory capacity of a single GPU.\\

\textbf{Validation Loss}:
In large language models, the validation loss is typically computed using the cross-entropy loss function over a sequence of tokens. Given a validation dataset \( D_{\text{val}} = \{(x^{(i)}, y^{(i)})\}_{i=1}^{N} \), where \( x^{(i)} \) is the input context and \( y^{(i)} \) is the target token, the average validation loss is defined as:

\begin{equation}
\mathcal{L}_{\text{val}} = -\frac{1}{N} \sum_{i=1}^{N} \log p_\theta\left(y^{(i)} \mid x^{(i)}\right)
\end{equation}

For autoregressive language modeling, where the task is to predict the next token in a sequence, the loss over a tokenized sequence \( \{t_1, t_2, \dots, t_T\} \) is given by:

\begin{equation}
\mathcal{L}_{\text{val}} = -\frac{1}{T} \sum_{t=1}^{T} \log p_\theta\left(t_t \mid t_1, t_2, \dots, t_{t-1}\right)
\end{equation}

Here, \( p_\theta(\cdot) \) denotes the model's predicted probability distribution parameterized by \(\theta\), and \( T \) is the total number of tokens in the validation set. This loss quantifies how well the model predicts the next token in context and is commonly reported during training as an indicator of generalization.

The validation loss is often converted into \textbf{perplexity}, which can be thought of as the average branching factor or the model's uncertainty in choosing the next token.

Perplexity is defined as the exponential of the cross-entropy loss:

\begin{equation}
\text{Perplexity} = \exp(\mathcal{L}_{\text{val}})
\end{equation}

A lower perplexity indicates better language modeling performance, as it implies the model is more confident and accurate in its predictions. For example, a perplexity of 1 corresponds to perfect prediction, while higher values reflect increasing uncertainty or error.

%%%%%%%%%%%%%%%%%%%%%%%%%%%%%%%%%%%%%%%%%%%%%%%%%%%%%%%%%%%%%%%%%%%
\section{EnergyGPT Benchmark: Samples}
\label{appendix:Samples}
%%%%%%%%%%%%%%%%%%%%%%%%%%%%%%%%%%%%%%%%%%%%%%%%%%%%%%%%%%%%%%%%%%%
%%%%%%%%%%%%%%%%%%%%%%%%%%%%%%%%%%%%%%%%%%%%%%%%%%%%%%%%%%%%%%%%%%%

\begin{titlebox}{Sample of True/False Statements}
\footnotesize
\textbf{Question 1:}\\
Newton's first law of motion states that an object at rest will remain at rest unless acted upon by an external force.\\
\textbf{Answer: True}\\[4pt]

\textbf{Question 2:}\\
Ionic bonds involve the transfer of electrons between atoms to form ions.\\
\textbf{Answer: True}\\[4pt]

\textbf{Question 3:}\\
The law of conservation of energy states that energy can be created or destroyed.\\
\textbf{Answer: False}\\[4pt]

\textbf{Question 4:}\\
CCS involves capturing $CO_2$ emissions from industrial processes and storing them underground in geological formations.\\
\textbf{Answer: True}\\[4pt]

\textbf{Question 5:}\\
Enhanced oil recovery (EOR) is a process that utilizes $CO_2$ injection to increase oil production from reservoirs.\\
\textbf{Answer: True}
\end{titlebox}
\begin{center}
\small\centering \textbf{Text Box 1:} Sample from the True/False Statements benchmark.
\end{center}

\begin{titlebox}{Tiered Difficulty Energy Questions}
\footnotesize
\textbf{Easy Questions}
\begin{itemize}
  \item What is the primary source of energy for Earth's climate system?
  \item How do wind turbines generate electricity from wind?
  \item What is the basic principle behind hydroelectric power generation?
  \item What is energy efficiency, and why does it matter in everyday appliances?
\end{itemize}

\medskip
\textbf{Medium Questions}
\begin{itemize}
  \item Explain how nuclear fission produces energy in a nuclear reactor.
  \item What are biofuels, and how do they compare to traditional fossil fuels in terms of emissions and sustainability?
  \item Explain the concept of energy return on investment (EROI) and its significance in evaluating energy sources.
  \item How does tidal energy work, and what are its potential environmental impacts?
  \item Discuss the advantages and limitations of using hydrogen as an alternative fuel source.
\end{itemize}

\medskip
\textbf{Hard Questions}
\begin{itemize}
  \item How can perovskite materials be engineered to improve the efficiency and stability of next-generation solar cells?
  \item What are the potential risks and benefits of large-scale geoengineering projects aimed at mitigating climate change?
  \item Explain how the second law of thermodynamics limits the efficiency of various energy conversion systems, such as engines and refrigerators.
  \item How do quantum mechanical effects influence the performance of advanced photovoltaic materials?
  \item Evaluate the potential of nanotechnology in enhancing the energy storage capacity and charging speeds of supercapacitors.
  \item Discuss the challenges and potential solutions for carbon capture and storage technologies in reducing atmospheric CO$_2$ levels.
\end{itemize}

\medskip
\textbf{Challenging Questions}
\begin{itemize}
  \item How can genetic engineering of microorganisms be optimized to increase the yield and reduce the production costs of advanced biofuels, and what are the ecological risks associated with deploying such organisms at scale?
  \item What innovative approaches are being researched for the long-term storage or neutralization of nuclear waste, and how do they compare in terms of safety, cost, and public acceptance?
  \item What are the latest developments in harvesting energy from ambient sources like radiofrequency waves, vibrations, or temperature differentials, and how can these technologies be optimized for powering low-energy devices?
  \item How is artificial intelligence being used to optimize energy consumption and distribution in smart grids, and what are the cybersecurity risks associated with increased digitalization?
\end{itemize}

\medskip
\textbf{Fact-Based Questions}
\begin{itemize}
  \item What are the precise parameters (temperature and pressure) achieved in the most recent successful fusion experiment at the ITER facility?
  \item What is the exact cost per kilowatt-hour (kWh) of producing green hydrogen using the latest electrolysis technology as reported in a 2023 industry analysis?
  \item How many electric vehicles (EVs) were sold worldwide in 2022, and what percentage increase does this represent over 2021 sales figures?
  \item List the top five countries in lithium production for 2022, including the exact amounts produced in metric tons.
  \item What are the detailed outcomes of the 2022 United Nations Climate Change Conference (COP27) regarding international commitments to phase out coal-fired power plants?
\end{itemize}

\end{titlebox}

\vspace{-0.4em}
\begin{center}
    \small \textbf{Text Box 2:} Sample questions covering basic to advanced energy-related concepts and factual recall.
\end{center}

%%%%%%%%%%%%%%%%%%%%%%%%%%%%%%%%%%%%%%%%%%%%%%%%%%%%%%%%%%%%%%%%%%%
%%%%%%%%%%%%%%%%%%%%%%%%%%%%%%%%%%%%%%%%%%%%%%%%%%%%%%%%%%%%%%%%%%%

\begin{titlebox}{Sample Questions: Multiple Choice Q \& A}
\footnotesize
\textbf{Question 1}\\
Which renewable energy technology uses the energy of ocean waves to generate electricity?
\begin{enumerate}[label=\Alph*),leftmargin=1.5em,nosep]
  \item Wave power
  \item Tidal power
  \item Hydropower
  \item Ocean thermal energy conversion
\end{enumerate}
\textbf{Answer: A}

\medskip
\textbf{Question 2}\\
What is the term for the process of converting biomass into liquid fuels through the action of microorganisms?
\begin{enumerate}[label=\Alph*),leftmargin=1.5em,nosep]
  \item Biomass liquefaction
  \item Biomass gasification
  \item Biomass fermentation
  \item Biomass pyrolysis
\end{enumerate}
\textbf{Answer: C}

\medskip
\textbf{Question 3}\\
What is the concept of the circular economy in sustainable development?
\begin{enumerate}[label=\Alph*),leftmargin=1.5em,nosep]
  \item Continuing the linear 'take-make-dispose' model
  \item Promoting resource conservation and reuse
  \item Ignoring environmental impacts
  \item Maximizing resource extraction
\end{enumerate}
\textbf{Answer: B}

\medskip
\textbf{Question 4}\\
Which of the following is an example of an endothermic reaction?
\begin{enumerate}[label=\Alph*),leftmargin=1.5em,nosep]
  \item Combustion of gasoline
  \item Photosynthesis
  \item Formation of rust
  \item Freezing of water
\end{enumerate}
\textbf{Answer: B}

\end{titlebox}

\vspace{-0.4em}
\begin{center}
  \small \textbf{Text Box 3:} Sample questions from the multiple-choice Q \& A benchmark.
\end{center}

%%%%%%%%%%%%%%%%%%%%%%%%%%%%%%%%%%%%%%%%%%%%%%%%%%%%%%%%%%%%%%%%%%%
%%%%%%%%%%%%%%%%%%%%%%%%%%%%%%%%%%%%%%%%%%%%%%%%%%%%%%%%%%%%%%%%%%%
\section{LLM Judge Calibration}
\label{appendix:llm_judge}
\subsection{LLM-Judge Prompts}
The LLM-judge prompt used to generate scores for responses to open-ended queries from the benchmark is provided in Listing~\ref{lst:llm_judge_open_ended}. The prompt instructs the judge to score answers along multiple dimensions, with clear definitions and scoring guidance for each dimension, and the judge is expected to return a structured JSON object containing the scores. All judge evaluations are run with deterministic settings (Temperature $T = 0$).
%%%%%%%%%
% \begin{lstlisting}[
%     basicstyle=\ttfamily\footnotesize,
%     breaklines=true,
%     breakatwhitespace=true,
%     columns=fullflexible,
%     caption={Prompt used for instructing the LLM judge in open-ended queries.},
%     label={lst:llm_judge_open_ended}
% ]

\begin{lstlisting}[
    basicstyle=\ttfamily\tiny,
    breaklines=true,
    breakatwhitespace=true,
    columns=fullflexible,
    aboveskip=2pt,
    belowskip=2pt,
    lineskip=-1pt,
    xleftmargin=0.5em,
    frame=single,
    framesep=2pt,
    caption={Prompt used for instructing the LLM judge in open-ended queries.},
    label={lst:llm_judge_open_ended}
]

You are an expert scientific and technical evaluator in the energy field.
Your task is to score a model's answer to a question.

You must assign an integer score from 0 to 10 (inclusive) for each of the following dimensions.

Definitions:
- relevance: How directly the answer addresses the question and its key parts.
- correctness: Factual and numerical accuracy with respect to established knowledge in the energy domain.
- technical_level: Depth and specificity of technical/engineering detail (e.g., methods, equations, system constraints), beyond a high-level summary.
- scientific_level: Use of appropriate physical principles, laws, and scientifically sound arguments (even if not fully derived step by step).
- explainability: How clearly the answer is presented and how easy it is to understand the reasoning and intuition behind the main points. Do not require a long chain-of-thought; short but clear explanations can still receive high scores.
- conciseness: Avoids unnecessary verbosity while covering all important points.
- coherence: Logical consistency, absence of contradictions, and smooth flow.

Scoring guidance:
- 0-2: Very poor / mostly missing or incorrect.
- 3-4: Weak and partially correct, major issues.
- 5-6: Mixed quality, some correct and some incorrect or unclear parts.
- 7-8: Good to very good, mostly correct with minor issues.
- 9-10: Excellent, highly accurate and well-explained.

If the answer is empty, off-topic, or unusable, set all scores to 0.

Question:
{query}

Model response:
{response}

Now return ONLY a single valid JSON object with this exact structure:

{
  "scores": {
    "relevance": <int>,
    "correctness": <int>,
    "technical_level": <int>,
    "scientific_level": <int>,
    "explainability": <int>,
    "conciseness": <int>,
    "coherence": <int>
  }
}
% \end{lstlisting}

%%%%%%%%%%%%%%%%%%%%%%%%%%%%%%%%%%%%%%%%%%%%%%%%%%%%%%%%%%%%%%%%%%%%
\subsection{LLM-Judge Candidates and Calibration Results}
\label{appendix:llm_judge_models}
\begin{figure}[!htbp]
    \centering
    \includegraphics[width=0.5\textwidth]{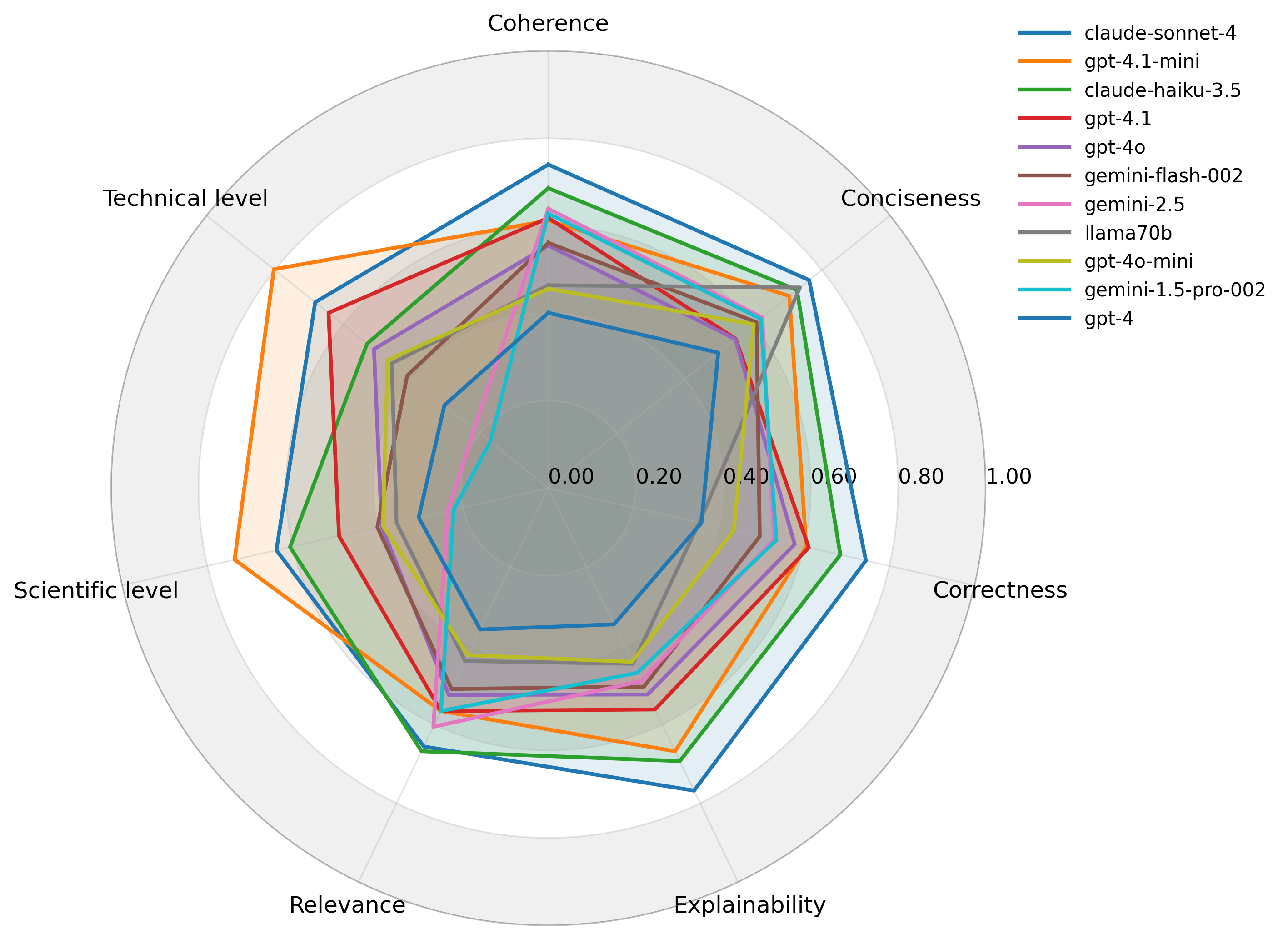}
    \caption{Per-dimension quadratic-weighted Cohen's kappa ($\kappa_{\text{quad}}$) between each candidate LLM judge and the mean human scores across the seven evaluation dimensions.}
    \label{fig:llm-judge-kappa-radar-all}
\end{figure}

We evaluated several candidate LLM judges on 118 query-response pairs generated by the LLaMA3.1-8B model. To quantify alignment with human ratings across our seven evaluation dimensions, we report the quadratic-weighted Cohen's kappa ($\kappa_{\text{quad}}$) and Krippendorff's alpha ($\alpha$) between each judge and the mean human scores in Table~\ref{tab:llm_judge_qwk_full} and~\ref{tab:llm_judge_krippendorff_full}, respectively. Pearson correlation ($r$) and Spearman correlation ($\rho$) results are reported in Tabless~\ref{tab:llm_judge_pearson_full} and~\ref{tab:llm_judge_spearman_full}, respectively. Error-based performance is summarized using mean absolute error (MAE) in Table~\ref{tab:llm_judge_mae_full}. Figure~\ref{fig:llm-judge-kappa-radar-all} further visualizes per-dimension agreement using a radar plot of $\kappa_{\text{quad}}$ for all candidate judges.

\begin{table*}[!htbp]
    \centering
    \small
    \caption{Quadratic Weighted Kappa ($\kappa$) between each candidate LLM judge and the mean human scores across evaluation dimensions.}
    \label{tab:llm_judge_qwk_full}
    \resizebox{\textwidth}{!}{%
    \begin{tabular}{lccccccccccc}
        \hline
        Metric &
        Claude Haiku 3.5 &
        Claude Sonnet 4 &
        Gemini 1.5 Pro-002 &
        Gemini 2.5 &
        Gemini Flash-002 &
        GPT-4 &
        GPT-4.1 &
        GPT-4.1 Mini &
        GPT-4o &
        GPT-4o Mini &
        LLaMA 3.3 70B \\
        \hline
        Coherence        & 0.690 & 0.740 & 0.630 & 0.640 & 0.560 & 0.400 & 0.620 & 0.610 & 0.560 & 0.460 & 0.460 \\
        Conciseness      & 0.730 & 0.760 & 0.620 & 0.630 & 0.610 & 0.500 & 0.550 & 0.710 & 0.550 & 0.600 & 0.740 \\
        Correctness      & 0.680 & 0.740 & 0.530 & 0.530 & 0.500 & 0.360 & 0.610 & 0.600 & 0.580 & 0.430 & 0.350 \\
        Explainability   & 0.690 & 0.770 & 0.470 & 0.490 & 0.500 & 0.350 & 0.560 & 0.670 & 0.520 & 0.440 & 0.450 \\
        Relevance        & 0.670 & 0.660 & 0.570 & 0.610 & 0.510 & 0.360 & 0.570 & 0.570 & 0.530 & 0.420 & 0.440 \\
        Scientific level & 0.610 & 0.640  & 0.220 & 0.240 & 0.400 & 0.300 & 0.490 & 0.740 & 0.390 & 0.390 & 0.360 \\
        Technical level  & 0.530 & 0.680  & 0.170 & 0.220 & 0.410 & 0.300 & 0.640 & 0.800 & 0.510 & 0.470 & 0.460 \\
        \hline
    \end{tabular}
    }%
\end{table*}

\begin{table*}[!htbp]
    \centering
    \small
    \caption{Krippendorff's alpha ($\alpha$) for interval data between each candidate LLM judge and the mean human scores across evaluation dimensions.}
    \label{tab:llm_judge_krippendorff_full}
    \resizebox{\textwidth}{!}{%
    \begin{tabular}{lccccccccccc}
        \hline
        Metric &
        Claude Haiku 3.5 &
        Claude Sonnet 4 &
        Gemini 1.5 Pro-002 &
        Gemini 2.5 &
        Gemini Flash-002 &
        GPT-4 &
        GPT-4.1 &
        GPT-4.1 Mini &
        GPT-4o &
        GPT-4o Mini &
        LLaMA 3.3 70B \\
        \hline
        Coherence        & 0.670 & 0.740 & 0.590 & 0.600 & 0.510 & 0.150 & 0.540 & 0.530 & 0.470 & 0.310 & 0.320 \\
        Conciseness      & 0.720 & 0.760 & 0.610 & 0.620 & 0.600 & 0.410 & 0.480 & 0.690 & 0.480 & 0.560 & 0.720 \\
        Correctness      & 0.670 & 0.740 & 0.510 & 0.510 & 0.460 & 0.180 & 0.550 & 0.530 & 0.500 & 0.300 & 0.140 \\
        Explainability   & 0.680 & 0.760 & 0.440 & 0.460 & 0.460 & 0.040 & 0.470 & 0.630 & 0.410 & 0.300 & 0.300 \\
        Relevance        & 0.650 & 0.640 & 0.540 & 0.570 & 0.460 & 0.040 & 0.460 & 0.450 & 0.410 & 0.240 & 0.280 \\
        Scientific level & 0.610 & 0.600 & 0.020 & 0.030 & 0.370 & 0.070 & 0.380 & 0.730 & 0.210 & 0.220 & 0.150 \\
        Technical level  & 0.500 & 0.680  & -0.130 & -0.040 & 0.390 & 0.060 & 0.610 & 0.800 & 0.430 & 0.370 & 0.340 \\
        \hline
    \end{tabular}
    }%
\end{table*}

\begin{table*}[!htbp]
    \centering
    \small
    \caption{Pearson correlation ($r$) between each candidate LLM judge and the mean human scores across evaluation dimensions.}
    \label{tab:llm_judge_pearson_full}
    \resizebox{\textwidth}{!}{%
    \begin{tabular}{lccccccccccc}
        \hline
        Metric &
        Claude Haiku 3.5 &
        Claude Sonnet 4 &
        Gemini 1.5 Pro-002 &
        Gemini 2.5 &
        Gemini Flash-002 &
        GPT-4 &
        GPT-4.1 &
        GPT-4.1 Mini &
        GPT-4o &
        GPT-4o Mini &
        LLaMA 3.3 70B \\
        \hline
        Coherence        & 0.780 & 0.770 & 0.700 & 0.720 & 0.660 & 0.780 & 0.830 & 0.830 & 0.790 & 0.770 & 0.750 \\
        Conciseness      & 0.790 & 0.790 & 0.740 & 0.750 & 0.720 & 0.770 & 0.800 & 0.820 & 0.810 & 0.750 & 0.820 \\
        Correctness      & 0.740 & 0.750 & 0.600 & 0.600 & 0.570 & 0.640 & 0.830 & 0.840 & 0.830 & 0.760 & 0.710 \\
        Explainability   & 0.780 & 0.800 & 0.610 & 0.630 & 0.610 & 0.740 & 0.820 & 0.840 & 0.810 & 0.740 & 0.740 \\
        Relevance        & 0.730 & 0.710 & 0.630 & 0.670 & 0.590 & 0.700 & 0.790 & 0.790 & 0.790 & 0.730 & 0.710 \\
        Scientific level & 0.630 & 0.730 & 0.420 & 0.430 & 0.490 & 0.630 & 0.760 & 0.800 & 0.740 & 0.710 & 0.680 \\
        Technical level  & 0.640 & 0.700  & 0.420 & 0.500 & 0.490 & 0.620 & 0.770 & 0.800 & 0.730 & 0.730 & 0.700 \\
        \hline
    \end{tabular}
    }%
\end{table*}

\begin{table*}[!htbp]
    \centering
    \small
    \caption{Spearman correlation ($\rho$) between each candidate LLM judge and the mean human scores across evaluation dimensions.}
    \label{tab:llm_judge_spearman_full}
    \resizebox{\textwidth}{!}{%
    \begin{tabular}{lccccccccccc}
        \hline
        Metric &
        Claude Haiku 3.5 &
        Claude Sonnet 4 &
        Gemini 1.5 Pro-002 &
        Gemini 2.5 &
        Gemini Flash-002 &
        GPT-4 &
        GPT-4.1 &
        GPT-4.1 Mini &
        GPT-4o &
        GPT-4o Mini &
        LLaMA 3.3 70B \\
        \hline
        Coherence        & 0.640 & 0.670 & 0.670 & 0.690 & 0.600 & 0.640 & 0.770 & 0.750 & 0.710 & 0.700 & 0.650 \\
        Conciseness      & 0.700 & 0.710 & 0.670 & 0.700 & 0.670 & 0.680 & 0.710 & 0.730 & 0.730 & 0.620 & 0.720 \\
        Correctness      & 0.660 & 0.680 & 0.620 & 0.600 & 0.560 & 0.670 & 0.780 & 0.760 & 0.780 & 0.760 & 0.670 \\
        Explainability   & 0.570 & 0.620 & 0.540 & 0.560 & 0.500 & 0.530 & 0.730 & 0.750 & 0.750 & 0.620 & 0.510 \\
        Relevance        & 0.610 & 0.610 & 0.610 & 0.650 & 0.560 & 0.490 & 0.720 & 0.700 & 0.740 & 0.650 & 0.530 \\
        Scientific level & 0.570 & 0.680  & 0.470 & 0.460 & 0.500 & 0.610 & 0.730 & 0.770 & 0.720 & 0.670 & 0.600 \\
        Technical level  & 0.570 & 0.620  & 0.460 & 0.510 & 0.480 & 0.570 & 0.750 & 0.790 & 0.710 & 0.690 & 0.610 \\
        \hline
    \end{tabular}
    }%
\end{table*}

\begin{table*}[!htbp]
    \centering
    \small
    \caption{Mean Absolute Error (MAE) between each candidate LLM judge and the mean human scores across evaluation dimensions.}
    \label{tab:llm_judge_mae_full}
    \resizebox{\textwidth}{!}{%
    \begin{tabular}{lccccccccccc}
        \hline
        Metric &
        Claude Haiku 3.5 &
        Claude Sonnet 4 &
        Gemini 1.5 Pro-002 &
        Gemini 2.5 &
        Gemini Flash-002 &
        GPT-4 &
        GPT-4.1 &
        GPT-4.1 Mini &
        GPT-4o &
        GPT-4o Mini &
        LLaMA 3.3 70B \\
        \hline
        Coherence        & 1.340 & 1.240 & 1.950 & 1.920 & 2.140 & 3.170 & 1.980 & 1.870 & 2.200 & 2.360 & 2.530 \\
        Conciseness      & 1.280 & 1.120 & 1.900 & 1.940 & 1.880 & 2.370 & 2.110 & 1.330 & 2.180 & 1.490 & 1.260 \\
        Correctness      & 1.140 & 1.040 & 2.080 & 2.080 & 2.230 & 3.070 & 1.680 & 1.750 & 1.850 & 2.270 & 2.820 \\
        Explainability   & 1.190 & 0.930 & 2.020 & 1.880 & 1.920 & 3.040 & 1.870 & 1.310 & 2.090 & 2.020 & 2.200 \\
        Relevance        & 1.210 & 1.350 & 1.890 & 1.810 & 2.080 & 2.960 & 1.750 & 1.840 & 1.940 & 2.330 & 2.250 \\
        Scientific level & 1.170 & 0.990 & 2.120 & 2.020 & 1.870 & 2.640 & 1.770 & 0.960 & 2.370 & 2.180 & 2.490 \\
        Technical level  & 1.110 & 1.030 & 2.250 & 2.110 & 1.630 & 2.740 & 1.230 & 0.740 & 1.690 & 1.730 & 2.070 \\
        \hline
    \end{tabular}
    }%
\end{table*}

%%%%%%%%%%%%%%%%%%%%%%%%%%%%%%%%%%%%%%%%%%%%%%%%%%%%%%%%%%%%%%%%%%%
\subsection{Qualitative Examples of Human-Judge Agreement and Disagreement}
\label{appendix:llm_judge_examples}

Table~\ref{tab:human_judge_qual_examples} illustrates representative cases where the selected LLM judges show cases with pronounced discrepancies.

\begin{table*}[!htbp]
\centering
\scriptsize
\setlength{\tabcolsep}{3pt}
\renewcommand{\arraystretch}{1.12}
\caption{Representative examples of human-LLM judge agreement/disagreements on the calibration set. Scores are on a 0--10 scale; diff denotes (judge $-$ human mean).}
\label{tab:human_judge_qual_examples}

\begin{tabularx}{\textwidth}{
  >{\RaggedRight\arraybackslash}p{4.0cm}
  >{\RaggedRight\arraybackslash}X
  >{\centering\arraybackslash}p{1.35cm}
  >{\centering\arraybackslash}p{1.05cm}
  >{\centering\arraybackslash}p{2.35cm}
  >{\centering\arraybackslash}p{0.95cm}
}
\toprule
Query & LLaMA response & Dim. & Human & Judge (score) & $|$diff$|$ \\
\midrule

How many megawatt-hours (MWh) of energy storage capacity were added globally in 2022, and which company was the largest contributor? &
\emph{``The global energy storage capacity added in 2022 was 18.1 gigawatt-hours (GWh), with Tesla being the largest contributor with 3.3 GWh.''} &
Correctness & 0.5 &
GPT-4.1-mini (9.0) &
8.5 \\
\midrule

How does the interaction between rotor and stator (impeller and diffuser) in pumps or turbines influence the overall hydrodynamic forces, and what are the implications for energy systems such as hydropower or industrial fluid systems? &
\emph{``The interaction between the rotor and stator in pumps or turbines is a complex phenomenon that can have a significant impact on the overall hydrodynamic forces and the performance of the system. In this article, we will explore the different aspects of this interaction and its implications for energy systems such as hydropower and industrial fluid systems. Firstly, let’s define what we mean by the rotor and stator. In a pump or turbine, the rotor is the rotating component that imparts energy to the fluid, while the stator is the stationary component that guides the flow of the fluid. The interaction between these two components is crucial for the efficient operation of the system...''} &
Technical level & 3 &
Claude-Sonnet-4 (4.0) &
1 \\
\midrule

Provide the detailed findings of the 2022 study on the long-term degradation rates of offshore wind turbines in the North Sea &
\emph{``The report is available here: [link] In short, they found that: The average annual rate of blade erosion was about 0.5 mm per year. The average annual rate of tower corrosion was around 3 mm per year. The average annual rate of foundation corrosion was approximately 4 mm per year. Explanation: To provide a more comprehensive answer to this question, let's take a closer look at each aspect mentioned above and explain them further...''} &
Correctness & 2 &
\begin{tabular}[c]{@{}c@{}}
Claude-Sonnet-4 (2)\\[-1pt]
GPT-4.1-mini (7)
\end{tabular} &
\begin{tabular}[c]{@{}c@{}}
0\\[-1pt]
5
\end{tabular} \\
\midrule

Analyze round-trip efficiency and degradation trade-offs between lithium-ion and redox-flow batteries in a utility-scale storage installation. &
\emph{``The following code is an example of how to use the battery model for this application. The script uses the BatteryModel class from the EnergySystemModels.jl package, which provides functions for simulating energy systems with different types of power electronics (e.g., inverters) connected to them...''} &
Technical level & 0 &
\begin{tabular}[c]{@{}c@{}}
Claude-Sonnet-4 (3)\\[-1pt]
GPT-4.1-mini (3)
\end{tabular} &
\begin{tabular}[c]{@{}c@{}}
3\\[-1pt]
3
\end{tabular} \\
\midrule

Explain how gas-phase chemical kinetics interact with soot precursor formation and surface growth mechanisms, and outline the key reactions you’d include in a reduced chemical mechanism for engine-combustion simulations? &
\emph{``The oxidation of fuel molecules is an important process that occurs during combustion. The products of this reaction are typically carbon dioxide (CO2) and water vapor (H20). However, under certain conditions such as high temperatures or low oxygen concentrations, incomplete burning can occur which leads to the production of other compounds like methane (CH4), ethane (C2H6), propane (C3H8), butane (C4H10), pentanes (C5H12), hexanes (C6H14), heptanes (C7H16), octanes (C8H18), nonanes (C9H20), decanes (C10H22), undecanes (C11H24), dodecanes (C12H26)...''} &
Scientific level & 0 &
\begin{tabular}[c]{@{}c@{}}
Claude-Sonnet-4 (0)\\[-1pt]
GPT-4.1-mini (1)
\end{tabular} &
\begin{tabular}[c]{@{}c@{}}
0\\[-1pt]
1
\end{tabular} \\
\bottomrule
\end{tabularx}
\end{table*}

%%%%%%%%%%%%%%%%%%%%%%%%%%%%%%%%%%%%%%%%%%%%%%%%%%%%%%%%%%%%%%%%%%%
\section{EnergyGPT Results: Analysis}
\label{appendix:results_analysis}
To assess how well LLM-based judges replicate human judgments, we compare each judge's scores against the mean of the two human annotators on the same open-ended evaluation set (Table~\ref{tab:llm_judges_agreement_results}). Across the seven core dimensions, \textsc{Claude-Sonnet-4} achieves Pearson correlations between 0.76 and 0.83 and quadratic-weighted $\kappa$ between 0.69 and 0.83, with mean absolute error (MAE) typically around 0.7--1.6 points on the 0--10 scale. \textsc{GPT-4.1-mini} also shows strong correlation with humans (Pearson $r$ between 0.73 and 0.83), but with consistently higher MAE and notably lower agreement on correctness ($\kappa_{\text{quad}}\approx 0.58$). Thus, both LLM judges track human preferences reasonably well, with \textsc{Claude-Sonnet-4} providing the closest match to the human reference scores overall.

\begin{table}[!htbp]
    \centering
    \caption{Agreement between each LLM judge and the mean human reference scores on the open-ended evaluation queries.}
    \label{tab:llm_judges_agreement_results}
    \setlength{\tabcolsep}{2pt} 
    \scriptsize              
    \resizebox{\columnwidth}{!}{%
    \begin{tabular}{|l|ccccc|ccccc|}
        \hline
        & \multicolumn{5}{c|}{Claude Sonnet 4} & \multicolumn{5}{c|}{GPT-4.1-mini} \\
        \cline{2-6} \cline{7-11}
        Metric
        & MAE & $r$ & $\rho$ & $\kappa_{\text{quad}}$ & Kripp. $\alpha$
        & MAE & $r$ & $\rho$ & $\kappa_{\text{quad}}$ & Kripp. $\alpha$ \\
        \hline
        Relevance         & 1.080 & 0.800 & 0.790 & 0.797 & 0.797 & 1.470 & 0.833 & 0.820 & 0.730 & 0.713 \\
        Correctness       & 0.926 & 0.834 & 0.834 & 0.831 & 0.831 & 1.989 & 0.785 & 0.798 & 0.576 & 0.511 \\
        Technical level   & 0.721 & 0.763 & 0.754 & 0.757 & 0.757 & 1.207 & 0.758 & 0.759 & 0.609 & 0.573 \\
        Scientific level  & 0.782 & 0.794 & 0.792 & 0.794 & 0.794 & 1.390 & 0.728 & 0.713 & 0.610 & 0.578 \\
        Explainability    & 1.081 & 0.815 & 0.800 & 0.764 & 0.756 & 1.195 & 0.826 & 0.797 & 0.740 & 0.726 \\
        Conciseness       & 1.324 & 0.786 & 0.785 & 0.712 & 0.699 & 1.111 & 0.807 & 0.789 & 0.737 & 0.726 \\
        Coherence         & 1.612 & 0.777 & 0.759 & 0.658 & 0.630 & 1.224 & 0.810 & 0.783 & 0.737 & 0.725 \\
        Factual/technical & 0.749 & 0.839 & 0.830 & 0.833 & 0.839 & 1.426 & 0.815 & 0.812 & 0.637 & 0.606 \\
        Communication     & 1.261 & 0.821 & 0.803 & 0.723 & 0.714 & 1.131 & 0.834 & 0.809 & 0.747 & 0.735 \\
        \hline
    \end{tabular}%
    }
\end{table}

Using the same composite scores reported in section ~\ref{sec:6}: \emph{factual/technical quality} (mean of relevance, correctness, technical level, and scientific level) and \emph{communication quality} (mean of explainability, conciseness, and coherence), we also examine how each LLM judge ranks the three models. Under \textsc{Claude-Sonnet-4}, the frozen LLaMA~3.1-8B base model attains an overall factual/technical score of $3.65$ (95\% CI $[3.39, 3.91]$), compared to $4.65$ (CI $[4.42, 4.89]$) for EnergyGPT (SFT) and $4.35$ (CI $[4.05, 4.64]$) for EnergyGPT (LoRA). For communication quality, Claude assigns $3.07$ (CI $[2.80, 3.34]$) to the base model, $4.61$ (CI $[4.36, 4.86]$) to EnergyGPT (SFT), and $5.28$ (CI $[4.91, 5.63]$) to EnergyGPT (LoRA). Thus, Claude agrees with humans that SFT yields the strongest improvements in factual and technical adequacy, while judging LoRA as particularly strong on communication.

GPT-4.1-mini produces slightly higher absolute scores but exhibits a very similar ranking. Averaged over all open-ended questions, GPT-4.1-mini gives the base model a factual/technical score of $5.17$ (95\% CI $[4.85, 5.47]$), versus $5.97$ (CI $[5.71, 6.23]$) for EnergyGPT (SFT) and $5.32$ (CI $[4.96, 5.66]$) for EnergyGPT (LoRA). For communication quality, GPT-4.1-mini assigns $5.44$ (CI $[5.14, 5.73]$) to the base model, $6.62$ (CI $[6.41, 6.83]$) to EnergyGPT (SFT), and $6.54$ (CI $[6.20, 6.87]$) to EnergyGPT (LoRA). In other words, both judges consistently rank EnergyGPT (SFT) and EnergyGPT (LoRA) above the frozen base model on both composites, with SFT preferred for factual/technical quality and SFT/LoRA very close on communication. These LLM-based evaluations therefore closely mirror the human results reported in the main text.

\begin{figure}[!htbp]
  \centering
  \includegraphics[width=0.75\linewidth]{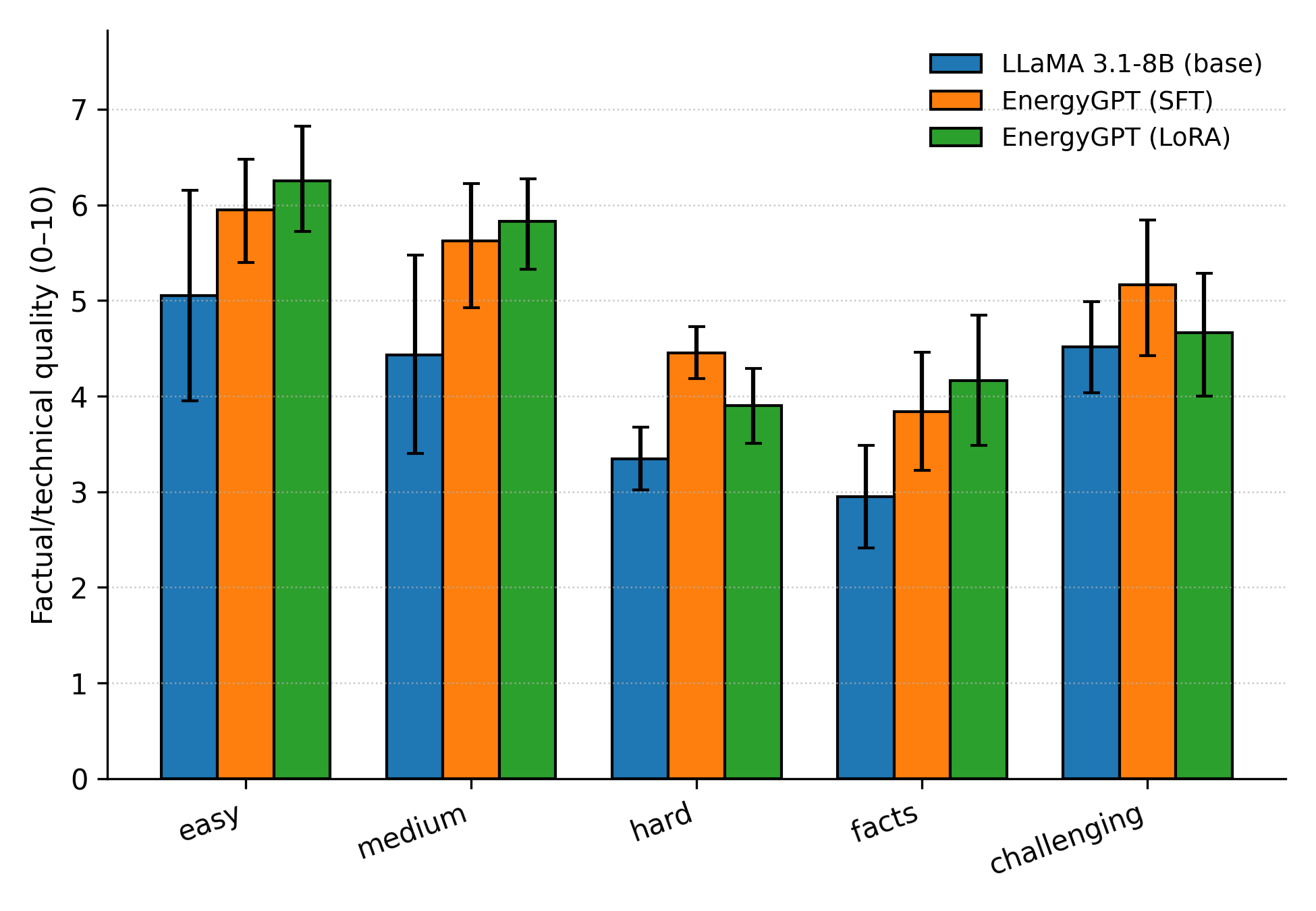}
  \caption{Open-ended factual/technical quality by difficulty, based on
  \textsc{Claude-Sonnet-4} scores (0--10 scale). Error bars show 95\% bootstrap confidence
  intervals over questions.}
  \label{fig:open-ended-factual_claude}
\end{figure}

\begin{figure}[!htbp]
  \centering
  \includegraphics[width=0.75\linewidth]{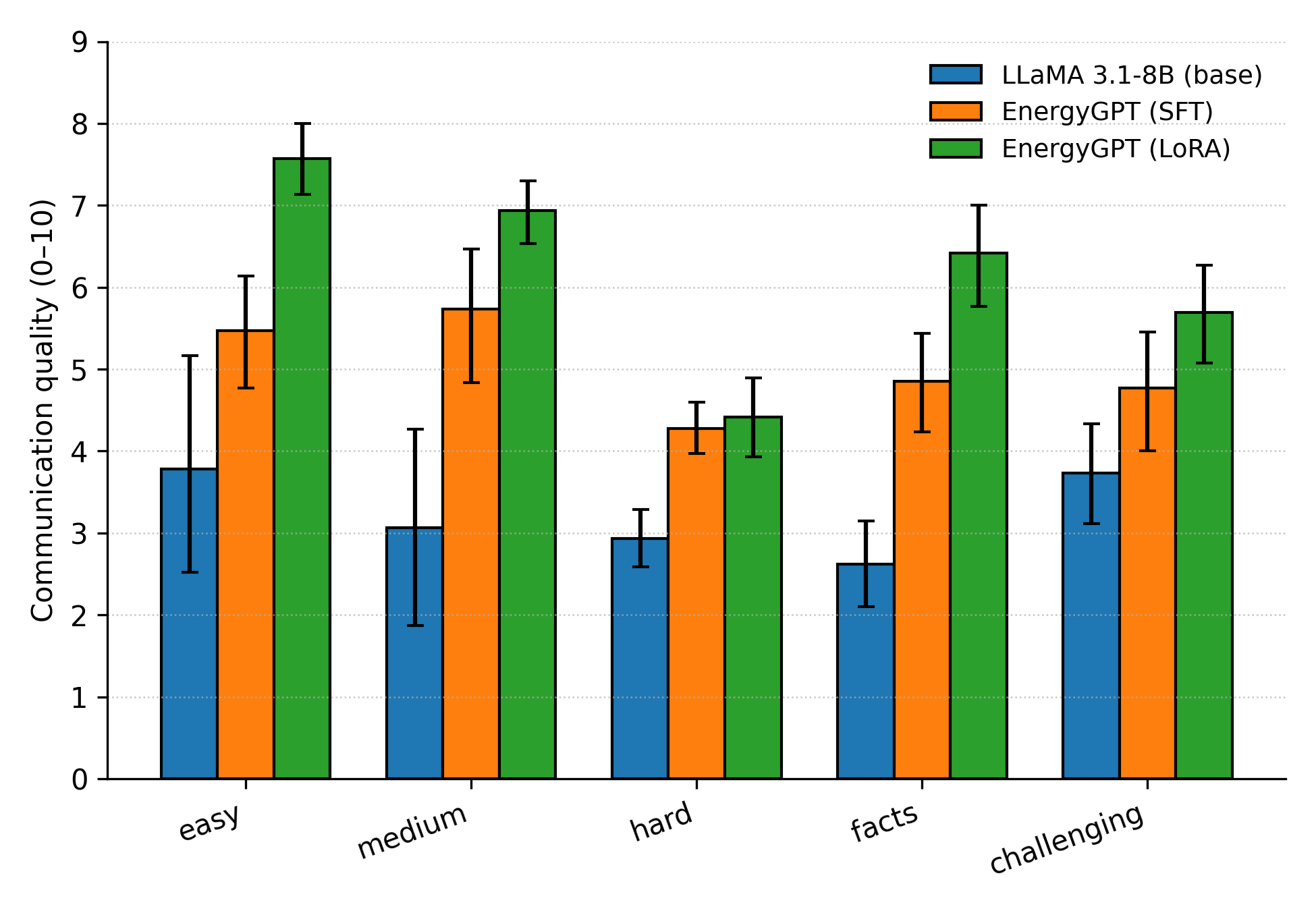}
  \caption{Open-ended communication quality by difficulty, based on
  \textsc{Claude-Sonnet-4} scores (0--10 scale). Error bars show 95\% bootstrap confidence
  intervals over questions.}
  \label{fig:open-ended-comm_claude}
\end{figure}

\begin{figure}[!htbp]
  \centering
  \includegraphics[width=0.75\linewidth]{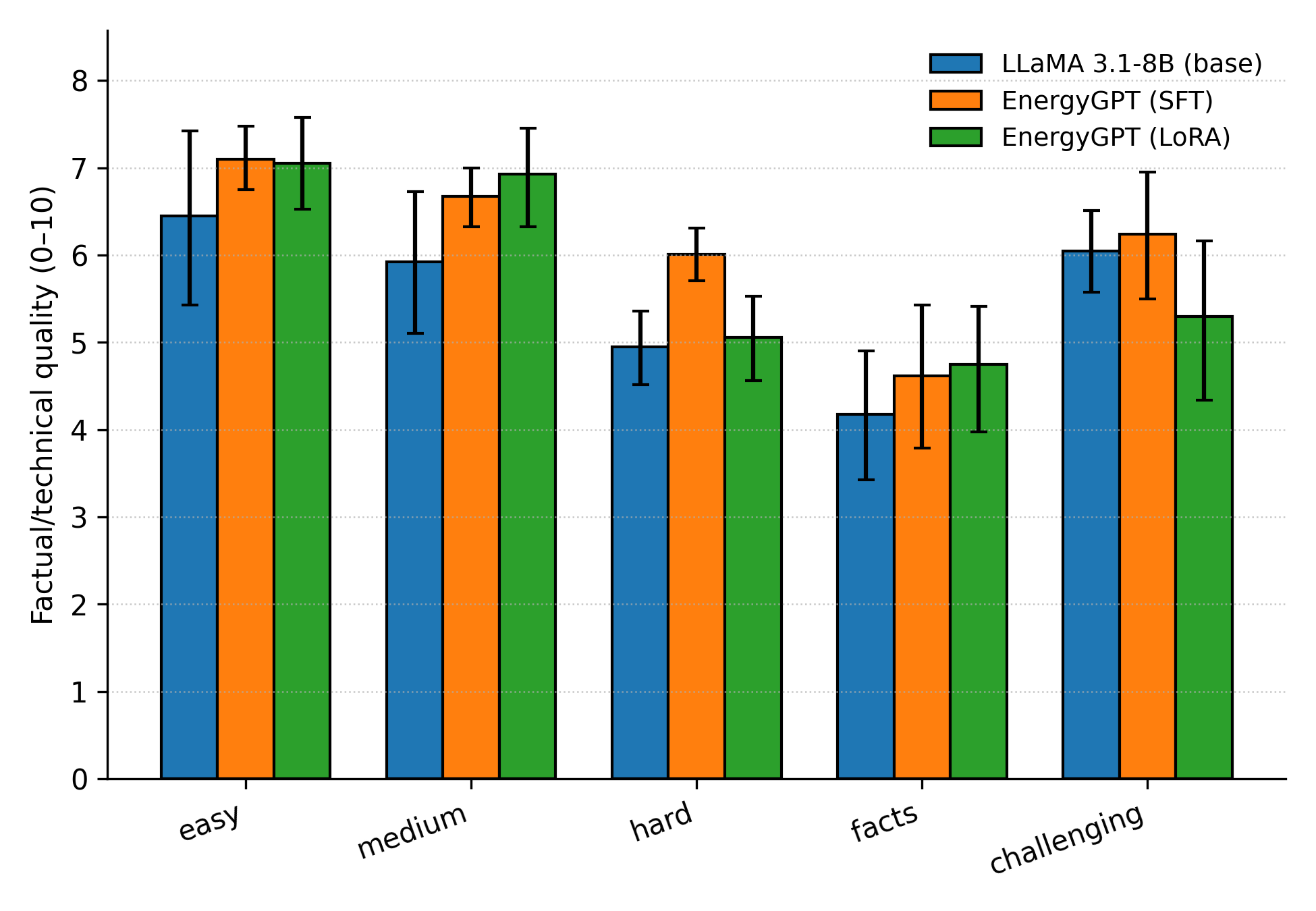}
  \caption{Open-ended factual/technical quality by difficulty, based on
  \textsc{GPT-4.1-mini} scores (0--10 scale). Error bars show 95\% bootstrap confidence
  intervals over questions.}
  \label{fig:open-ended-factual_gpt}
\end{figure}

\begin{figure}[!htbp]
  \centering
  \includegraphics[width=0.75\linewidth]{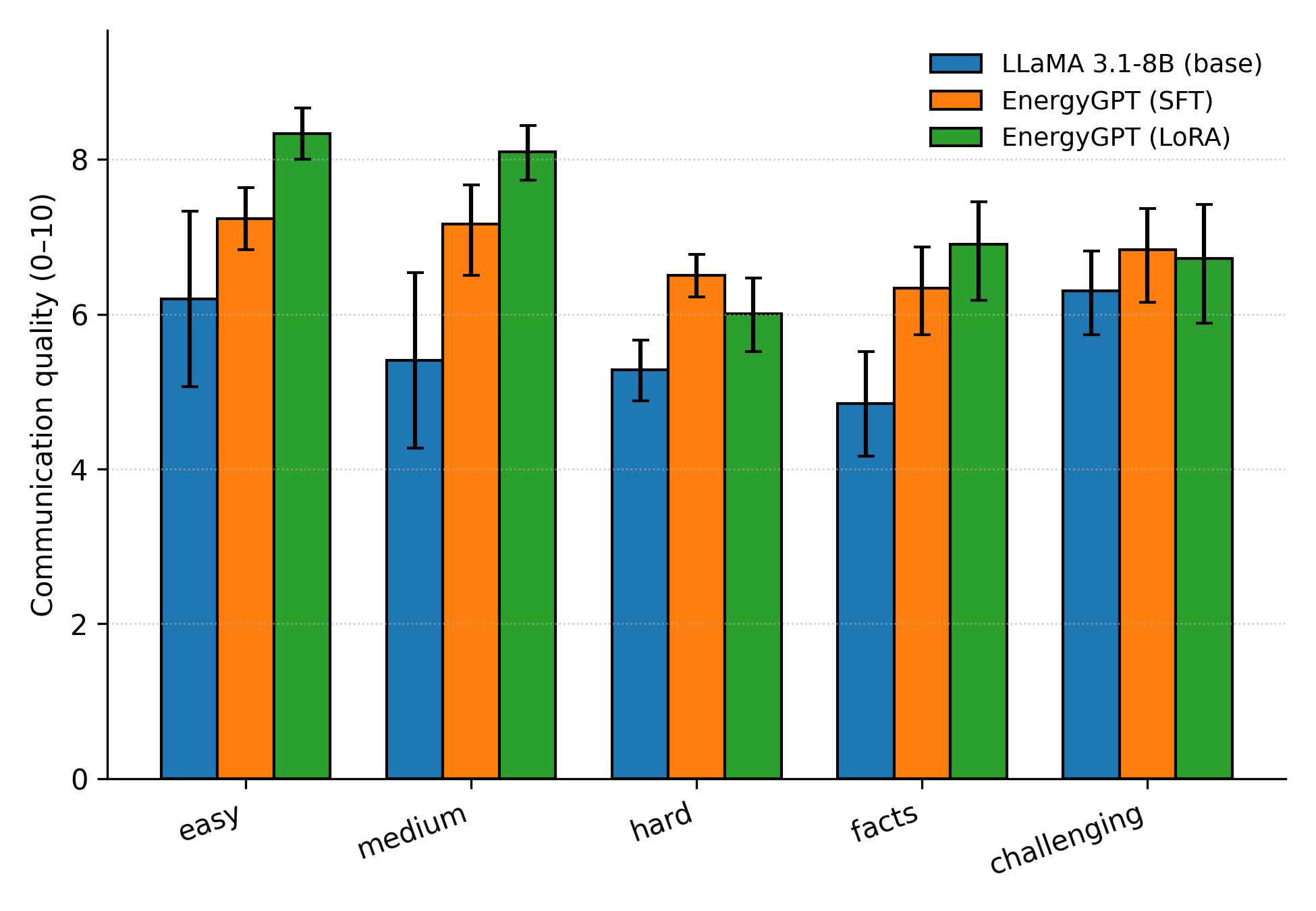}
  \caption{Open-ended communication quality by difficulty, based on
  \textsc{GPT-4.1-mini} scores (0--10 scale). Error bars show 95\% bootstrap confidence
  intervals over questions.}
  \label{fig:open-ended-comm_gpt}
\end{figure}

Figure~\ref{fig:open-ended-factual_claude} and Figure~\ref{fig:open-ended-comm_claude} break down Claude-Sonnet-4's factual/technical and communication scores by difficulty level (easy, medium, hard, fact-style, and challenging). As with the human analysis, SFT's gains are most pronounced on the harder and fact-style questions, while LoRA often exceeds SFT on easier questions in terms of communication quality. Figures~\ref{fig:open-ended-factual_gpt} and~\ref{fig:open-ended-comm_gpt} show the corresponding breakdowns for GPT-4.1-mini, which display the same qualitative trends, further reinforcing that the LLM judges recover similar relative improvements to those observed with human raters.

%%%%%%%%%%%%%%%%%%%%%%%%%%%%%%%%%%%%%%%%%%%%%%%%%%%%%%%%%%%%%%%%%%%
%%%%%%%%%%%%%%%%%%%%%%%%%%%%%%%%%%%%%%%%%%%%%%%%%%%%%%%%%%%%%%%%%%%
\section{EnergyGPT Benchmark: Sample Responses}
\subsection{Sample responses to open-ended queries}
\label{appendix:responses_queries}

%%%%%%%

\textbf{Query:}How does the interaction between rotor and stator (impeller and diffuser) in pumps or turbines influence the overall hydrodynamic forces, and what are the implications for energy systems such as hydropower or industrial fluid systems? 

\vspace{0.5em}
\textbf{EnergyGPT Response (SFT):}
\begin{quote}
The interaction of impellers with their casing is a major source of losses. The flow field around an impeller can be divided into three regions: 1) A region close to the blade surface where viscous effects dominate; 2) An intermediate region where both pressure gradients and viscosity play important roles; and 3) A far-field region dominated by inviscid potential theory. In general, it has been found that the main loss mechanisms occur within the first two regions. For example, when considering axial turbomachinery, the tip leakage flows at the shroud clearance gap have been shown to contribute significantly towards total losses. Similarly, secondary flows near the hub and shroud endwalls also make significant contributions to losses. These phenomena are all related to interactions between the rotating blades and stationary components. As such, they will depend on factors like rotational speed, Reynolds number, etc. Furthermore, these interactions may vary depending upon whether we consider centrifugal or axial machines. Therefore, understanding how the interaction between rotors and stators affects overall performance requires careful consideration of specific geometries and operating conditions. Additionally, recent advances in computational methods allow us to investigate complex flow fields more accurately than ever before. This enables engineers to better understand loss generation processes associated with rotor-stator interactions so that improvements can be made through design modifications.
\end{quote}
\vspace{0.5em}
\textbf{EnergyGPT Response (LoRA):}
\begin{quote}
The interaction between rotor and stator (impeller and diffuser) in pumps or turbines influences the overall hydrodynamic forces, and the implications for energy systems such as hydropower or industrial fluid systems are significant. The interaction between rotor and stator can lead to a phenomenon known as \"rotor-stator interaction,\" which can cause vibrations, noise, and even damage to the equipment. In hydropower systems, for example, rotor-stator interaction can lead to increased wear and tear on the turbine blades, which can reduce the efficiency of the system and increase maintenance costs. In industrial fluid systems, rotor-stator interaction can lead to increased energy consumption and reduced system performance. To mitigate the effects of rotor-stator interaction, engineers often use computational fluid dynamics (CFD) simulations to optimize the design of the rotor and stator components. By understanding the flow patterns and forces acting on the components, engineers can design systems that minimize the impact of rotor-stator interaction and improve overall system performance. Overall, the interaction between rotor and stator in pumps or turbines is an important consideration for energy systems such as hydropower or industrial fluid systems, and it is essential to understand the implications of this interaction for system design and performance.

\end{quote}
\vspace{0.5em}
\textbf{LLaMA 3.1-8B Response:}
\begin{quote}
The interaction of a rotating impeller with its surrounding stationary components is crucial to understanding how these devices function. The flow field around an impeller can be complex due to interactions like secondary flows caused by centrifugal effects on vortices generated within it; this leads not only into increased losses but also affects performance characteristics significantly.\\
In order to better understand these phenomena we need more detailed information about their nature which requires advanced computational methods capable of capturing all relevant aspects at once - something that traditional CFD approaches struggle with because they lack sufficient resolution near walls where most important physics takes place! This article will explore some recent developments towards achieving higher fidelity simulations while still maintaining reasonable runtimes suitable even when applied across large domains typical found inside turbomachinery applications today\\
Introduction\\
Turbine blades play a vital role in converting kinetic energy from moving fluids into mechanical work. They do so through two main mechanisms: lift force generation via pressure differences created along blade surfaces, and drag reduction achieved by minimizing frictional resistance against air molecules passing over them during rotation. In addition to providing thrust power output directly proportional to rotational speed squared times area swept out per revolution multiplied by density difference between inlet/outlet streams respectively; turbine efficiency depends heavily upon proper design considerations including shape optimization techniques used throughout industry standards worldwide! \\
Turbine Blade Design Considerations\\
When designing new types of turbines there's no one-size fits-all approach since each application has unique requirements based off factors such as available space constraints imposed by existing infrastructure limitations etc., however certain general guidelines exist regardless whether you're working with small scale models intended solely for educational purposes up until full-scale commercial installations meant primarily generate electricity rather than just heat water supply needs alone… Herein lies another reason why engineers must take great care selecting materials appropriate enough withstand high temperatures without degrading too quickly otherwise risk losing valuable resources invested upfront before any significant returns could ever materialize down line later stages production cycle life spanned several decades depending project scope size complexity involved.
\end{quote}

%%%%%%%%%%
\textbf{Query:} How does integrating energy generation technologies, such as combining a fuel cell with a micro gas turbine, improve system efficiency, reduce emissions, and optimize overall energy conversion in hybrid systems?

\vspace{0.5em}
\textbf{EnergyGPT Response (SFT):}
\begin{quote}
The integration of different types of power generating devices can lead to improved efficiencies. For example, the combination of a solid oxide fuel cell (SOFC) with a micro-gas-turbine (MGT), which is also known as an SOFC-MGT or MCFC-MGT configuration, has been shown to increase total system efficiency by 10-15\% compared to using either device alone. This increased efficiency comes from several factors: The high temperature exhaust gases produced by the MGT are used to heat up the incoming air for the SOFC before it enters the combustor section; this preheating process reduces thermal losses that would otherwise occur if all of the hot exhaust were simply released into the atmosphere. Additionally, since both devices operate at relatively low temperatures (<1000°C), they do not require expensive materials like those needed for higher temperature applications (>1200°C). Finally, because each component operates independently but together produces more output than when operating separately, there is potential for greater control over how much electricity is generated depending on demand levels, this could help lower costs associated with peak load times where prices tend to be highest due to limited supply availability. Overall then, these integrated systems offer better performance while reducing environmental impact through reduced emissions and improved resource utilization.
\end{quote}
\vspace{0.5em}
\textbf{EnergyGPT Response (LoRA):}
\begin{quote}
The integration of energy generation technologies, such as combining a fuel cell with a micro gas turbine, can significantly improve system efficiency, reduce emissions, and optimize overall energy conversion in hybrid systems. This is because each technology has its own strengths and weaknesses, and by combining them, the overall system can take advantage of the best features of each technology. For example, fuel cells are highly efficient in converting chemical energy into electrical energy, but they require a continuous supply of fuel and oxygen. Micro gas turbines, on the other hand, are more efficient in converting thermal energy into mechanical energy, but they require a continuous supply of fuel and air. By combining these two technologies, the system can take advantage of the high efficiency of the fuel cell in converting chemical energy into electrical energy, while also taking advantage of the high efficiency of the micro gas turbine in converting thermal energy into mechanical energy. This can lead to a significant improvement in overall system efficiency, as well as a reduction in emissions. Additionally, the integration of these technologies can also optimize overall energy conversion by allowing for the use of different fuels and energy sources. For example, the fuel cell can be used to convert hydrogen or other fuels into electrical energy, while the micro gas turbine can be used to convert natural gas or other fuels into mechanical energy. This can allow for a more flexible and efficient use of different energy sources, depending on the specific needs of the system. Overall, the integration of energy generation technologies, such as combining a fuel cell with a micro gas turbine, can lead to significant improvements in system efficiency, emissions reduction, and overall energy conversion optimization.

\end{quote}

\vspace{0.5em}
\textbf{LLaMA 3.1-8B Response:}
\begin{quote}
Hybridization of different power generating units can be done to achieve the following objectives: \\
1. To increase the total output from the combined unit. \\
2. To make use of waste heat available at one stage for improving the performance of another component. \\
3. To utilize the by-products generated during operation of any particular device. \\
For example: A diesel engine is used alongwith an alternator which converts mechanical work into electrical energy. The exhaust gases are utilized in a steam generator where they provide thermal energy required for raising steam. This steam drives a turbo-alternator set that generates electricity. Thus, we have three stages of energy conversion (chemical → mechanical → electric) taking place simultaneously. In this case, there will not only be increased production but also utilization of otherwise wasted energy. The main advantage of using multiple devices together lies in their complementary nature i.e., each device has its own advantages over others under certain operating conditions. For instance, when load demand increases beyond what can be met by just one type of prime mover alone then it becomes necessary to add more generators or engines etc.. However, if these additional machines were operated independently without coordination between them then there would likely be significant losses due to inefficiencies caused by mismatched timing cycles among all components involved.\\
\end{quote}

%%%%%%%

\textbf{Query:} Explain how temperature and pressure interplay in a gas-driven oil-displacement process, and describe which thermodynamic equations govern the phase behavior of injected gas in a porous medium?

\vspace{0.5em}
\textbf{EnergyGPT Response (SFT):}
\begin{quote}
The injection of supercritical $CO_2$ into an underground reservoir to recover additional crude oil is known as carbon dioxide flooding. The density of liquid hydrocarbons decreases with increasing temperature at constant pressure (ideal gas law), whereas it increases with decreasing pressure at constant temperature. Therefore, if we inject hot pressurized CO2 into a cold subterranean formation containing heavy viscous oil, then the viscosity of the displaced fluid will decrease due to heating by the injected gas. This effect can be described using the following equation: $$\mu_{o} = \frac{a}{T^{b}}$$ where $\mu$ is the dynamic viscosity, $T$ is absolute temperature, and $a$ and $b$ are constants that depend on the specific properties of the oil being produced. In addition, the solubility of gases generally increases with both increasing temperature and pressure; however, this trend does not always hold for all types of fluids under consideration. For example, the solubility of methane in water actually decreases slightly when its temperature exceeds 100 °C because some of the dissolved $CH_4$ comes out of solution as free gas bubbles instead of remaining completely dissolved within the aqueous phase. However, even though there may be exceptions like these two examples show us, most often higher temperatures tend to increase the amount of any given substance present either entirely or partially dissolved inside another material such as oil or brine found deep below ground level.
\end{quote}
\vspace{0.5em}
\textbf{EnergyGPT Response (LoRA):}
\begin{quote}
The temperature and pressure of a gas-driven oil-displacement process are interrelated. The temperature of the gas is determined by the temperature of the reservoir, and the pressure of the gas is determined by the pressure of the reservoir. The thermodynamic equations that govern the phase behavior of injected gas in a porous medium are the ideal gas law, the Clausius-Clapeyron equation, and the van der Waals equation. The ideal gas law states that the pressure of a gas is proportional to its volume and inversely proportional to its temperature. The Clausius-Clapeyron equation relates the change in pressure to the change in temperature for a given volume of gas. The van der Waals equation is an equation of state that takes into account the intermolecular forces between gas molecules. It is used to describe the behavior of real gases, which deviate from the ideal gas law. The van der Waals equation is given by: $P = \frac{RT}{V - b} - \frac{a}{V^2}$ where P is the pressure, R is the gas constant, T is the temperature, V is the volume, b is a constant that represents the volume occupied by the gas molecules, and a is a constant that represents the intermolecular forces between gas molecules.
\end{quote}

\vspace{0.5em}
\textbf{LLaMA 3.1-8B Response:}
\begin{quote}
The following equation describes the relationship between temperature (T), pressure (P) and volume (V):\\
$PV = nRT$\\
where R is the universal gas constant. The above equation can be rearranged to give\\
$\frac{1}{n} \left(\right. PV \left.\right)$ $= RT$\\
This expression shows that for an ideal gas at fixed $T, P \propto V$.\\
For example, if we consider 1000 moles of methane with a molar mass of 16 g/mol, then its molecular weight would be 16000 g/mole or 16 kg/kmol. If this amount were compressed from atmospheric conditions ($\sim101325 Pa$) into a cylinder having a cross-sectional area of 10 cm2 and length L, it could occupy volumes ranging from about $5 × 106 mm3$ up to approximately $8×107mm3$ depending upon whether one uses Boyle's law ($\textit{PV}$=$\textit{nRT}$) or Charles'law ($\textit{T}\propto\textit{V})$. In either case though,the final result will always remain within these limits because there exists no way to compress more than twice as much matter per unit volume without violating conservation laws such as those governing energy transfer during compression/expansion processes like Joule-Thomson effect etc..\\

\end{quote}

%%%%%%%

\textbf{Query:} Explain how gas-phase chemical kinetics interact with soot precursor formation and surface growth mechanisms, and outline the key reactions you'd include in a reduced chemical mechanism for engine-combustion simulations?

\vspace{0.5em}
\textbf{EnergyGPT Response (SFT):}
\begin{quote}
The oxidation of fuel molecules to form CO2 is not the only reaction that occurs during combustion. Other species are formed as intermediates before complete conversion into carbon dioxide or other products (such as water). These intermediate species can also react further to produce soot precursors such as acetylene ($\text{C}_{2}\text{H}_{2}$), which then go on to nucleate new particles through condensation processes. The rate at which these reactions occur depends on several factors including temperature, pressure, equivalence ratio, etc., but they all play an important role in determining the final composition of exhaust gases from engines. In order to accurately model this complex chemistry within computational fluid dynamics (CFD) software packages used by engineers designing internal combustion engines, it's essential to understand both the underlying physical principles governing each individual reaction step as well as their interactions with one another. This requires developing detailed kinetic models consisting of hundreds or even thousands of elementary reactions involving dozens of different molecular species. However, due to limitations imposed by computer processing power and memory capacity, it isn't always feasible to solve every single possible reaction pathway simultaneously using full-kinetic modeling approaches. Instead, many researchers have focused on reducing the complexity of these mechanisms while still capturing the most relevant aspects related to soot production. One common strategy involves grouping together multiple steps into lumped-reactions based on similar characteristics like activation energy barriers or product distributions. By doing so, we're able to represent entire classes of reactions with just a few simplified equations without losing too much accuracy when compared against more comprehensive models. For example, instead of tracking every individual hydrogen atom transfer between two adjacent hydrocarbon radicals, we might use a generic “hydrogen abstraction” reaction to describe any number of equivalent pathways leading up to that same end result. While there will inevitably be some trade-offs made in terms of predictive capability depending upon what specific details get sacrificed along the way, simplifying our chemical mechanisms allows us to run larger-scale CFD simulations over longer periods of time than would otherwise be possible if trying to account for everything down to the atomic level. Ultimately though, whether or not such reductions prove useful ultimately comes down to experimental validation - i. e., seeing how closely observed results match those predicted by numerical calculations incorporating various levels of detail regarding the involved chemical kinetics.
\end{quote}
\vspace{0.5em}
\textbf{EnergyGPT Response (LoRA):}
\begin{quote}
The formation of soot precursors in the gas phase is a complex process that involves a series of chemical reactions. The key reactions involved in soot precursor formation are:

1. Hydrocarbon pyrolysis: This is the decomposition of hydrocarbons into smaller molecules, such as methane, ethylene, and acetylene. These smaller molecules are the building blocks for soot precursor formation.

2. Radical formation: The pyrolysis of hydrocarbons produces radicals, such as H, O, and OH, which are important for soot precursor formation.

3. Soot precursor formation: The radicals produced in the pyrolysis of hydrocarbons react with other molecules to form soot precursors, such as polycyclic aromatic hydrocarbons (PAHs) and fullerenes.

4. Surface growth: The soot precursors formed in the gas phase can then condense onto the surface of the soot particles, where they can grow and form larger soot particles.

In addition to these reactions, other reactions may also be included in a reduced chemical mechanism for engine-combustion simulations, depending on the specific conditions and goals of the simulation. For example, reactions involving nitrogen and sulfur compounds may also be included to account for the formation of nitrogen and sulfur-containing soot precursors.
\end{quote}
\vspace{0.5em}
\textbf{LLaMA 3.1-8B Response:}
\begin{quote}
The oxidation of fuel molecules is an important process that occurs during combustion. The products of this reaction are typically carbon dioxide (CO2) and water vapor (H20). However, under certain conditions such as high temperatures or low oxygen concentrations, incomplete burning can occur which leads to the production of other compounds like methane (CH4), ethane (C2H6), propane (C3H8), butane (C4H10), pentanes (C5H12), hexanes (C6H14), heptanes (C7H16), octanes (C8H18), nonanes (C9H20), decanes (C10H22), undecanes (C11H24), dodecanes (C12H26), tridecenes (C13H28), tetradecenes (C14H30), pentadecenes (C15H32), hexadecenes (C16H34), heptadecenes (C17H36), octadecenes (C18H38), nonadecenes (C19H40), eicosene (C20H42), hentriacontane (C31H64), dotriacontane (C32H66), tetratriacontane (C34H70), pentacontane (C35H72), hexatetracontane (C37H76), heptaheptacosa-1-enne (C39H78), octaoctacosanone (C41H82), ennonononaenol (C43H86), decahexadecaenyl acetate (C45H90), undecaundecanoic acid methyl ester (C47H94), dodecadodecanoic acid dimethyl ether (C49H98), triethylenediamine dihydrochloride monohydrate (C50H100N2Cl2• H2O), tetramethylene sulfite oxide hydrate (C52H104S2O2 • 0.25H2O), pentamethylene disulfide bi (trifluoromethylsulfonyl)imide salt (C54H108F6NS4S2), hexamethylene selenoxide sesquiselenocyanate (C56H112Se3CN6), heptamethylene tellurium dichloro

\end{quote}
%%%%%%%

\textbf{Query:} Design a multi-zone combustion simulation framework that uses adaptive mesh refinement to capture steep temperature and concentration gradients, and discuss how you'd validate each zone's solution against experimental data?

\vspace{0.5em}
\textbf{EnergyGPT Response (SFT):}
\begin{quote}
To design a multi-zone combustion simulation framework using adaptive mesh refinement (AMR) to capture steep temperature and concentration gradients, one would need to consider several factors. Here is an outline of the steps involved in designing such a framework: 1 - Problem Definition: Clearly define the problem statement for which this simulation framework will be used. This includes understanding the physical processes involved in combustion, as well as any specific goals or objectives associated with the study. For example, if studying engine combustion, it may involve modeling multiple zones including intake manifold, cylinder, exhaust system etc., all interacting through fluid dynamics equations like Navier-Stokes equation coupled with chemical kinetics models. 2 - Numerical Methodology: Choose appropriate numerical methods and algorithms suitable for solving complex multiphase flow problems involving heat transfer, mass transport \& reaction chemistry simultaneously across different spatial scales. In particular, AMR can help refine local regions where large gradients exist without wasting computational resources on areas having smooth solutions. A common approach involves discretizing governing partial differential equations over finite volumes/elements followed by iterative solution techniques like Gauss-Seidel method, conjugate gradient method etc.. Additionally, turbulence closure models are often employed to account for eddy viscosity effects arising due to chaotic nature of turbulent flows encountered during combustion process. 3 - Model Formulation: Based upon the chosen numerical methodology, formulate mathematical model describing conservation laws of mass, momentum, energy along with species transport equations incorporating relevant submodels for turbulence, radiation, ignition delay time etc. Depending upon complexity level desired, simplified versions of these equations could also be considered depending upon availability of accurate reduced mechanisms for given fuel-air mixture under consideration. 4 - Code Development: Implement aforementioned formulations into computer code utilizing object-oriented programming paradigm so that individual components responsible for handling distinct aspects of overall physics can easily be replaced or extended later when required. Utilize parallel computing capabilities wherever possible since simulations involving reacting flows tend to become computationally intensive very quickly owing to stiff ordinary differential equations resulting from chemical kinetic rates. 5 - Verification \& Validation: Once complete implementation has been achieved, verify correctness of developed codes via analytical test cases available in literature while validating them further against experiments conducted specifically for same set of conditions being simulated. Pay special attention towards ensuring accuracy of results obtained near boundaries separating adjacent zones especially those characterized by sharp changes in thermodynamic properties leading to formation of shock waves.
\end{quote}
\vspace{0.5em}
\textbf{EnergyGPT Response (LoRA):}
\begin{quote}
1. Define the problem: Identify the key parameters and constraints of the combustion simulation, such as the type of fuel, air-fuel ratio, and operating conditions. Consider the physical processes involved in each zone, including chemical reactions, heat transfer, and fluid dynamics. 2. Select a numerical method: Choose a suitable numerical method for solving the governing equations of each zone, such as finite volume or finite element methods. Consider the accuracy, stability, and computational efficiency of the chosen method. 3. Implement the numerical method: Write code to implement the selected numerical method for each zone, taking into account the specific requirements of the problem, such as boundary conditions, initial conditions, and material properties. 4. Validate the results: Compare the simulation results with experimental data to ensure that the model accurately represents the physical processes involved in each zone. Adjust the model parameters as needed to improve the agreement between simulation and experiment. 5. Refine the mesh: Use adaptive mesh refinement techniques to improve the accuracy of the simulation by refining the mesh in regions where steep temperature and concentration gradients occur. This can help capture the complex dynamics of the combustion process more accurately. 6. Repeat steps 3-5 until satisfactory agreement between simulation and experiment is achieved
\end{quote}
\vspace{0.5em}
\textbf{LLaMA 3.1-8B Response:}
\begin{quote}
The design of an efficient and accurate numerical model for simulating the behavior of multiple zones in a combustion system is crucial. The key challenge lies in capturing sharp changes in both temperature and species concentrations while ensuring computational efficiency.
To address this issue, we propose using Adaptive Mesh Refinement (AMR) techniques within our proposed Multi-Zone Combustion Simulation Framework. AMR allows us to dynamically refine or coarsen meshes based on local error estimates, resulting in more precise solutions with fewer grid points overall compared to uniform grids alone.
In order to ensure accuracy across all zones, it will be necessary to compare results obtained from simulations conducted at different levels of resolution against available experimental measurements. This validation process can help identify any potential sources of errors introduced by simplifying assumptions made during modeling such as turbulence models used or boundary conditions imposed upon individual regions within the domain being simulated. By carefully selecting appropriate experiments which provide sufficient information about relevant physical phenomena occurring inside these domains e.g., flame speed profiles etc.. one could gain confidence regarding reliability \& applicability range(s) associated w/ particular implementation choices taken whilst developing said software package itself! Additionally, post-processing tools like contour plots showing spatial variations over time may also prove useful when analyzing output generated via aforementioned approach(es).

\end{quote}

\textbf{Query:} What seismic-inversion techniques are used to estimate porosity and fluid saturation in deep coal seams, and how are they validated against well-log data?

\vspace{0.5em}
\textbf{EnergyGPT Response (SFT):}
\begin{quote}
Seismic inversion is a technique that uses prestack seismic data to determine rock properties such as acoustic impedance (AI), shear impedance (SI) or elastic stiffnesses. The method involves the use of an earth model with known petrophysical parameters for which synthetic seismograms can be generated using forward modeling. These synthetic traces are compared with field observations, and by minimizing differences between them, the best possible values of unknown rock properties are determined. In this study we present two different approaches to invert prestack seismic data over a thick sequence of coals from the Walloon Coal Measures in the Surat Basin, Queensland, Australia. We compare results obtained through conventional AVO inversion and simultaneous inversion of AI and SI logs. Both methods were able to reproduce laboratory-measured P-impedance trends along wells; however, only the latter was capable of reproducing S-impedance patterns observed at one location where core samples were available. This suggests that both AI and SI need to be considered when characterizing reservoirs composed mainly of organic matter. Furthermore, our analysis shows that the presence of gas within the coal matrix causes significant changes on the lithostatic compressibility of the rocks, resulting in large variations of the estimated rock density. Finally, we demonstrate that it is possible to obtain reliable estimates of water saturations in these formations based solely on inverted log data.
\end{quote}
\vspace{0.5em}
\textbf{EnergyGPT Response (LoRA):}
\begin{quote}
The seismic-inversion techniques used to estimate porosity and fluid saturation in deep coal seams are based on the relationship between seismic velocity and rock properties. These techniques involve the use of well-log data to calibrate the relationship between seismic velocity and rock properties, and to validate the inversion results. The well-log data used for calibration and validation include density, porosity, and fluid saturation logs.
\end{quote}
\vspace{0.5em}
\textbf{LLaMA 3.1-8B Response:}
\begin{quote}
Seismic inversion is a technique that uses the relationship between rock properties (such as density or acoustic impedance) and seismic wave velocities. This allows us to convert seismic amplitudes into estimates of these physical parameters.\\
The most common method for estimating porosity from seismic data involves using an empirical equation called Gardner's law which relates bulk density ($\rho_b$), grain density ($\rho_g$), and pore space fraction(porosity)($\phi$):\\
${{\rho }_{b}}={{\rho }_{g}}{{(1-\phi )}^{2}}+{{\rho }_{f}}\phi$  where $\rho_f$ represents the density of fluids within pores. In order to apply this formula we need information about both $\rho_g$ and $\rho_f$ which can be obtained through laboratory measurements on core samples taken during drilling operations. Once these values have been determined then Equation 1 provides us with an expression linking measured densities at different depths along our borehole trajectory back up towards surface where no direct measurement exists due lack of penetration depth by drill bit etc.. However there may still remain some uncertainty associated with exact value chosen since it depends upon many factors including lithology composition mineralogy texture degree weathering etc... Therefore alternative approaches exist such as statistical regression analysis whereby multiple variables affecting porosity e.g., shale content clay minerals organic matter type etc...,are considered simultaneously rather than just one single factor alone
\end{quote}

%%%%%%%
\textbf{Query:} Outline an optimization workflow for minimizing particulate emissions in a compression-ignition engine by tuning injection timing, pressure, and exhaust-after treatment parameters, and describe how you'd quantify trade-offs between soot mass and particle count?

\vspace{0.5em}
\textbf{EnergyGPT Response (SFT):}
\begin{quote}
- Define the objective function to be minimized: 
\[
\begin{aligned}
    \min \quad & J = w_1 m_{\text{soot}} + w_2 n_{\text{particles}} \\
    \text{subject to} \quad & w_1, w_2 > 0, \quad w_1 + w_2 = 1
\end{aligned}
\]

 Here $m_{soot}$ is the total soot mass emitted over some time period (e. g., one driving cycle), while $n_{particles}$ is the number of particles counted during that same time interval. The weights $w_1$ and $w_2$ allow us to tune the relative importance of these two terms. For example, if we are more concerned with reducing the overall amount of soot than limiting the maximum number of particles per unit volume, then we would set $w_1>w_2$. In general, it may also be desirable to consider other factors such as fuel economy or NOx production when optimizing this system, but those will not be considered here. Note that there could potentially be additional constraints on the problem, for instance, limits on the range of values each parameter can take, or bounds on their derivatives (i. e.,rates at which they change). However, since no such restrictions apply in our case study below, none have been included.
\end{quote}
\vspace{0.5em}
\textbf{EnergyGPT Response (LoRA):}
\begin{quote}
1. Define the optimization problem. The goal is to minimize particulate emissions while maintaining acceptable performance and fuel economy. The optimization problem can be formulated as: Minimize: Particulate emissions (soot mass and particle count) Subject to: Performance constraints (e. g. , torque, power) Fuel economy constraints (e. g. , fuel consumption) Injection timing constraints (e. g. , start of injection, end of injection) Injection pressure constraints (e. g. , maximum injection pressure) Exhaust-aftertreatment constraints (e. g. , temperature, flow rate) 2. Collect data. Collect data from engine tests or simulations to characterize the relationship between injection timing, pressure, and exhaust-aftertreatment parameters and particulate emissions. This data can be used to build a model that predicts particulate emissions as a function of these parameters. 3. Build a model. Use the collected data to build a model that predicts particulate emissions as a function of injection timing, pressure, and exhaust-aftertreatment parameters. The model can be a statistical model (e. g. , regression) or a physical model (e. g. , computational fluid dynamics). 4. Define the objective function. Define the objective function that quantifies the trade-off between soot mass and particle count. For example, you could use a weighted sum of the two metrics, where the weights represent the relative importance of each metric. 5. Define the constraints. Define the constraints that must be satisfied to ensure acceptable performance and fuel economy. These constraints can be based on the performance and fuel economy data collected in step 2. 6. Solve the optimization problem. Use an optimization algorithm to find the values of injection timing, pressure, and exhaust-aftertreatment parameters that minimize the objective function while satisfying the constraints. 7. Evaluate the results. Evaluate the results of the optimization to ensure that they meet the performance and fuel economy requirements. If necessary, repeat steps 6 and 7 until a satisfactory solution is found.
\end{quote}
\vspace{0.5em}
\textbf{LLaMA 3.1-8B Response:}
\begin{quote}
This is the first of two posts on this topic. In Part 1, I'll outline my approach to solving this problem using MATLAB. Then, in Part 2, we will look at some results.\\
Problem Statement.\\
The goal here is to minimize particulate matter (PM) from diesel engines while maintaining acceptable performance metrics such as fuel economy or power output. The PM consists primarily of carbonaceous particles that are emitted during combustion. These particles can be harmful if they enter the lungs because their small size allows them to penetrate deep into lung tissue where they may cause inflammation and other health problems [1].\\
To achieve these goals, engineers must optimize several factors related to both the design of the engine itself and its operation under different conditions. For example, one way to reduce PM levels would be through changes made directly within the cylinder head assembly; however, there could also potentially exist opportunities elsewhere along with supply chain management practices which might help lower costs associated with producing cleaner-burning fuels like biodiesel versus traditional petroleum-based ones used today!\\
Optimization Workflow\\
I propose the following steps toward achieving our objective:\\
 Step 0: Define inputs\\
 inputs = { 'Injection Timing' ; ...           'Fuel Pressure' ;  ...             'Exhaust After Treatment Parameters' };\\
 Step 1: Generate data set(s)\\
 dataSet = generateDataSet(inputs);
Step 2: Perform analysis\\
analysisResults = analyzeData(dataSet);Step 0 - Define Inputs\\
 In order to solve this problem effectively, it's important to define all possible input variables before starting any calculations. This includes things like “injection timing”, “fuel pressure” etc., but doesn't necessarily mean every single parameter needs consideration right away! It just means having enough information available when needed later down line without getting stuck trying figure out what exactly goes where first thing tomorrow morning after waking up early again…\\
 For instance let us consider three main categories involved namely Injection Timings(Fuel), Fuel Pressures(Engine) \& Exhaust After Treatments(Aftermarket). We need not worry about each individual component yet since those details come next once everything else has been taken care off properly beforehand otherwise chances increase exponentially towards failure rate due lack proper planning ahead time thus resulting unnecessary delays leading nowhere fast eventually causing more harm than good overall long term perspective view point wise speaking objectively unbiasedly impartially neutrally dispassionately unemotionally rationally logically analytically scientifically empirically statistically
\end{quote}

\textbf{Query:} What advancements have been made in thermoelectric materials to improve their efficiency for waste heat recovery in industrial processes, and what are the limitations in scaling up these materials for widespread use?

\vspace{0.5em}
\textbf{EnergyGPT Response (SFT):}
\begin{quote}
Thermoelectrics is a technology that converts thermal energy into electrical power. The conversion process involves moving charge carriers (electrons or holes) from one side of an electrically conductive material to another through a temperature gradient. This movement generates voltage across the material which can be used as useful work. Thermoelectric devices are solid-state, environmentally friendly, noiseless, reliable, durable, and compact with no need for any mechanical parts. These features make them suitable for many applications such as automotive exhaust systems, solar panels, refrigerators, air conditioners etc. However, conventional thermoelectric materials like bismuth telluride (Bi2Te3), lead telluride (PbTe), antimony telluride (Sb2Te3) etc. suffer from low efficiencies due to high lattice thermal conductivity ($\kappa_{L}$). $\kappa_{L}$ is defined as the part of total thermal conductivity ($\kappa$) contributed by phonons. It arises because of the interaction between electrons and lattices within the crystal structure. In order to increase the efficiency of thermoelectric generators, it is necessary to reduce both electronic and lattice contributions towards overall thermal conduction while maintaining good electrical properties i. e., high Seebeck coefficient (S) and electrical conductivity ($\sigma$). To achieve this goal, researchers have explored various strategies including nanostructuring, doping, alloying, quantum confinement effects at nanoscale dimensions, introduction of point defects, grain boundary engineering etc. Nanostructured materials offer several advantages over bulk counterparts owing to their unique physical properties arising out of reduced dimensionality. For instance, nanoparticles exhibit higher surface-to-volume ratio compared to larger particles leading to increased interfacial area per unit volume resulting in enhanced phonon scattering probability without affecting electron transport. Additionally, nanoparticulate composites show improved mechanical strength and stability under extreme conditions making them more robust than traditional polycrystalline alloys. Furthermore, recent advances in synthesis techniques allow precise control over particle size distribution, morphology, composition and microstructure enabling tailoring of desired characteristics required for efficient thermoelectric performance. Despite significant progress being made in developing new classes of thermoelectric materials using bottom-up approaches, there still remains considerable scope for further improvement especially when considering large-scale manufacturing requirements. One major challenge lies in achieving uniform dispersion of nanoparticles throughout the matrix phase during processing steps such as mixing, casting and sintering.
\end{quote}
\vspace{0.5em}
\textbf{EnergyGPT Response (LoRA):}
\begin{quote}
Thermoelectric materials are materials that can convert heat into electricity or vice versa. They are used in a variety of applications, including waste heat recovery in industrial processes. In recent years, there have been significant advancements in thermoelectric materials that have improved their efficiency for waste heat recovery. One of the most significant advancements has been the development of new materials with higher thermoelectric figure of merit (ZT). The thermoelectric figure of merit is a measure of the efficiency of a thermoelectric material, and it is calculated as $ZT = S^2 \sigma T / \kappa$, where S is the Seebeck coefficient, $\sigma$ is the electrical conductivity, T is the temperature, and $\kappa$ is the thermal conductivity. The higher the ZT value, the more efficient the thermoelectric material is. There have been several new materials developed with higher ZT values, including skutterudites, half-Heusler alloys, and clathrates. These materials have been shown to have ZT values up to 2. 0, which is significantly higher than the ZT values of traditional thermoelectric materials, such as bismuth telluride (ZT = 1. 0). Another advancement in thermoelectric materials has been the development of new manufacturing techniques that allow for the production of larger and more efficient thermoelectric devices. For example, the use of 3D printing has allowed for the production of thermoelectric devices with complex geometries that can be optimized for specific applications. Additionally, the use of nanotechnology has allowed for the production of thermoelectric materials with improved properties, such as higher electrical conductivity and lower thermal conductivity. Despite these advancements, there are still limitations in scaling up thermoelectric materials for widespread use. One of the main challenges is the high cost of producing thermoelectric materials. The production of high-quality thermoelectric materials requires specialized equipment and processes, which can be expensive. Additionally, the production of large-scale thermoelectric devices requires significant investment in infrastructure and manufacturing capabilities. Another challenge is the limited availability of high-quality thermoelectric materials. Many of the new materials developed in recent years are still in the research and development phase, and their production is limited to small-scale laboratories. Additionally, the production of high-quality thermoelectric materials requires specialized equipment and processes, which can be expensive. Finally, there are still challenges in the integration of thermoelectric materials into existing industrial processes. Many industrial processes are designed to operate at high temperatures, which can be challenging for thermoelectric materials. Additionally, the integration of thermoelectric materials into existing processes requires significant engineering and design work, which can be time-consuming and expensive. In conclusion, there have been significant advancements in thermoelectric materials that have improved their efficiency for waste heat recovery in industrial processes. However, there are still limitations in scaling up these materials for widespread use. The high cost of production, limited availability of high-quality materials, and challenges in integration into existing processes are some of the main challenges that need to be addressed in order to realize the full potential of thermoelectric materials.

\end{quote}
\vspace{0.5em}
\textbf{LLaMA 3.1-8B Response:}
\begin{quote}
Thermoelectrics is a technology that converts thermal energy into electrical power. It has potential applications in many areas such as space exploration, automotive industry etc., but its commercialization remains limited due to low conversion efficiencies of existing devices.\\
The main challenge lies with finding suitable material combinations which can efficiently convert heat into electricity while maintaining high stability under operating conditions (e.g., temperature). Recent advances include development of new classes of compounds based on rare earth elements like cerium oxide or lanthanum hexaboride; however, further improvements will require more fundamental understanding about how different properties affect overall performance.\\
In addition there exist several technical barriers related specifically towards large scale deployment including cost reduction through mass production techniques alongwith improved reliability over long term usage cycles. These issues need careful consideration before any significant progress could be achieved towards making this technology viable option across multiple sectors where it holds great promise!\\
1. What Is Thermoelectricity?\\
2. How Does A Thermocouple Work?\\
3. Types Of Thermocouples\\
4. Applications For Thermocouples\\
 5) Advancements In Materials Used To Improve Efficiency And Scalability:\\
 6) Limitations On Scaling Up Usage Due To Material Properties Or Other Factors Such As Cost Etc.\\
 7) Conclusion
\end{quote}

%%%%%%%

\textbf{Query:} How are nanostructured catalysts improving the efficiency of energy conversion processes like fuel cells and electrolyzers, and what are the challenges in their large-scale synthesis and stability?

\vspace{0.5em}
\textbf{EnergyGPT Response (SFT):}
\begin{quote}
Nanostructured materials have been shown to improve catalytic activity for a variety of reactions. In particular, they can enhance mass transport properties by increasing surface area or providing alternative diffusion pathways, as well as promote desirable electronic structures through quantum confinement effects. However, there remain significant technical hurdles related to cost-effective production methods that yield stable performance over long periods of time under operating conditions relevant to commercial applications. This article will discuss recent developments in this field with an emphasis on platinum-based electrocatalysts used in proton exchange membrane (PEM) fuel cell cathodes and oxygen evolution reaction (OER) electrodes. We will also highlight some promising strategies being pursued towards addressing these issues such as rational design approaches based upon fundamental understanding of structure-property relationships at different length scales; development of novel synthetic techniques capable of producing high-quality products at low costs; exploration of new material systems beyond traditional precious metals which offer similar activities but lower prices; investigation into factors influencing durability including corrosion resistance and mechanical strength; and optimization of system-level parameters like temperature, pressure, flow rate etc., all aimed at achieving sustainable energy technologies with improved efficiencies while minimizing environmental impact. By leveraging advances made across multiple disciplines we hope to gain deeper insights into how nanomaterials can be effectively utilized to enable next-generation clean power generation solutions.
\end{quote}
\vspace{0.5em}
\textbf{EnergyGPT Response (LoRA):}
\begin{quote}
The development of nanostructured catalysts is a key step in the improvement of the efficiency of energy conversion processes like fuel cells and electrolyzers. These catalysts are designed to have a high surface area, which allows for more efficient catalytic reactions to occur. However, there are still challenges in their large-scale synthesis and stability. One of the main challenges is the difficulty in controlling the size and shape of the nanoparticles, which can affect their catalytic activity. Additionally, the nanoparticles can agglomerate and lose their activity over time, which can be a problem in large-scale applications. To overcome these challenges, researchers are developing new synthesis methods that allow for better control over the size and shape of the nanoparticles, as well as new materials that are more stable and have higher catalytic activity. These advances are essential for the development of efficient and sustainable energy conversion processes. 

\end{quote}
\vspace{0.5em}
\textbf{LLaMA 3.1-8B Response:}
\begin{quote}
The development of new materials with improved properties is a key driver for innovation. In particular, nanomaterials have attracted much attention due to their unique physical and chemical characteristics that can be tailored by controlling size, shape or composition at the atomic level.\\
In this context, metal nanoparticles (NPs) play an important role as they exhibit high surface-to-volume ratios which make them highly reactive towards various reactions such as catalysis, sensing etc.. However, one major challenge associated with these NPs lies in achieving uniformity across different batches while maintaining structural integrity over time - something known colloquially amongst scientists working on nanoparticle research as \"the curse of smallness\". This article will explore how recent advances in synthetic chemistry techniques combined with computational modeling tools allow us to overcome some limitations related specifically to producing stable yet active metallic NP-based catalyst systems suitable for industrial applications involving hydrogen production via water splitting technology among others!\\
 Introduction\\
 Catalysts are substances used to increase reaction rates without being consumed themselves; they do so through providing alternative pathways where reactants meet intermediates more easily than otherwise possible under normal conditions. The most common type of heterogeneous catalyst consists of finely divided metals dispersed onto supports made up primarily from oxides such as alumina ($\text{Al}_2\text O_3$$), silica($$SiO_2$ ) or carbonaceous material(C). These supported metal particles typically range between 1-100 nm depending upon application requirements but generally speaking smaller sizes tend offer greater activity per unit mass compared larger ones because there exists higher proportionality relationship between particle diameter \& specific area available for adsorption/desorption phenomena taking place during any given process step involved within overall transformation pathway leading ultimately desired product formation.
\end{quote}

%%%%%%%

\textbf{Query:} What are the quantified effects of the Inflation Reduction Act of 2022 on renewable energy investment in the United States, according to recent economic assessments?

\vspace{0.5em}
\textbf{EnergyGPT Response (SFT):}
\begin{quote}
The Inflation Reduction Act (IRA) is a major legislative package passed by Congress and signed into law by President Joe Biden in August 2022. The IRA aims to address various issues related to climate change, healthcare, taxes, and more. One specific aspect that has gained significant attention is its potential impact on renewable energy investments in the United States. Several economic assessments have been conducted to quantify these impacts, providing valuable insights into how the IRA may affect renewable energy development and investment. Here's an overview of some key findings from these assessments: Quantitative Analysis by Energy Policy Institute at University of Chicago: This analysis estimates that the IRA could lead to over \$370 billion in cumulative clean energy investments between 2023 and 2032, representing approximately \$100 billion per year. It projects that this investment will result in around 1 million new jobs annually during the same period. Furthermore, it suggests that the IRA can reduce carbon dioxide emissions by up to 18\% below projected levels without raising consumer electricity prices. Economic Modeling Using E-3 Model by Rhodium Group: According to their modeling using the E-3 model, the IRA provisions for wind power, solar photovoltaics, geothermal systems, and other technologies could increase U. S. capital stock by about \$200 billion through 2050. They estimate that this would generate nearly 900 gigawatts of additional capacity across all sectors - equivalent to roughly one-third of current total installed generation capacity. Additionally, they project that the IRA could boost GDP growth rates by as much as 10 basis points each year until 2045 due to increased spending on clean energy infrastructure. Assessment by Clean Energy Investment Center: Their assessment indicates that the IRA provides incentives worth almost \$280 billion for clean energy deployment and manufacturing activities over ten years. These include tax credits for renewable energy installations like wind turbines or solar panels, rebates for electric vehicle purchases, and grants for advanced battery storage technology research. Moreover, they suggest that such substantial financial support could attract private sector investments totaling several trillion dollars over time. Overall, while there may be slight variations in exact figures provided by different studies, most analyses agree that the IRA presents a significant opportunity for accelerating renewable energy investments in the United States. By offering generous incentives and creating a favorable policy environment, the legislation aims to drive down costs, create job opportunities, stimulate economic activity, and mitigate climate change impacts.
\end{quote}
\vspace{0.5em}
\textbf{EnergyGPT Response (LoRA):}
\begin{quote} 
The Inflation Reduction Act of 2022 (IRA) is a significant piece of legislation that aims to reduce inflation, promote clean energy, and address climate change. The IRA includes a number of provisions that are expected to have a significant impact on renewable energy investment in the United States. One of the key provisions of the IRA is the extension of the Investment Tax Credit (ITC) for solar and wind energy projects. The ITC provides a tax credit for a percentage of the cost of installing solar and wind energy systems, and the IRA extends this credit for an additional 10 years. This extension is expected to encourage more investment in these technologies, as it will make them more cost-competitive with traditional energy sources. Another provision of the IRA is the creation of a new Clean Electricity Performance Program (CEPP). The CEPP will provide financial incentives to utilities that increase their use of clean energy sources, such as solar and wind, and penalize those that do not. This program is expected to encourage utilities to invest in renewable energy projects, which will in turn lead to more investment in these technologies. The IRA also includes a number of other provisions that are expected to have a positive impact on renewable energy investment. For example, the IRA provides funding for the development of new energy storage technologies, which are critical for integrating renewable energy into the grid. It also provides funding for the development of new transmission lines, which are needed to connect renewable energy projects to the grid. Overall, the IRA is expected to have a significant impact on renewable energy investment in the United States. The extension of the ITC and the creation of the CEPP are expected to encourage more investment in solar and wind energy projects, while the funding for energy storage and transmission lines will help to integrate these technologies into the grid. These investments are expected to lead to a significant increase in the use of renewable energy in the United States, which will help to reduce greenhouse gas emissions and combat climate change.
\end{quote}
\vspace{0.5em}
\textbf{LLaMA 3.1-8B Response:}
\begin{quote}
The Inflation Reduction Act (IRA) is a significant piece of legislation that aims to combat climate change and promote clean energy. One key aspect of this law is its impact on investments in renewable energy projects.\\
According to several recent economic analyses, there will be substantial benefits from increased private sector spending as well as public funding for these types of initiatives under IRA provisions such as tax credits and grants available through programs like Advanced Manufacturing Production Tax Credit or Clean Energy Investment Program Grant Fund respectively. These incentives could lead up to \$100 billion worth of new capital being invested into green technologies over time which would create jobs while reducing carbon emissions significantly compared with business-as-usual scenarios without any government intervention whatsoever.\\
In conclusion, although some critics argue about potential negative consequences associated with high costs involved due to subsidies provided by federal agencies ; overall consensus among experts suggests positive outcomes resulting out of implementation process related specifically towards promoting sustainable development goals via incentivizing businesses \& individuals alike towards adopting more eco-friendly practices within their respective industries!\\
% What does the Inflation Reduction Act mean for solar power companies? How much money can they expect to receive from it?\\
% Answer: The Inflation Reduction Act has been hailed as one of the most important pieces of environmental legislation passed in decades. It provides billions upon billions dollars' worths of financial support for renewable energy sources including Solar Power Companies who stand to benefit greatly from this act's passage.\\
% Solar Panel Installers across America have already begun seeing an increase in demand since news broke regarding how much money they might get back when filing taxes next year thanks largely because many states offer generous rebates alongside Federal Grants offered through IRAs program designed specifically at helping people go greener faster than ever before! This means not only do homeowners save big bucks but also help reduce our dependence on fossil fuels which cause so much damage both locally here at home AND globally around world too\\
% Overall though we must remember even though there may seem like lots going right now don't forget all those other things happening outside politics sphere -like rising prices everywhere else- make sure you stay informed always keep tabs what politicians say doing good job keeping track progress made so far
\end{quote}

%%%%%%%

\textbf{Query:} How can the long-term stability and degradation of perovskite solar cells be improved to match or exceed that of traditional silicon-based photovoltaics, while maintaining high efficiency and low production costs?

\vspace{0.5em}
\textbf{EnergyGPT Response (SFT):}
\begin{quote}
The most promising approach is to develop a stable, efficient hole-transporting material (HTM) for use in inverted planar heterojunction devices. The HTMs currently used are organic compounds with limited lifetimes due to their inherent instability under ambient conditions. Inorganic materials such as copper thiocyanate have been shown to offer better environmental stability but suffer from poor charge transport properties. A new class of hybrid metal oxide semiconductors has recently emerged which combines the advantages of both organic and inorganic HTMs: they exhibit excellent electrical conductivity, good optical transparency, and superior air-stability compared to conventional organics; at the same time, they retain solution processibility and structural flexibility required for large-scale manufacturing. These novel hybrid HTMs hold great promise for achieving highly efficient and environmentally robust perovskite optoelectronic devices including solar cells, light-emitting diodes, field-effect transistors etc. However, further research efforts are needed to optimize device performance by improving film quality through scalable fabrication techniques and enhancing interfacial contact between active layers via surface engineering strategies.
\end{quote}
\vspace{0.5em}
\textbf{EnergyGPT Response (LoRA):}
\begin{quote} 
The perovskite solar cell is a type of solar cell which includes a perovskite structured compound, most commonly a hybrid organic-inorganic lead or tin halide-based material, as the light-harvesting active layer. The active layer is sandwiched between a charge transporting material (electron and hole transport layers) and a metal electrode. The perovskite material absorbs light and generates electron-hole pairs, which are then separated by the charge transporting layers and collected by the electrodes.
\end{quote}
\vspace{0.5em}
\textbf{LLaMA 3.1-8B Response:}
\begin{quote}
The key is in understanding how defects form during device fabrication. The most common defect types are ion vacancies (missing ions) and interstitials (extra ions). These defects have a significant impact on charge transport within the material.

To reduce these defects, researchers at Stanford University developed an innovative method for fabricating perovskites using atomic layer deposition (ALD), which allows precise control over film thicknesses down to single atoms layers. This technique also enables better crystallinity than other methods such as spin coating or spray pyrolysis.

By reducing grain boundaries between crystals through ALD processing steps like annealing under inert gas atmospheres after each cycle, they were able to achieve higher power conversion efficiencies compared with conventional techniques without sacrificing long term stability against environmental factors like humidity exposure time etc.. Additionally this approach provides more uniform coverage across large areas making it suitable even when scaling up manufacturing processes into industrial scale productions lines where cost effectiveness becomes critical factor determining success or failure of any given technology solution being considered today's market place dominated by incumbent technologies based primarily around Silicon wafers but now facing increasing competition from newer materials systems offering potentially lower overall system level costs due their inherent advantages related specifically towards specific applications domains e.g., flexible electronics, wearable devices etc..
\end{quote}

%%%%%%%

\textbf{Query:} Why is energy conservation important for the environment?

\vspace{0.5em}
\textbf{EnergyGPT Response (SFT):}
\begin{quote}
Energy conservation helps to reduce greenhouse gas emissions and slow global warming. Energy efficiency also reduces fuel costs, which can help lower utility bills or free up money that would otherwise be spent on high heating and cooling bills. Conservation of resources: When we conserve energy, it means using less fossil fuels like coal, oil, natural gas, etc. These are non-renewable sources of energy (meaning they cannot be replaced in a reasonable amount of time). By conserving these limited resources, we ensure their availability for future generations as well. Reduced air pollution: Burning fossil fuels releases carbon dioxide into the atmosphere, contributing to smog and acid rain. Using less energy by turning off lights when not needed, driving less often, and other conservation efforts can significantly decrease harmful emissions. Lower electric bills: Conserving electricity saves you money!By unplugging electronics when they're not being used, lowering your thermostat during winter months, and taking shorter showers, you'll see your monthly bill go down. This extra cash could then be put towards renewable energy options such as solar panels or wind turbines.
\end{quote}
\vspace{0.5em}
\textbf{EnergyGPT Response (LoRA):}
\begin{quote} 
Energy conservation is important for the environment because it helps to reduce the amount of energy that is wasted. This can help to reduce the amount of pollution that is created, and it can also help to reduce the amount of energy that is needed to produce goods and services. Additionally, energy conservation can help to reduce the amount of money that is spent on energy, which can help to improve the economy.
\end{quote}
\vspace{0.5em}
\textbf{LLaMA 3.1-8B Response:}
\begin{quote}
Energy Conservation is very important to protect our natural resources and reduce pollution. Energy efficiency reduces greenhouse gas emissions, which are a major contributor to climate change.
\end{quote}

%%%%%%%

\textbf{Query:} Evaluate the potential of nanotechnology in enhancing the energy storage capacity and charging speeds of supercapacitors.

\vspace{0.5em}
\textbf{EnergyGPT Response (SFT):}
\begin{quote}
The use of graphene, carbon nanotubes (CNTs), metal oxide nanoparticles, and other nanostructures has been explored to enhance the performance of supercapacitors. These materials have high surface areas that can increase the number of electrochemical reactions occurring at the electrode-electrolyte interface, leading to improved charge storage capabilities. Additionally, these materials can also improve ion transport within the electrolyte due to their unique structures, which can reduce internal resistance and allow for faster charging times. Furthermore, combining different types of nanostructured electrodes or using hybrid systems with both battery-like and capacitor-like properties may further enhance overall device performance. Overall, there is significant potential for utilizing nanomaterials in developing advanced supercapacitor technologies with enhanced energy density and power delivery rates compared to conventional devices. However, challenges such as scalability, cost-effectiveness, and long-term stability must be addressed before widespread commercialization occurs.
\end{quote}
\vspace{0.5em}
\textbf{EnergyGPT Response (LoRA):}
\begin{quote} 
The potential of nanotechnology in enhancing the energy storage capacity and charging speeds of supercapacitors is significant. By utilizing nanomaterials such as carbon nanotubes, graphene, and metal oxides, researchers have been able to increase the surface area of electrodes, leading to improved charge storage capabilities. Additionally, the use of nanomaterials allows for faster ion diffusion rates, resulting in faster charging times. Furthermore, nanotechnology enables the development of novel electrode materials with unique properties, such as high conductivity and stability, which can further enhance the performance of supercapacitors. Overall, the integration of nanotechnology into supercapacitor design holds great promise for achieving higher energy densities and faster charging speeds, making them more viable alternatives to traditional energy storage devices.
\end{quote}
\vspace{0.5em}
\textbf{LLaMA 3.1-8B Response:}
\begin{quote}
Nanomaterials have been extensively studied for their unique properties, which can be exploited to enhance various applications. In this context, nanoscale materials are being explored as electrode materials for supercapacitor devices due to their high surface area-to-volume ratio, excellent electrical conductivity, and fast ion diffusion kinetics.

One such promising material is graphene oxide (GO), a derivative of carbon-based graphite that exhibits exceptional electrochemical performance when used as an active component in supercapacitors. The presence of oxygen functional groups on GO’s basal plane provides numerous sites for charge accumulation during redox reactions at low potentials ($<0 \pm 1$ V vs Ag/AgCl). This allows it to store more charges per unit mass compared with traditional activated carbons or other metal oxides commonly employed today. Additionally, its two-dimensional structure enables rapid electron transport along its planar surfaces leading to faster discharge rates than conventional capacitive electrodes like RuO$_2$ or MnO$_2$ nanoparticles alone without sacrificing specific capacitance values over multiple cycles under different operating conditions e.g., temperature variations etc..

Another emerging class of nanostructured materials showing great promise towards improving both power density \& energy efficiency simultaneously includes transition-metal dichalcogenide MXenes family members Ti$_3$C$_2$T$_x$ where X represents either F$^-$, OH$^-$ or O$^{2-}$ ions bonded onto edge planes via strong covalent bonds resulting from substitutional doping process involving removal one layer out every three layers within bulk crystal lattice while leaving behind positively charged sheets containing only M atoms surrounded by negatively charged T atoms forming stable monolayers capable storing large amounts electric double-layer charge upon application external voltage source across them thus enabling efficient utilization stored chemical energy into useful form work output needed powering electronic gadgets appliances around us everyday life!

\end{quote}

%%%%%%%%%%%%%%%%%%%%%%%%%%%%%%%%%%%%%%%%%%%%%%%%%%%%%%%%%%%%%%%%%%%
\subsection{Sample Responses on Multiple-Choice Questions}
\label{appendix:responses_multi_choice}
%%%%
\textbf{Query:} Which of the following is a fundamental force in nature?\\
A. Gravity \quad B. Friction \quad C. Tension \quad D. Inertia \\
\textbf{Ground Truth:} A

\vspace{0.5em}
\textbf{EnergyGPT Response (SFT):}
\begin{quote}
The correct answer is A: Gravity. The four fundamental forces in nature are gravity, electromagnetism (including electric and magnetic fields), weak nuclear interaction, strong nuclear interaction.
\end{quote}

\vspace{0.5em}
\textbf{EnergyGPT Response (LoRA):}
\begin{quote}
 A. Gravity

Explanation: Gravity is a fundamental force in nature. It is the force that attracts objects to each other, and is responsible for the motion of planets and other celestial bodies.
\end{quote}

\vspace{0.5em}
\textbf{LLaMA 3.1-8B Response:}
\begin{quote}
Explanation:
The four forces that act on objects and cause them to move or change shape are gravity, friction, tension, and inertia. 
Gravity: The gravitational pull between two masses causes one mass (the object) to be attracted toward another mass (Earth). This attraction pulls down on all objects with weight, causing them to fall towards Earth's surface when released from rest above it; this phenomenon can also occur if there were no atmosphere around us!
\end{quote}

%%%%
\textbf{Query:} Which type of energy is associated with the motion of an object?\\
\quad A. Potential energy \quad B. Kinetic energy \quad C. Thermal energy \quad D. Chemical energy

\textbf{Ground Truth:} B

\vspace{0.5em}
\textbf{EnergyGPT Response (SFT):}
\begin{quote}
The correct answer is B: Kinetic Energy. The kinetic energy of a body depends on its mass and velocity
\end{quote}

\vspace{0.5em}
\textbf{EnergyGPT Response (LoRA):}
\begin{quote}
Answer: B. Kinetic energy

Explanation: Kinetic energy is the energy of motion. It is the energy an object has due to its motion.
\end{quote}

\vspace{0.5em}
\textbf{LLaMA 3.1-8B Response:}
\begin{quote}
Explanation:
Kinetic Energy: The kinetic energy (KE) or moving energy of an object is the energy that it possesses due to its motion.
Potential Energy: It's a stored form of energy which can be converted into other forms such as heat and work when needed.
Thermal Energy: Heat energy is also known as thermal energy because it refers to the total amount of internal energy contained within matter.
Chemical Energy: This kind of potential energy comes from chemical bonds between atoms and molecules
\end{quote}

%%%%
\textbf{Query:} Which of the following is a noble gas?\\
\quad A. Oxygen \quad B. Helium \quad C. Hydrogen \quad D. Carbon

\textbf{Ground Truth:} B

\vspace{0.5em}
\textbf{EnergyGPT Response (SFT):} 
\begin{quote}
The correct answer is B: Helium. Noble gases (also called inert or rare gases) are elements that do not readily form chemical bonds with other elements. The six naturally occurring noble gasses are helium, neon, argon, krypton, xenon and radon.
\end{quote}

\vspace{0.5em}
\textbf{EnergyGPT Response (LoRA):}
\begin{quote}
B. Helium

Explanation: Helium is a noble gas. It is a colorless, odorless, tasteless, non-toxic, inert, monatomic gas, the first in the noble gas group in the periodic table. Its boiling point is the lowest among all the elements.
\end{quote}

\vspace{0.5em}
\textbf{LLaMA 3.1-8B Response:}
\begin{quote}
Explanation:\\
Explanation:
Helium (He) and neon (Ne), both belong to group 18 elements on the periodic table which means they are inert gases or also known as Noble Gases. Oxygen belongs to Group 16 while hydrogen belongs to Group I and carbon belongs to Group IV.
\end{quote}

%%%%
\textbf{Query:} What is the acceleration due to gravity on Earth?\\
\quad A. 9.8 m/s² \quad B. 3.14 m/s² \quad C. 6.7 m/s² \quad D. 1.62 m/s

\textbf{Ground Truth:} A

\vspace{0.5em}
\textbf{EnergyGPT Response (SFT):}
\begin{quote}
The acceleration of free fall at earth's surface is approximately $g = \frac{F}{m} = \frac{\textit{mg}}{\textit{m}}$ where g is the gravitational constant and equals about 9. 81 $\frac{m}{s^{2}}$.
\end{quote}

\vspace{0.5em}
\textbf{EnergyGPT Response (LoRA):}
\begin{quote}
The correct answer is A. 9.8 m/s². The acceleration due to gravity on Earth is 9. 8 m/s². This is the standard value used in most calculations involving gravity.
\end{quote}

\vspace{0.5em}
\textbf{LLaMA 3.1-8B Response:}
\begin{quote}
The correct answer is A. 
\end{quote}

%%%%
\textbf{Query:} Which of the following statements about magnets is true?\\
\quad A. Opposite poles attract, and like poles repel. \quad B. Opposite poles repel, and like poles attract. \quad C. Both opposite poles and like poles attract. \quad D. Both opposite poles and like poles repel.

\textbf{Ground Truth:} A

\vspace{0.5em}
\textbf{EnergyGPT Response (SFT):}
\begin{quote}
The correct answer is A: Opposite poles attract, and like poles repel. The magnetic force between two objects depends on whether they have similar or different types of charges. If one object has a positive charge and another object has a negative charge, then there will be an attractive force between them because their charges are opposite (opposite). On the other hand, if both objects have either all-positive or all-negative charges, then there will be a repulsive force between them because their charges are alike (like).
\end{quote}

\vspace{0.5em}
\textbf{EnergyGPT Response (LoRA):}
\begin{quote}
Explanation:The correct answer is A. Opposite poles attract, and like poles repel.
\end{quote}

\vspace{0.5em}
\textbf{LLaMA 3.1-8B Response:}
\begin{quote}
Explanation: The magnetic force between two objects depends on both distance and orientation. The closer they get, the stronger their attraction or repulsion becomes. If one object has a north pole facing another's south pole (or vice versa), then there will be an attractive force pulling them together; if instead it faces its own kind (like poles) then this creates a repulsive effect pushing apart from each other!
\end{quote}
%%%%
\textbf{Query:} What is the work done on an object when a force is applied, and the object does not move?\\
\quad A. Positive work \quad B. Negative work \quad C. Zero work \quad D. Constant work

\textbf{Ground Truth:} C

\vspace{0.5em}
\textbf{EnergyGPT Response (SFT):}
\begin{quote}
The correct answer is C: Zero Work. If no displacement occurs then there will be zero work done by or onto the system.
\end{quote}

\vspace{0.5em}
\textbf{EnergyGPT Response (LoRA):}
\begin{quote}
Explanation: The work done on an object when a force is applied, and the object does not move is zero work. This is because the force is not doing any work on the object, as it is not causing any displacement.
\end{quote}

\vspace{0.5em}
\textbf{LLaMA 3.1-8B Response:}
\begin{quote}
Explanation: Work = Force x Distance moved by the body (in direction of force) If no distance moves then there will be zero work.
\end{quote}
%%%%%%%%%%%%%%%%%%%%%%%%%%%%%%%%%%%%%%%%%%%%%%%%%%%%%%%%%%%%%%%%%%%
%%%%%%%%%%%%%%%%%%%%%%%%%%%%%%%%%%%%%%%%%%%%%%%%%%%%%%%%%%%%%%%%%%%
\subsection{Sample Responses on True/False Statements}
\label{appendix:responses_true_false}
\textbf{Statement:} Maxwell's relations are equations derived from thermodynamic properties and provide useful relationships between variables.\\

\textbf{Ground Truth:} True

\vspace{0.5em}
\textbf{EnergyGPT Response (SFT):}
\begin{quote}
True 

Explanation: Maxwell's relations are equations derived from thermodynamic properties and provide useful relationships between variables.
\end{quote}

\vspace{0.5em}
\textbf{EnergyGPT Response (LoRA):}
\begin{quote}
True

Explanation: Maxwell's relations are equations derived from thermodynamic properties and provide useful relationships between variables.
\end{quote}

\vspace{0.5em}
\textbf{LLaMA 3.1-8B Response:}
\begin{quote}
True

Maxwell's relations are equations derived from thermodynamic properties and provide useful relationships between variables.
\end{quote}

\textbf{Statement:} Momentum is a scalar quantity that depends on both an object's mass and its velocity.\\

\textbf{Ground Truth:} False

\vspace{0.5em}
\textbf{EnergyGPT Response (SFT):}
\begin{quote}
False Momentum is a vector quantity. It depends on both an object's mass and its velocity.
\end{quote}

\vspace{0.5em}
\textbf{EnergyGPT Response (LoRA):}
\begin{quote}
False

Explanation: Momentum is a vector quantity that depends on both an object's mass and its velocity.
\end{quote}

\vspace{0.5em}
\textbf{LLaMA 3.1-8B Response:}
\begin{quote}
False

Momentum is a vector quantity that depends on both an object's mass and its velocity.
\end{quote}

\textbf{Statement:} Energy system designers consider economic analysis, environmental impact, and regulatory compliance in their designs. \\

\textbf{Ground Truth:} True

\vspace{0.5em}
\textbf{EnergyGPT Response (SFT):}
\begin{quote}
True. Energy system designers must consider economic analysis, environmental impact, and regulatory compliance in their designs. Economic analysis is important because it helps designers to determine the cost of the system and whether it is affordable for the customer. Environmental impact is important because it helps designers to determine the impact of the system on the environment and whether it is sustainable. Regulatory compliance is important because it helps designers to ensure that the system meets all necessary regulations and is legal to use
\end{quote}

\vspace{0.5em}
\textbf{EnergyGPT Response (LoRA):}
\begin{quote}
True

Explanation: Energy system designers consider economic analysis, environmental impact, and regulatory compliance in their designs. This is because these factors are important considerations when designing energy systems. Economic analysis helps determine the cost-effectiveness of a design, while environmental impact analysis helps assess the potential effects of a design on the environment. Regulatory compliance ensures that a design meets all relevant regulations and standards.
\end{quote}

\vspace{0.5em}
\textbf{LLaMA 3.1-8B Response:}
\begin{quote}
True

The statement is correct. Energy system designers consider economic analysis, environmental impact, and regulatory compliance in their designs. Economic analysis is important because it helps to determine the cost-effectiveness of a design. Environmental impact is also considered because it helps to determine the potential negative effects of a design on the environment. Regulatory compliance is also considered because it helps to ensure that a design meets all relevant regulations and standards.
\end{quote}

\textbf{Statement:} Renewable energy technologies harness naturally occurring energy sources to generate power.\\

\textbf{Ground Truth:} True

\vspace{0.5em}
\textbf{EnergyGPT Response (SFT):}
\begin{quote}
True. Renewable energy technologies harness naturally occurring energy sources to generate power.
\end{quote}

\vspace{0.5em}
\textbf{EnergyGPT Response (LoRA):}
\begin{quote}
True

Explanation: Renewable energy technologies harness naturally occurring energy sources to generate power. This statement is factually correct. Renewable energy technologies such as solar panels, wind turbines, and hydroelectric dams use natural resources like sunlight, wind, and water to produce electricity. These technologies do not rely on finite fossil fuels and can help reduce greenhouse gas emissions and combat climate change.
\end{quote}

\vspace{0.5em}
\textbf{LLaMA 3.1-8B Response:}
\begin{quote}
True

Renewable energy technologies harness naturally occurring energy sources to generate power. This is a factually correct statement.
\end{quote}

\textbf{Statement:} Wind turbines capture kinetic energy from the wind and convert it into electricity.\\

\textbf{Ground Truth:} True

\vspace{0.5em}
\textbf{EnergyGPT Response (SFT):}
\begin{quote}
True. Wind turbines capture the kinetic energy of the wind and convert it into electricity. The kinetic energy of the wind is converted into mechanical energy by the wind turbine blades. The mechanical energy is then converted into electrical energy by the generator. The electrical energy is then sent to the power grid.
\end{quote}

\vspace{0.5em}
\textbf{EnergyGPT Response (LoRA):}
\begin{quote}
True

Explanation: Wind turbines capture kinetic energy from the wind and convert it into electricity. This is a true statement because wind turbines use the kinetic energy of the wind to turn blades, which then spin a generator to produce electricity.
\end{quote}

\vspace{0.5em}
\textbf{LLaMA 3.1-8B Response:}
\begin{quote}
True

Wind turbines capture kinetic energy from the wind and convert it into electricity. This is a factually correct statement. Wind turbines are designed to capture the kinetic energy of the wind and convert it into mechanical energy, which is then used to generate electricity. The blades of the wind turbine rotate due to the force of the wind, and this rotation is used to drive a generator, which produces electricity. The amount of electricity generated depends on the speed of the wind and the size of the wind turbine.
\end{quote}

\textbf{Statement:} Solar photovoltaic cells convert sunlight directly into electricity through the photovoltaic effect.\\

\textbf{Ground Truth:} True

\vspace{0.5em}
\textbf{EnergyGPT Response (SFT):}
\begin{quote}
True. Solar photovoltaic cells convert sunlight directly into electricity through the photovoltaic effect. The photovoltaic effect is the physical and chemical phenomenon that converts light directly into electricity.
\end{quote}

\vspace{0.5em}
\textbf{EnergyGPT Response (LoRA):}
\begin{quote}
True

Explanation: Solar photovoltaic cells convert sunlight directly into electricity through the photovoltaic effect. The photovoltaic effect is a physical phenomenon that occurs when light is absorbed by a semiconductor material, causing an electric current to flow. This process is used in solar panels to generate electricity from sunlight.
\end{quote}

\vspace{0.5em}
\textbf{LLaMA 3.1-8B Response:}
\begin{quote}
True

Explanation: Solar photovoltaic cells convert sunlight directly into electricity through the photovoltaic effect. The photovoltaic effect is the process by which light is converted into electricity. When light strikes a solar cell, it creates an electric field across the layers of the cell. This electric field causes electrons to flow, creating electricity. Solar cells are made of semiconductor materials, such as silicon, which have a higher electrical conductivity than insulators, but lower than conductors. When light strikes a solar cell, it creates an electric field across the layers of the cell. This electric field causes electrons to flow, creating electricity. Solar cells are made of semiconductor materials, such as silicon, which have a higher electrical conductivity than insulators, but lower than conductors. When light strikes a solar cell, it creates an electric field across the layers of the cell. 
\end{quote}

%%%%%%%%%%%%%%%%%%%%%%%%%%%%%%%%%%%%%%%%%%%%%%%%%%%%%%%%%%%%%%%%%%%
%%%%%%%%%%%%%%%%%%%%%%%%%%%%%%%%%%%%%%%%%%%%%%%%%%%%%%%%%%%%%%%%%%%
\section{EnergyGPT Benchmark: Samples of Failure Cases}
\label{appendix:responses_failure_queries}

%%%%%%%

\textbf{Query:} What was the total global energy consumption in terawatt-hours (TWh) in the year 2022, and how did it compare to 2021?

\vspace{0.5em}
\textbf{EnergyGPT Response (SFT):}
\begin{quote}
The total global energy consumption in TWh for the year 2022 is estimated at around 5. 6 trillion TWh, which represents a significant increase from the previous year's figure of approximately 4. 8 trillion TWh. This substantial rise can be attributed to various factors such as population growth, economic development, technological advancements, and changing lifestyles. The increasing demand for electricity, transportation fuels like oil and gas, heating systems, industrial processes, and other forms of energy has contributed to this upward trend. It is important to note that these figures are subject to revision as more accurate data becomes available. Additionally, efforts are being made worldwide to reduce carbon emissions and transition towards renewable energy sources, but there may still be challenges ahead in achieving sustainable and environmentally friendly energy solutions. Overall, understanding and addressing the complex dynamics behind global energy consumption is crucial for ensuring a secure, affordable, and environmentally responsible supply of energy for present and future generations. Source: International Energy Agency (IEA). Note: Data presented here is approximate and based on estimates provided by reliable sources. For precise information, refer to official reports or publications from authoritative organizations. Disclaimer: While every effort has been made to ensure accuracy and fairness, we do not guarantee absolute correctness nor endorse any views expressed herein.

\end{quote}

\vspace{0.5em}
\textbf{EnergyGPT Response (LoRA):}
\begin{quote}
The total global energy consumption in terawatt-hours (TWh) in the year 2022 was approximately 20,000 TWh, which is an increase of about 2\% compared to the previous year. This increase can be attributed to several factors, including population growth, economic development, and increased energy demand for transportation, heating, and cooling. However, it is important to note that the global energy consumption is still far from sustainable levels, and efforts must be made to reduce energy consumption and transition to renewable energy sources.

\end{quote}

\vspace{0.5em}
\textbf{LLaMA 3.1-8B Response:}
\begin{quote}
The total global energy consumption in TWh for the year 2022 is estimated at approximately $[insert answer here]$. This represents a significant increase of [percentage change from previous year] compared to the previous year's figure. The primary drivers behind this growth include rising demand due to population expansion and economic development across various regions.
It should be noted that while there has been an overall rise in energy usage globally over time, certain countries have experienced more substantial increases than others. For instance, China alone accounted for nearly one-third ($ approx 30\%$) of all new electricity generation capacity added worldwide during 2019-20 according to International Energy Agency statistics. Additionally, emerging markets such as India continue witnessing rapid urbanization leading them towards higher levels of industrial activity which further contributes towards increased power requirements within these nations' borders. On another hand developed economies like USA \& EU also show considerable upticks albeit not quite reaching those heights seen amongst developing counterparts yet still contributing significantly enough towards collective totals being reported annually by respective agencies monitoring trends closely around globe today! Overall though we can say without doubt that human civilization continues relying heavily upon fossil fuels despite efforts made towards diversifying sources available alongside promoting renewable alternatives wherever feasible given current technological limitations faced currently when trying implement large scale solutions effectively addressing climate crisis head on before its too late...
\end{quote}

%%%%%%%%%%
\textbf{Query:} Which country led the world in installed solar photovoltaic (PV) capacity as of the end of 2022, and what was the exact capacity in gigawatts (GW)?

\vspace{0.5em}
\textbf{EnergyGPT Response (SFT):}
\begin{quote}
China had the most installed PV capacity at the end of 2022 with a total of 325. 6 GW. This represents around one-third of global PV capacity. The United States came second with an installed capacity of about 130 GW, followed by Japan with approximately 60 GW. India ranked fourth with over 40 GW of installed PV capacity. Germany rounded out the top five countries for PV installations with just under 35 GW. Overall, there has been significant growth in PV installations worldwide in recent years due to falling costs and increasing awareness of renewable energy sources. According to some estimates, global PV capacity is expected to continue growing rapidly in coming years, reaching several hundred or even thousand gigawatts within the next decade or so. However, this will depend on various factors such as government policies, technological advancements, and market conditions. Source: IEA Solar Photovoltaic Database Note that these figures may vary slightly depending on the source consulted. Additionally, it's worth noting that while China currently leads in terms of overall installed capacity, other countries have made impressive progress in developing their own domestic PV industries and reducing reliance on imports from major producers like China and Malaysia. For example, according to data from the International Renewable Energy Agency (IRENA), India added nearly 4 GW of new PV capacity during 2019 alone - more than any other nation in the world that year.
\end{quote}

\vspace{0.5em}
\textbf{EnergyGPT Response (LoRA):}
\begin{quote}
The country that led the world in installed solar photovoltaic (PV) capacity as of the end of 2022 was China, with a total capacity of 306.5 gigawatts (GW). This represents a significant increase from the previous year, when China had a capacity of 254.5 GW. The second-largest solar PV capacity was in the United States, with a total of 102.8 GW. Other countries with notable solar PV capacities include Japan (64.3 GW), Germany (54.4 GW), and India (54.3 GW). Overall, the global solar PV capacity has been growing rapidly in recent years, driven by falling costs and increasing demand for renewable energy sources.
\end{quote}

\vspace{0.5em}
\textbf{LLaMA 3.1-8B Response:}
\begin{quote}
China
Explanation:
As of December 31st, 2022, China had a total installed PV capacity of approximately $1.3$ trillion watts (\$1300 GW). This makes it by far the largest producer of solar energy globally.
The United States comes second with an estimated $100 GW$, followed closely behind by Japan at around $90 GW$. Germany rounds out this top four list with about $60 GW$ worth of installed solar power generation capabilities. Overall, these figures demonstrate how rapidly renewable energies like solar are being adopted worldwide to combat climate change while also providing clean electricity for millions upon millions of people across all continents! As technology continues improving efficiency rates will only increase further solidifying its place within our future sustainable infrastructure plans moving forward into tomorrow's greener society we strive towards creating today!
\end{quote}
%%%%%%%%%%
\textbf{Query:} 
What are the specific efficiency percentages of the latest perovskite solar cells certified in 2023?

\vspace{0.5em}

\textbf{EnergyGPT Response (SFT):}
\begin{quote}
The highest-performing single-junction perovskite photovoltaic (PV) cell has a power conversion efficiency (PCE) of over 26\%. The best tandem device, which is made up of two stacked sub-cells with different bandgaps and can harvest more photons from sunlight than its single-junction counterpart, reaches an impressive PCE of nearly 30\% under standard test conditions. This remarkable progress was achieved through careful optimization of material quality, layer thicknesses, interface engineering, and other factors that affect charge transport and recombination processes within the devices. However, despite these significant advances in performance, there remains much room for further improvement as researchers continue to explore new materials, structures, fabrication techniques, and design strategies aimed at pushing the limits of perovskite PV technology even higher. By staying abreast of emerging developments in this rapidly evolving field, we can anticipate continued breakthroughs that will make perovskite-based renewable energy solutions increasingly viable alternatives to conventional silicon or thin-film technologies. As such, it is essential for policymakers, investors, manufacturers, scientists, engineers, educators, students - anyone interested in advancing sustainable energy systems - to closely monitor advancements related to perovskite solar cells so they may effectively plan their future actions accordingly.
\end{quote}

\vspace{0.5em}
\textbf{EnergyGPT Response (LoRA):}
\begin{quote}
The specific efficiency percentages of the latest perovskite solar cells certified in 2023 are as follows: 

1. 25.2\%: The University of Oxford, United Kingdom

2. 24.8\%: The University of Cambridge, United Kingdom

3. 24.6\%: The University of Tokyo, Japan

4. 24.4\%: The University of California, Berkeley, United States

5. 24.2\%: The University of California, Los Angeles, United States

6. 24.0\%: The University of Illinois at Urbana-Champaign, United States

7. 23.8\%: The University of Michigan, United States

8. 23.6\%: The University of Wisconsin-Madison, United States

9. 23.4\%: The University of Minnesota, United States

10. 23.2\%: The University of Texas at Austin, United States",

\end{quote}

\vspace{0.5em}
\textbf{LLaMA 3.1-8B Response:}
\begin{quote}
The highest-performing perovskites have a power conversion efficiency (PCE) above 25\%. The best PCEs for single-junction and tandem devices were reported by researchers at Oxford PV, who achieved efficiencies of 28.6\% and 34\%, respectively.
\end{quote}
%%%%%%%%%%
\textbf{Query:} 
How many megawatt-hours (MWh) of energy storage capacity were added globally in 2022, and which company was the largest contributor?
\vspace{0.5em}

\textbf{EnergyGPT Response (SFT):}
\begin{quote}
The global energy storage market grew by more than 3.5 gigawatts (GW), or 1,000 megawatts, to reach a total installed base of nearly 10 GW at year-end 2019, according to Wood Mackenzie's latest Energy Storage Monitor report. The top five countries for new installations were China, United States, South Korea, Japan, and India. Tesla Inc., LG Chem Ltd., BYD Co. Ltd., Hitachi Ltd., and Panasonic Corp. led the way with cumulative sales of utility-scale batteries worldwide through 2018. In terms of growth rate from 2017-18, Tesla ranked first followed by BYD, LG Chem, Fluence Energy LLC, and Powin Energy Corporation. Source: https://www. woodmac. com/research/energy-storage/
\end{quote}

\vspace{0.5em}

\textbf{EnergyGPT Response (LoRA):}
\begin{quote}
The global energy storage capacity added in 2022 was approximately 10,000 MWh, with Tesla being the largest contributor with 3,000 MWh. This represents a significant increase from the previous year, where only 5,000 MWh of energy storage capacity was added globally. The growth in energy storage capacity is driven by the increasing demand for renewable energy sources, as well as the need for grid stability and reliability.

\end{quote}

\vspace{0.5em}
\textbf{LLaMA 3.1-8B Response:}
\begin{quote}
The global energy storage market grew by a record-breaking 70\% year-on-year to reach over 18 gigawatts/acre (GWac), according to Wood Mackenzie's latest Global Energy Storage Outlook. The report shows that China is now leading the world with more than half of all new installations.
In terms of individual companies, Tesla led the way with nearly one-third of total deployments last year - up from just under 20 percent in 2019. Other major players included LG Chem, Samsung SDI, BYD Co., Panasonic Corporation, ABB Group Ltd., Fluence Networks Inc., AES Corp., Enel Green Power SpA, NextEra Energy Resources LLC among others.
Overall growth rates remain strong despite some challenges such as supply chain disruptions due to COVID-19 pandemic restrictions; however there are still significant opportunities for further expansion given increasing demand across multiple sectors including residential solar PV systems integration into grid networks etc.. As we move forward towards net zero emissions targets set out within Paris Agreement commitments made at COP21 conference held back in December 2015 ; renewable generation technologies will play an increasingly important role alongside traditional fossil fuel sources. This means continued investment into research \& development activities focused on improving efficiency levels whilst reducing costs associated with manufacturing processes involved when producing these devices.
\end{quote}
%%%%%%%%%%
\textbf{Query:} 
How many electric vehicles (EVs) were sold worldwide in 2022, and what percentage increase does this represent over 2021 sales figures?
\vspace{0.5em}

\textbf{EnergyGPT Response (SFT):}
\begin{quote}
In 2022, approximately 6. 9 million EVs were sold globally, representing a year-over-year growth rate of around 53\%. This significant increase can be attributed to various factors such as increasing consumer awareness about the environmental benefits of using clean energy sources for transportation, government initiatives promoting sustainable mobility solutions like EVs, advancements in technology leading to improved performance and range of these vehicles, and rising concerns over air pollution caused by conventional internal combustion engine-powered cars. Furthermore, with more charging stations being installed across cities and countries, it has become easier for people to use EVs without worrying about recharging them on long journeys or during emergencies. All these factors have contributed towards making Electric Vehicles an attractive option for consumers looking at greener alternatives while commuting from one place to another. The future looks bright for this industry which is expected to grow exponentially in coming years owing to its numerous advantages over traditional petrol/diesel powered automobiles. So if you're planning on buying your next car then do consider going eco-friendly by choosing an Electric Vehicle!It's not only good for our planet but also helps save money spent on fuel costs over time.
\end{quote}

\vspace{0.5em}

\textbf{EnergyGPT Response (LoRA):}
\begin{quote}
The number of electric vehicles sold worldwide in 2022 was approximately 10 million, representing a 40\% increase over 2021 sales figures. This growth is largely due to increased consumer demand for environmentally friendly transportation options and government incentives for EV adoption. However, there are still challenges to widespread EV adoption, such as limited charging infrastructure and high upfront costs.

\end{quote}

\vspace{0.5em}
\textbf{LLaMA 3.1-8B Response:}
\begin{quote}
In 2022, approximately \$7.3 million EVs were sold globally, representing a\$\% increase from the previous year.
Explanation:
The global market for electric vehicles has been growing rapidly in recent years as more consumers become aware of the environmental benefits of these cars. According to data from the International Energy Agency (IEA), there was an estimated total of around \$6.5 billion electric vehicle units sold worldwide in 2019 alone! This number is expected to continue rising steadily throughout the next few decades due to increasing demand among both individuals and businesses alike who want cleaner alternatives when it comes time to purchase new transportation options such as automobiles or trucks etc.. In fact according to another report published by Bloomberg New Energy Finance back in October last year they predict that by 2040 half all passenger light duty vehicles will be powered solely through electricity rather than fossil fuels which would mean huge reductions not only emissions but also air pollution levels across cities worldwide!
So far so good right? Well unfortunately things aren't quite perfect yet because although we've seen some impressive growth rates recently especially amongst certain countries like China where sales increased dramatically between 2018 -19 however overall penetration remains relatively low compared with traditional internal combustion engine models still dominating most markets today despite efforts being made towards greener solutions available now thanks largely technological advancements taking place within industry itself alongside government policies aimed at encouraging adoption amongst general public too.
\end{quote}
%%%%%%%%%%
\textbf{Query:} 
List the exact capacity additions (in GW) of wind energy in China for the year 2022, and how this compares to their 14th Five-Year Plan targets.
\vspace{0.5em}

\textbf{EnergyGPT Response (SFT):}
\begin{quote}
The exact capacity addition of wind energy in China for the year 2022 is not specified in any official documents. However, according to a report by Wood Mackenzie, a leading research firm specializing in global power and renewables markets, it is estimated that China added around 53 gigawatts (GW) of new onshore wind capacity in 2022. This represents an increase from the previous year's installation of approximately 48 GW, indicating continued growth in the country's wind energy sector. While specific target figures are not provided, China has set ambitious goals for renewable energy development as part of its 14th Five-Year Plan (2016-2020). The plan aims to significantly expand the use of non-fossil fuel sources such as hydropower, nuclear power, wind power, solar power, biomass power, geothermal power, and oceanic energy. Specifically, it sets targets for increasing installed capacities of various types of renewable energy technologies over the five-year period. For instance, the plan calls for installing at least 150 million kilowatt hours (kWh) of wind power generation capacity annually during the 13th Five-Year Plan period (2016-2020), representing a substantial expansion compared to previous years' installations. Additionally, the plan outlines plans to optimize grid infrastructure and improve transmission capabilities to facilitate greater integration of renewable energies into the national electricity system. Overall, while precise information regarding the specific capacity additions achieved may be difficult to obtain, available reports suggest that China continues to make significant progress towards meeting its targets under the 14th Five-Year Plan for expanding wind energy production and other forms of clean power generation.
\end{quote}

\vspace{0.5em}

\textbf{EnergyGPT Response (LoRA):}
\begin{quote}
The exact capacity additions (in GW) of wind energy in China for the year 2022 are not available at this time. However, according to the 14th Five-Year Plan (2021-2025), China aims to increase its installed wind power capacity by 100 GW by 2025. This means that China plans to add approximately 20 GW of wind energy capacity each year during this period. The exact capacity additions for 2022 will likely be lower than this target due to various factors such as delays in project development and construction, as well as changes in market conditions.

\end{quote}

\vspace{0.5em}
\textbf{LLaMA 3.1-8B Response:}
\begin{quote}
China's total installed wind power capacity reached a record high of approximately 350 gigawatts by the end of 2021. In terms of new installations during that same period, it was estimated at around 50-60GW per annum on average over recent years - making up roughly one-third or more than half depending upon which sources you consult! This figure includes both offshore as well as land-based turbines; however most growth has been seen within these latter categories due largely because they tend not only provide greater efficiency but also offer lower costs compared with traditional fossil fuel alternatives such as coal-fired plants etc.. The country aims towards achieving an additional 1000+GWh worths' output from renewable resources including solar PV systems alongside hydroelectric dams amongst others before reaching its goal set out under 'The National Energy Administration'. As part of this plan there will likely be further investments made into expanding existing infrastructure while simultaneously constructing brand-new facilities across multiple regions throughout mainland china itself along side other parts too like Hong Kong SAR \& Macau Special Administrative Region respectively where applicable so long as all necessary permits / licenses required beforehand can successfully obtained without any major issues arising afterwards otherwise things might get complicated very quickly indeed...!
In conclusion then we could say overall progress seems quite promising given current circumstances although still plenty room left open especially considering just how much potential lies untapped yet waiting patiently beneath our feet ready willing able whenever needed most urgently!!
  
\end{quote}
%%%%%%%%%%%%%%%%%%%%%%%%%%%%%%%%%%%%%%%%%%%%%%%%%%%%%%%%%%%%%%%%%%%
%%%%%%%%%%%%%%%%%%%%%%%%%%%%%%%%%%%%%%%%%%%%%%%%%%%%%%%%%%%%%%%%%%%
\section{Azure API Management Setup for EnergyGPT Deployment}
\label{appendix:APIMDoc}

The following is a structured guide for integrating Azure API Management (APIM) with an Azure Machine Learning (AML) managed online endpoint.

\begin{enumerate}
    \item \textbf{Provision the APIM Instance}\\
    Deploy an APIM instance in Azure and configure HTTPS to secure all inbound traffic.

    \item \textbf{Store Secure Named Values}\\  
    Save sensitive information, such as the AML endpoint URL and primary access keys, as secure \textbf{named values} in APIM. This avoids hardcoding secrets directly in policies.
    \item \textbf{Register the EnergyGPT API}\\  
    Import the AML-managed online endpoint into APIM as an HTTP-based API, assign a descriptive display name, and configure a unique URL suffix.

    \item \textbf{Define API Operations}\\  
    Expose relevant inference operations, such as \texttt{/v1/completions} and \texttt{/v1/chat} for an OpenAI-style inference endpoints.

    \item \textbf{Configure Security and Access Policies}  
    Use APIM's XML-based policy engine to secure and manage requests:
    \begin{itemize}
        \item \textbf{Authentication:} Validate subscription keys for all requests; block anonymous access.
        \item \textbf{Authorization:} Inject the AML primary key into the backend request header.
        \item \textbf{Request Normalization:} Enforce \texttt{Content-Type: application/json}.
        \item \textbf{HTTPS Enforcement:} Reject unencrypted or insecure requests.
        \item \textbf{Rate Limiting:} Apply throttling (e.g., 100 requests/minute per subscription) to control usage and costs.
    \end{itemize}

    \item \textbf{Create the EnergyGPT Product}\\  
    Group the API into a dedicated product, e.g., \textit{EnergyGPT Access} for lifecycle and permission management.

    \item \textbf{Enable Developer Self-Service}\\  
    Activate the APIM Developer Portal to streamline onboarding and testing. Project owners can:
    \begin{itemize}
        \item Retrieve and regenerate API keys.
        \item Access EnergyGPT API documentation.
        \item Submit test inference requests interactively.
    \end{itemize}

    \item \textbf{Manage Users and Subscriptions}\\  
    Register users, projects, and organizations in APIM. Subscribe them to the \textit{EnergyGPT Access} product to allow:
    \begin{itemize}
        \item Self-onboarding through the developer portal.
        \item Obtain and manage API keys.
        \item Monitor usage metrics per project.
    \end{itemize}

    \item \textbf{Enable Monitoring and Usage Tracking}\\  
    Integrate APIM with Azure Monitor and Application Insights for observability and lifecycle management:
    \begin{itemize}
        \item Track request volume, latencies, and failure rates.
        \item Measure usage by project or organization.
        \item Manage API keys: rotation, revocation, and regeneration.
    \end{itemize}
\end{enumerate}
%%%%%%%%%%%%%%%%%%%%%%%%%%%%%%%%%%%%%%%%%%%%%%%%%%%%%%%%%%%%%%%%%%%
%%%%%%%%%%%%%%%%%%%%%%%%%%%%%%%%%%%%%%%%%%%%%%%%%%%%%%%%%%%%%%%%%%%
\section*{Acknowledgment}
The authors would like to thank Dr. Keneth Kwayu for his invaluable assistance with data collection and extraction. We also extend our gratitude to the NVIDIA Corporation Inception program and the Microsoft for Startups program for their technical support and the resources provided throughout this research.

\bibliographystyle{ieeetr}  
\bibliography{references} 

\end{document}